%% file: main.tex
\documentclass{article}
\pdfoutput=1
\input{imports}

\title{\papertitle}
\date{}
\author[1,*,$\dagger$]{Shayegan~Omidshafiei}
\author[1,*]{Karl~Tuyls}
\author[2]{Wojciech~M.~Czarnecki}
\author[3]{Francisco~C.~Santos}
\author[2]{Mark~Rowland}
\author[2]{Jerome~Connor}
\author[1]{Daniel~Hennes}
\author[1]{Paul~Muller}
\author[1]{Julien~P\'erolat}
\author[1]{Bart~De~Vylder}
\author[2]{Audrunas~Gruslys}
\author[1]{R\'emi~Munos}

\affil[1]{DeepMind, Paris, France}
\affil[2]{DeepMind, London, UK}
\affil[3]{INESC-ID and Instituto Superior T\'ecnico, Universidade de Lisboa, Portugal}
\affil[*]{Equal contributors.}
\affil[$\dagger$]{Corresponding author.}

\begin{document}

\maketitle
\input{main_text.tex}

\clearpage
\bibliographystyle{plainnat}
\bibliography{references}
\clearpage

\input{main_additional_info.tex}

\newpage
\section*{\centering Supplementary Information:\\ \papertitle}
\input{supplementary_info_text.tex}

\end{document}

%% file: imports.tex
\usepackage[utf8]{inputenc}
\usepackage{hyperref}
\usepackage{comment}
\usepackage[labelfont=bf,labelformat=simple,singlelinecheck=off,justification=raggedright]{subcaption}
\DeclareCaptionLabelFormat{bold}{\textbf{#2}}
\usepackage[pdftex]{graphicx}
\usepackage[super,comma,sort]{natbib}
\usepackage{multirow}
\usepackage{booktabs}
\usepackage{nicefrac}
\usepackage[margin=1in]{geometry}
\usepackage{xcolor}
\usepackage{xspace}
\usepackage{nameref}
\usepackage{authblk}
\usepackage{array}
\usepackage{lineno}

\input{math_commands.tex}

\newcommand{\papertitle}{Navigating the Landscape of Multiplayer Games}

\newcommand{\bi}{\begin{itemize}}
\newcommand{\ei}{\end{itemize}}
\newcommand{\be}{\begin{enumerate}}
\newcommand{\ee}{\end{enumerate}}
\newcommand{\bb}{\begin{block}}
\newcommand{\eb}{\end{block}}

\newcommand{\alpharank}{$\alpha$-Rank\xspace}

\newcommand{\reals}{\mathbb{R}}

\usepackage{tikz}
\usetikzlibrary{shapes,arrows.meta,calc,fit,backgrounds,shapes.multipart,positioning}
\tikzset{
    ib/.style={
    rectangle split, rectangle split parts=2,
    rectangle split draw splits=false,
    rounded corners=\blockroundedcorners,
    line width=\blocklinewidth,
    minimum height=\IBheight,
    minimum width=\IBwidth,
    text width=\IBwidth-\blocklinewidth*2,
    draw=#1,
    rectangle split part fill={#1, none},
    text=white, text centered
    },
    every two node part/.style={align=left, text=black}
}
\newlength\IBheaderheight 
\newlength\IBwidth 
\newlength\IBheight 
\newlength\blockroundedcorners
\newlength\blocklinewidth
\definecolor{myblue}{RGB}{18,54,147}
\definecolor{myteal}{RGB}{0,190,200}

\newtheorem{theorem}{Theorem}

\newtheorem{definition}[theorem]{Definition}

\usepackage[capitalise,noabbrev]{cleveref} 
\crefname{equation}{}{} 

\newlength{\tempdima}
\newcommand{\rowname}[1]{
    \rotatebox{90}{\hspace{-\tempdima}\makebox[\tempdima][c]{#1}}
}

\def\gameSubfigWidth{0.15\textwidth}

\newcommand{\gameRowPhantomSubcaptions}[1]{
    \centering{
        \phantomsubcaption\label{fig:#1_plot_payoffs}
        \phantomsubcaption\label{fig:#1_plot_networkx}
        \phantomsubcaption\label{fig:#1_plot_cycles_histogram}
        \phantomsubcaption\label{fig:#1_plot_networkx_spectral}
        \phantomsubcaption\label{fig:#1_plot_networkx_spectral_clustered}
        \phantomsubcaption\label{fig:#1_plot_networkx_spectral_clustered_contracted}
    }
}

\def\gameSubfigWidth{0.15\textwidth}
\newcommand{\gameRowMotivating}[2]{
    \settoheight{\tempdima}{\includegraphics[width=\gameSubfigWidth]{figs/motivating_example/complexity_#1_plot_payoffs.pdf}}%
    \rowname{#2}&
    \subcaptionbox{\label{fig:#1_plot_payoffs}}{%
        \centering%
        \includegraphics[width=\gameSubfigWidth]{figs/motivating_example/complexity_#1_plot_payoffs.pdf}}&
    \subcaptionbox{\label{fig:#1_plot_networkx}}{%
        \centering%
        \includegraphics[width=\gameSubfigWidth]{figs/motivating_example/complexity_#1_plot_networkx.pdf}%
    }&
    \subcaptionbox{\label{fig:#1_plot_cycles_histogram}}{%
        \centering%
        \includegraphics[width=\gameSubfigWidth]{figs/motivating_example/complexity_#1_plot_cycles_histogram.pdf}%
    }&
    \subcaptionbox{\label{fig:#1_plot_networkx_spectral}}{%
        \centering%
        \includegraphics[width=\gameSubfigWidth]{figs/motivating_example/complexity_#1_plot_networkx_spectral.pdf}%
    }&
    \subcaptionbox{\label{fig:#1_plot_networkx_spectral_clustered}}{%
        \centering%
        \includegraphics[width=\gameSubfigWidth]{figs/motivating_example/complexity_#1_plot_networkx_spectral_clustered.pdf}%
    }&
    \subcaptionbox{\label{fig:#1_plot_networkx_spectral_clustered_contracted}}{%
        \centering%
        \includegraphics[width=\gameSubfigWidth]{figs/motivating_example/complexity_#1_plot_networkx_spectral_clustered_contracted.pdf}%
    }\\
}

\def\gameSubfigWidthLarge{0.22\textwidth}
\newcommand{\gameRowFiletype}[3]{
    \settoheight{\tempdima}{\includegraphics[width=\gameSubfigWidthLarge]{figs/complexity_#1_plot_payoffs.#3}}%
    \rowname{#2}&
    \subcaptionbox{\label{fig:#1_plot_payoffs}}{%
        \centering%
        \includegraphics[width=\gameSubfigWidthLarge]{figs/complexity_#1_plot_payoffs.#3}}&
    \subcaptionbox{\label{fig:#1_plot_networkx_spectral}}{%
        \centering%
        \includegraphics[width=\gameSubfigWidthLarge]{figs/complexity_#1_plot_networkx_spectral.#3}%
    }&
    \subcaptionbox{\label{fig:#1_plot_networkx_spectral_clustered}}{%
        \centering%
        \includegraphics[width=\gameSubfigWidthLarge]{figs/complexity_#1_plot_networkx_spectral_clustered.#3}%
    }&
    \subcaptionbox{\label{fig:#1_plot_networkx_spectral_clustered_contracted}}{%
        \centering%
        \includegraphics[width=\gameSubfigWidthLarge]{figs/complexity_#1_plot_networkx_spectral_clustered_contracted.#3}%
    }\\
}

\newcommand{\gameRowFiletypePhantomSubcaptions}[1]{
    \centering{
        \phantomsubcaption\label{fig:#1_plot_payoffs}
        \phantomsubcaption\label{fig:#1_plot_networkx_spectral}
        \phantomsubcaption\label{fig:#1_plot_networkx_spectral_clustered}
        \phantomsubcaption\label{fig:#1_plot_networkx_spectral_clustered_contracted}
    }
}

\def\gameSubfigWidthLarge{0.22\textwidth}
\newcommand{\gameRowFiletypeNoPayoffs}[3]{
    \settoheight{\tempdima}{\includegraphics[width=\gameSubfigWidthLarge]{figs/complexity_#1_plot_networkx_spectral.#3}}%
    \rowname{#2}&
    \subcaptionbox{\label{fig:#1_plot_networkx}}{%
        \centering%
        \includegraphics[width=\gameSubfigWidthLarge]{figs/complexity_#1_plot_networkx.#3}%
    }&
    \subcaptionbox{\label{fig:#1_plot_networkx_spectral}}{%
        \centering%
        \includegraphics[width=\gameSubfigWidthLarge]{figs/complexity_#1_plot_networkx_spectral.#3}%
    }&
    \subcaptionbox{\label{fig:#1_plot_networkx_spectral_clustered}}{%
        \centering%
        \includegraphics[width=\gameSubfigWidthLarge]{figs/complexity_#1_plot_networkx_spectral_clustered.#3}%
    }&
    \subcaptionbox{\label{fig:#1_plot_networkx_spectral_clustered_contracted}}{%
        \centering%
        \includegraphics[width=\gameSubfigWidthLarge]{figs/complexity_#1_plot_networkx_spectral_clustered_contracted.#3}%
    }\\
}

\newcolumntype{Y}{>{\centering\arraybackslash}X}

\newcolumntype{M}[1]{>{\centering\arraybackslash\hspace{0pt}}p{#1}}

%% file: math_commands.tex
\usepackage{amsmath,amsfonts,bm}

\def\eqref#1{equation~\ref{#1}}

\def\1{\bm{1}}

\def\vc{{\bm{c}}}

\def\ve{{\bm{e}}}

\def\vv{{\bm{v}}}

\def\vpi{{\bm{\pi}}}

\def\mA{{\mathbf{A}}}

\def\mD{{\mathbf{D}}}
\def\mE{{\mathbf{E}}}

\def\mI{{\mathbf{I}}}

\def\mL{{\mathbf{L}}}
\def\mM{{\mathbf{M}}}

\def\mOmega{{\bm{\Omega}}}

\DeclareMathAlphabet{\mathsfit}{\encodingdefault}{\sfdefault}{m}{sl}
\SetMathAlphabet{\mathsfit}{bold}{\encodingdefault}{\sfdefault}{bx}{n}

\def\emM{{M}}



%% file: main_text.tex
\begin{abstract}
Multiplayer games have long been used as testbeds in artificial intelligence research, aptly referred to as the Drosophila of artificial intelligence. Traditionally, researchers have focused on using well-known games to build strong agents. This progress, however, can be better informed by characterizing games and their topological landscape. Tackling this latter question can facilitate understanding of agents and help determine what game an agent should target next as part of its training. Here, we show how network measures applied to response graphs of large-scale games enable the creation of a landscape of games, quantifying relationships between games of varying sizes and characteristics. We illustrate our findings in domains ranging from canonical games to complex empirical games capturing the performance of trained agents pitted against one another. Our results culminate in a demonstration leveraging this information to generate new and interesting games, including mixtures of empirical games synthesized from real world games.
\end{abstract}

\section*{Introduction}

Games have played a prominent role as platforms for the development of learning algorithms and the measurement of progress in artificial intelligence (AI)~\citep{turing1953digital,samuel1960programming,schaeffer2001gamut,yannakakis2018artificial}.
Multiplayer games, in particular, have played a pivotal role in AI research and have been extensively investigated in machine learning, ranging from abstract benchmarks in game theory over popular board games such as Chess~\citep{campbell2002deep,silver2017mastering} and Go~\citep{Silver2016} (referred to as the Drosophila of AI research~\citep{mccarthy1997ai}), to realtime strategy games such as StarCraft~II~\citep{vinyals19} and Dota~2~\citep{Berner19}. 
Overall, AI research has primarily placed emphasis on training of strong agents;
we refer to this as the Policy Problem, which entails the search for super human-level AI performance.
Despite this progress, the need for a task theory, a framework for taxonomizing, characterizing, and decomposing AI tasks has become increasingly important in recent years~\citep{thorisson2016artificial,hernandez2017new}.
Naturally, techniques for understanding the space of games are likely beneficial for the algorithmic development of future AI entities~\citep{thorisson2016artificial,clune2019ai}.
Understanding and decomposing the characterizing features of games can be leveraged for downstream training of agents via curriculum learning~\citep{bengio2009curriculum}, which seeks to enable agents to learn increasingly-complex tasks.

A core challenge associated with designing such a task theory has been recently coined the Problem Problem, defined as ``the engineering problem of generating large numbers of interesting adaptive environments to support research"~\citep{leibo2019autocurricula}.
Research associated with the Problem Problem has a rich history spanning over 30 years, including the aforementioned work on task theory~\citep{thorisson2016artificial,hernandez2017new,bieger2018task}, procedurally-generated videogame features~\citep{toy1980,braben1984elite,nomanssky2016}, generation of games and rule-sets for General Game Playing~\citep{genesereth2005general,browne2010evolutionary,togelius2014characteristics,togelius2008experiment,kowalski2016evolving,nelson2016rules,perez2019general}, and procedural content generation techniques~\citep{togelius2011search,shaker2016procedural,risi2019increasing,smith2011answer,nelson2007towards,cook2011multi,cook2016angelina,juliani2019obstacle,wang2019paired,wang2020enhanced};
we refer readers to Supplementary Note 1 for detailed discussion of these and related works.
An important question that underlies several of these interlinked fields is: 
what makes a game interesting enough for an AI agent to learn to play? 
Resolving this requires techniques that can characterize the topological landscape of games, which is the topic of interest in this paper.
We focus, in particular, on the characterization of multiplayer games (i.e., those involving interactions of multiple agents), and henceforth use the shorthand of games to refer to this class.

\begin{figure}[t]
	\centering
	\includegraphics[width=0.8\textwidth]{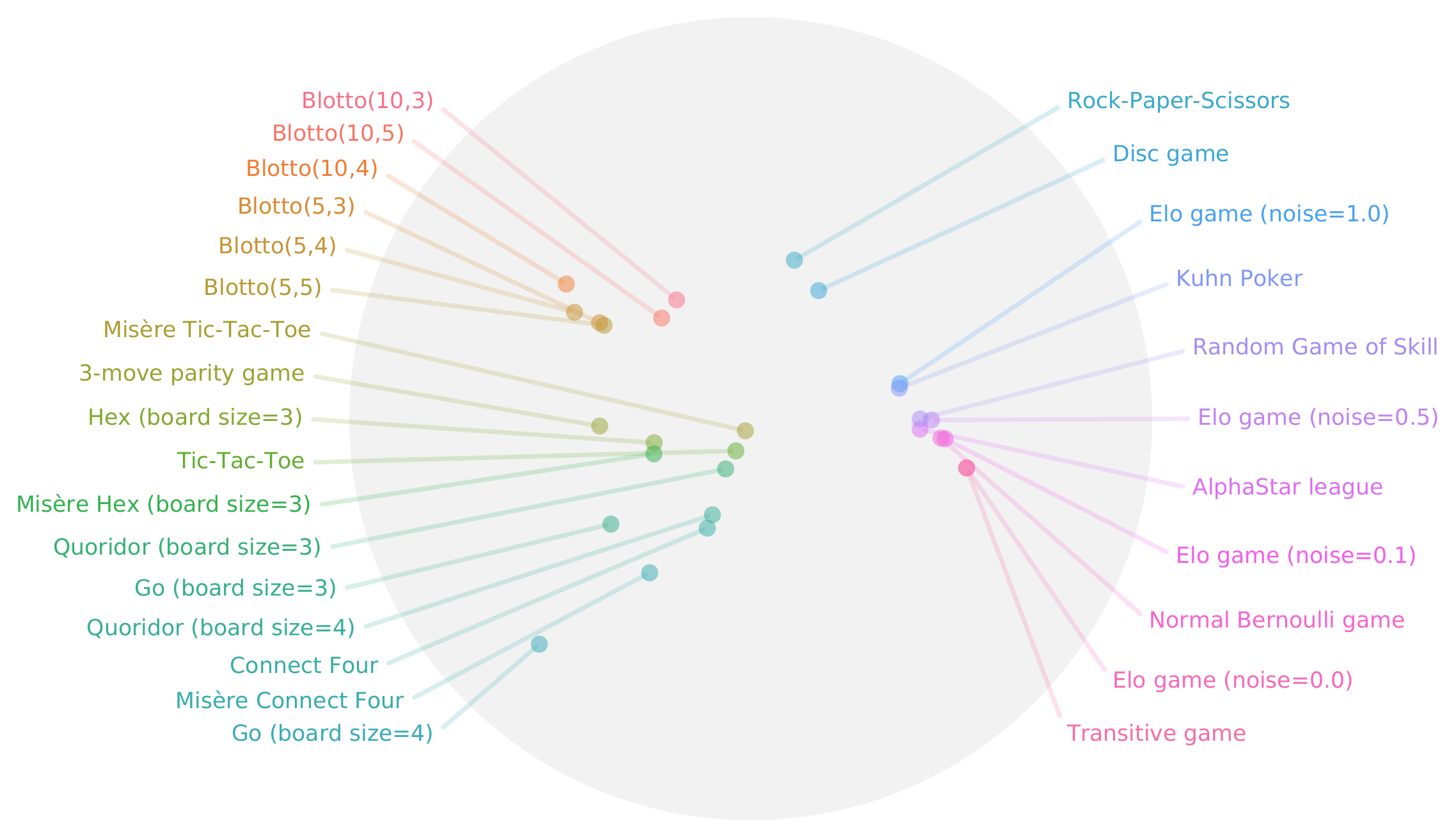}
	\caption{A landscape of games revealed by the proposed response graph-based workflow. 
	This landscape is generated by collecting features associated with the response graph of each game, and plotting the top two principal components.
	At a high level, games whose response graphs are characteristically similar are situated close to one another in this landscape.
    Notably, variations of games with related rules are well-clustered together, indicating strong similarity despite their widely-varying raw sizes.
    Instances of Blotto cluster together, despite their payoff table sizes ranging from $20 \times 20$ for Blotto(5,3) to $1000 \times 1000$ for Blotto(10,5).
    Games with strong transitive components (e.g., variations of Elo games, AlphaStar League, Random Game of Skill, and Normal Bernoulli Game) can be observed to be well separated from strongly cyclical games (Rock--Paper--Scissors and the Disc game).
    Closely-related real-world games (i.e., games often played by humans in the real world, such as Hex, Tic-Tac-Toe, Connect Four and each of their respective Mis\`ere counterparts) are also clustered together.
    }
	\label{fig:rwg_embeddings}
\end{figure}

The objective of this paper is to establish tools that enable discovery of a topology over games, regardless of whether they are interesting or not;
we do not seek to answer the interestingness question here, although such a toolkit can be useful for 
subsequently considering it.
Naturally, many notions of what makes a game interesting exist, from the perspectives of human-centric game design, developmental learning, curriculum learning, AI training, and so on. 
Our later experiments link to the recent work of \citet{czarnecki2020}, which investigated properties that make a game interesting specifically from an AI training perspective, as also considered here. 
We follow the interestingness characterization of \citet{czarnecki2020}, which defines so-called Games of Skill that are engaging for agents due to:
i) a notion of progress;
ii) availability of diverse play styles that perform similarly well.
We later show how clusters of games discovered by our approach align with this notion of interestingness.
An important benefit of our approach is that it applies to adversarial and cooperative games alike.
Moreover, while the procedural game structure generation results we later present target zero-sum games due to the payoff parameterization chosen in those particular experiments, they readily extend to general-sum games.

How does one topologically analyze games? 
One can consider characterizations of a game as quantified by measures such as the number of strategies available, players involved, whether the game is symmetric, and so on.
One could also order the payouts to players to taxonomize games, as done in prior works exploring $2 \times 2$ games~\citep{rapoport1966taxonomy,liebrand1983classification,bruns2015names}.
For more complex games, however, such measures are crude, failing to disambiguate differences in similar games.
One may also seek to classify games from the standpoint of computational complexity.
However, a game that is computationally challenging to solve may not necessarily be interesting to play.
Overall, designation of a single measure characterizing games is a non-trivial task.

It seems useful, instead, to consider measures that characterize the possible strategic interactions in the game.
A number of recent works have considered problems involving such interactions~\citep{balduzzi2018re,BellemareNVB13,MachadoBTVHB18,omidshafiei2019alpha,rowland2019multiagent,balduzzi2019open,Muller2020A,lanctot2017unified,HO2017,Hernandez-Orallo17,Hernandez-OralloFF12}.
Many of these works analyze agent populations, relying on game-theoretic models capturing pairwise agent relations.
Related models have considered transitivity (or lack thereof) to study games from a dynamical systems perspective~\citep{tuyls2018generalised,balduzzi2018mechanics};
here, a transitive game is one where strategies can be ordered in terms of strength, whereas an intransitive game may involve cyclical relationships between strategies (e.g., Rock--Paper--Scissors).
Fundamentally, the topology exposed via pairwise agent interactions seems a key enabler of the powerful techniques introduced in the above works.
In related literature, graph theory is well-established as a framework for topological analysis of large systems involving interacting entities~\citep{van2010graph,dehmer2010structural,boccaletti2006complex}.
Complexity analysis via graph-theoretic techniques has been applied to social networks~\citep{scott1988social,wasserman1994social}, the webgraph~\citep{donato2004large,georgeot2010spectral}, biological systems~\citep{bonchev2007complexity,lesne2006complex,pavlopoulos2011using}, econometrics~\citep{hausmann2014atlas,tacchella2013economic}, and linguistics~\citep{vitevitch2008can}.
Here we demonstrate that the combination of graph and game theory provides useful tools for analyzing the structure of general-sum, many-player games.

The primary contribution of this work is a graph-based toolkit for analysis and comparison of games. 
As detailed below, the nodes in our graphs are either strategies (in abstract games) or AI agents (in empirical games, where strategies correspond to learned or appropriately-sampled player policies). 
The interactions between these agents, as quantified by the game's payoffs, constitute the structure of the graph under analysis. 
We show that this set of nodes and edges, also known as the \alpharank response graph~\citep{omidshafiei2019alpha,rowland2019multiagent,Muller2020A}, yields useful insights into the structure of individual games and can be used to generate a landscape over collections of games (as in \cref{fig:rwg_embeddings}).
We subsequently use the toolkit to analyze various games that are both played by humans or wherein AI agents have reached human-level performance, including Go, MuJoCo Soccer, and StarCraft~II.
Our overall analysis culminates in a demonstration of how the topological structure over games can be used to tackle the interestingness question of the Problem Problem, which seeks to automatically generate games with interesting characteristics for learning agents~\citep{leibo2019autocurricula}.

\section*{Results}\label{sec:results}
\subsection*{Overview}

We develop a foundational graph-theoretic toolkit that facilitates analysis of canonical and large-scale games, providing insights into their related topological structure in terms of their high-level strategic interactions.
The prerequisite game theory background and technical details are provided in the \nameref{sec:methods} section, with full discussion of related works and additional details in Supplementary Note 1.

Our results are summarized as follows.
We use our toolkit to characterize a number of games, first analyzing motivating examples and canonical games with well-defined structures, then extending to larger-scale empirical games datasets.
For these larger games, we rely on empirical game-theoretic analysis~\citep{wellman2006methods,Walsh02}, where we characterize an underlying game using a sample set of policies.
While the empirical game-theoretic results are subject to the policies used to generate them, we rely on a sampling scheme designed to capture a diverse variety of interactions within each game, and subsequently conduct sensitivity analysis to validate the robustness of the results.
We demonstrate correlation between the complexity of the graphs associated with games and the complexity of solving the game itself.
In Supplementary Note 2, we evaluate our proposed method against baseline approaches for taxonomization of 2 $\times$ 2 games~\citep{bruns2015names}.
We finally demonstrate how this toolkit can be used to automatically generate interesting game structures that can, for example, subsequently be used to train AI agents.

\subsection*{Motivating example}

\begin{figure}[t]
    \centering{
        \phantomsubcaption\label{fig:elo_game_variations_transitive}
    	\phantomsubcaption\label{fig:elo_game_variations_transitive_unpermuted}
    	\phantomsubcaption\label{fig:elo_game_variations_transitive_do_iterations}
    	\phantomsubcaption\label{fig:elo_game_variations_cyclical}
    	\phantomsubcaption\label{fig:elo_game_variations_cyclical_unpermuted}
    	\phantomsubcaption\label{fig:elo_game_variations_cyclical_do_iterations}
    	\phantomsubcaption\label{fig:elo_game_variations_random}
    	\phantomsubcaption\label{fig:elo_game_variations_random_unpermuted}
    	\phantomsubcaption\label{fig:elo_game_variations_random_do_iterations}
    }
    \includegraphics[width=\linewidth]{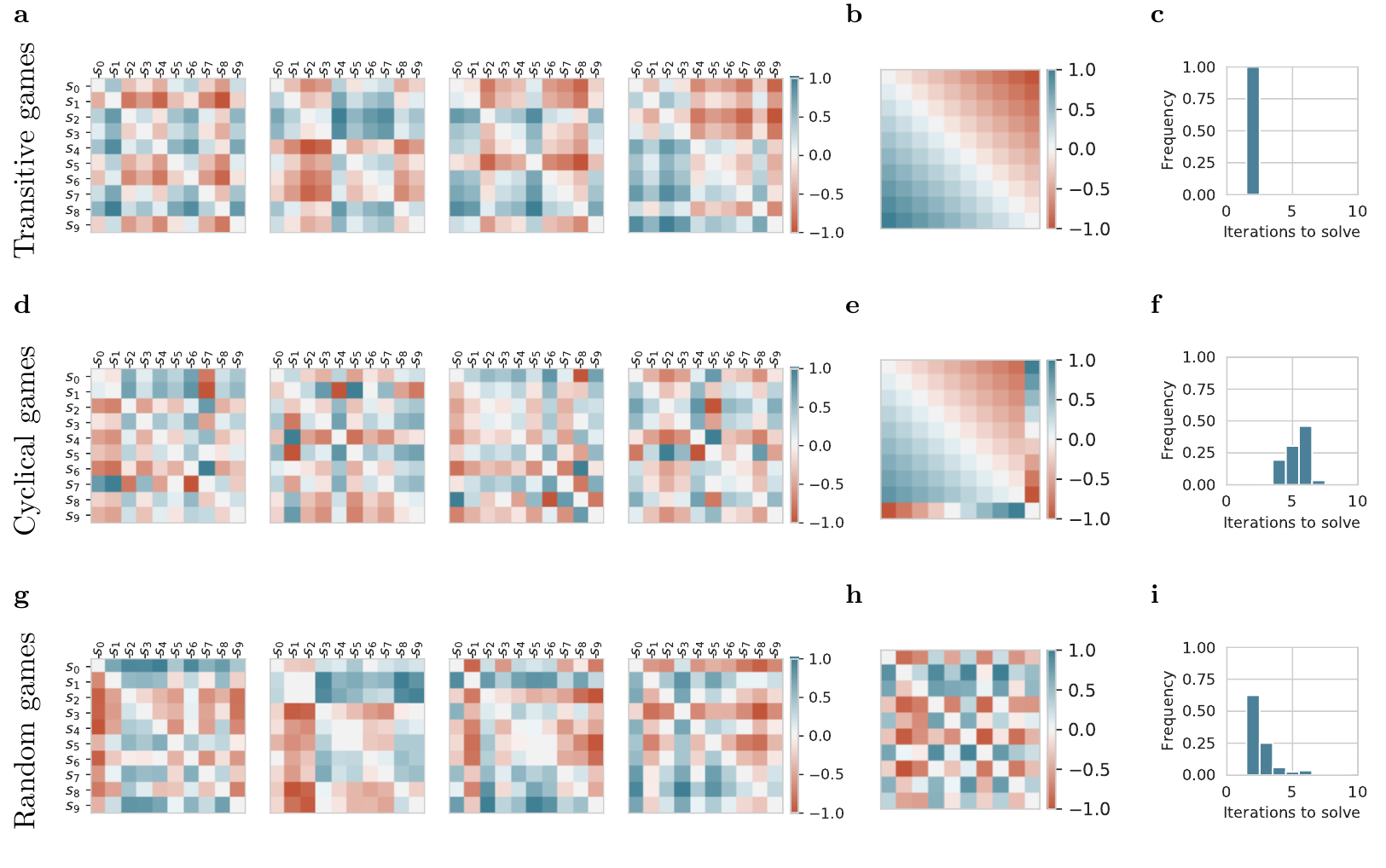}
	\caption{
    	Motivating example of three classes of two-player, symmetric zero-sum games. 
    	\subref{fig:elo_game_variations_transitive}, \subref{fig:elo_game_variations_cyclical}, and \subref{fig:elo_game_variations_random}, respectively, visualize payoffs for instances of games with transitive, cyclical, and random structure.
    	Each exemplified game consists of two players with 10 strategies each (with payoff row and column labels, $\{s_0, \ldots, s_9\}$, indicating the strategies).
		Despite the numerous payoff variations possible in each class of games illustrated, each shares the underlying payoff structure shown, respectively, in \subref{fig:elo_game_variations_transitive_unpermuted}, \subref{fig:elo_game_variations_cyclical_unpermuted}, and \subref{fig:elo_game_variations_random_unpermuted}.
		Moreover, variations in payoffs can notably impact the difficulty of solving (i.e., finding the Nash equilibrium) of these games, as visualized in \subref{fig:elo_game_variations_transitive_do_iterations}, \subref{fig:elo_game_variations_cyclical_do_iterations}, \subref{fig:elo_game_variations_random_do_iterations}.
	}
	\label{fig:elo_game_variations}
\end{figure}

Let us start with a motivating example to solidify intuitions and explain the workflow of our graph-theoretic toolkit, using classes of games with simple parametric structures in the player payoffs.
Specifically, consider games of three broad classes (generated as detailed in the Supplementary Methods 1): 
games in which strategies have a clear transitive ordering (\cref{fig:elo_game_variations_transitive}); 
games in which strategies have a cyclical structure wherein all but the final strategy are transitive with respect to one another (\cref{fig:elo_game_variations_cyclical}); 
and games with random (or no clear underlying) structure (\cref{fig:elo_game_variations_random}).
We shall see that the core characteristics of games with shared underlying structure is recovered via the proposed analysis.

Each of these figures visualizes the payoffs corresponding to 4 instances of games of the respective class, with each game involving 10 strategies per player;
more concretely, entry $\emM(s_i,s_j)$ of each matrix visualized in \Cref{fig:elo_game_variations_transitive,fig:elo_game_variations_cyclical,fig:elo_game_variations_random} quantifies the payoff received by the first player if the players, respectively, use strategies $s_i$ and $s_j$ (corresponding, respectively, to the $i$-th and $j$-th row and column of each payoff table).
Despite the variance in payoffs evident in the instances of games exemplified here, each essentially shares the payoff structure exposed by re-ordering their strategies, respectively, in \cref{fig:elo_game_variations_transitive_unpermuted,fig:elo_game_variations_cyclical_unpermuted,fig:elo_game_variations_random_unpermuted}.
In other words, the visual representation of the payoffs in this latter set of figures succinctly characterizes the backbone of strategic interactions within these classes of games, despite not being immediately apparent in the individual instances visualized.

More importantly, the complexity of learning useful mixed strategies to play in each of these games is closely associated with this structural backbone.
To exemplify this, consider the computational complexity of solving each of these games; for brevity, we henceforth refer to solving a game as synonymous with finding a Nash equilibrium (similar to prior works~\citep{bowling2015heads,waugh2015solving,burch2014solving,robinson1951iterative}, wherein the Nash equilibrium is the solution concept of interest). 
Specifically, we visualize this computational complexity by using the Double Oracle algorithm~\citep{mcmahan2003planning}, which has been well-established as a Nash solver in multiagent and game theory literature~\citep{bosansky2015combining,lanctot2017unified,jain2011double,regan2012regret}.
At a high level, Double Oracle starts from a sub-game consisting of a single randomly-selected strategy, iteratively expands the strategy space via best responses (computed by an oracle), until discovery of the Nash equilibrium of the full underlying game. 

\Cref{fig:elo_game_variations_transitive_do_iterations,fig:elo_game_variations_cyclical_do_iterations,fig:elo_game_variations_random_do_iterations} visualize the distribution of Double Oracle iterations needed to solve the corresponding games, under random initializations.
Note, in particular, that although the underlying payoff structure of the transitive and cyclical games respectively visualized in \cref{fig:elo_game_variations_transitive} and \cref{fig:elo_game_variations_cyclical} is similar, the introduction of a cycle in the latter class of games has a substantial impact on the complexity of solving them (as evident in \cref{fig:elo_game_variations_cyclical_do_iterations}). 
In particular, whereas the former class of games are solved using a low (and deterministic) number of iterations, the latter class requires additional iterations due to the presence of cycles increasing the number of strategies in the support of the Nash equilibrium. 

\begin{figure}[t]
    \centering
    \includegraphics[width=\linewidth]{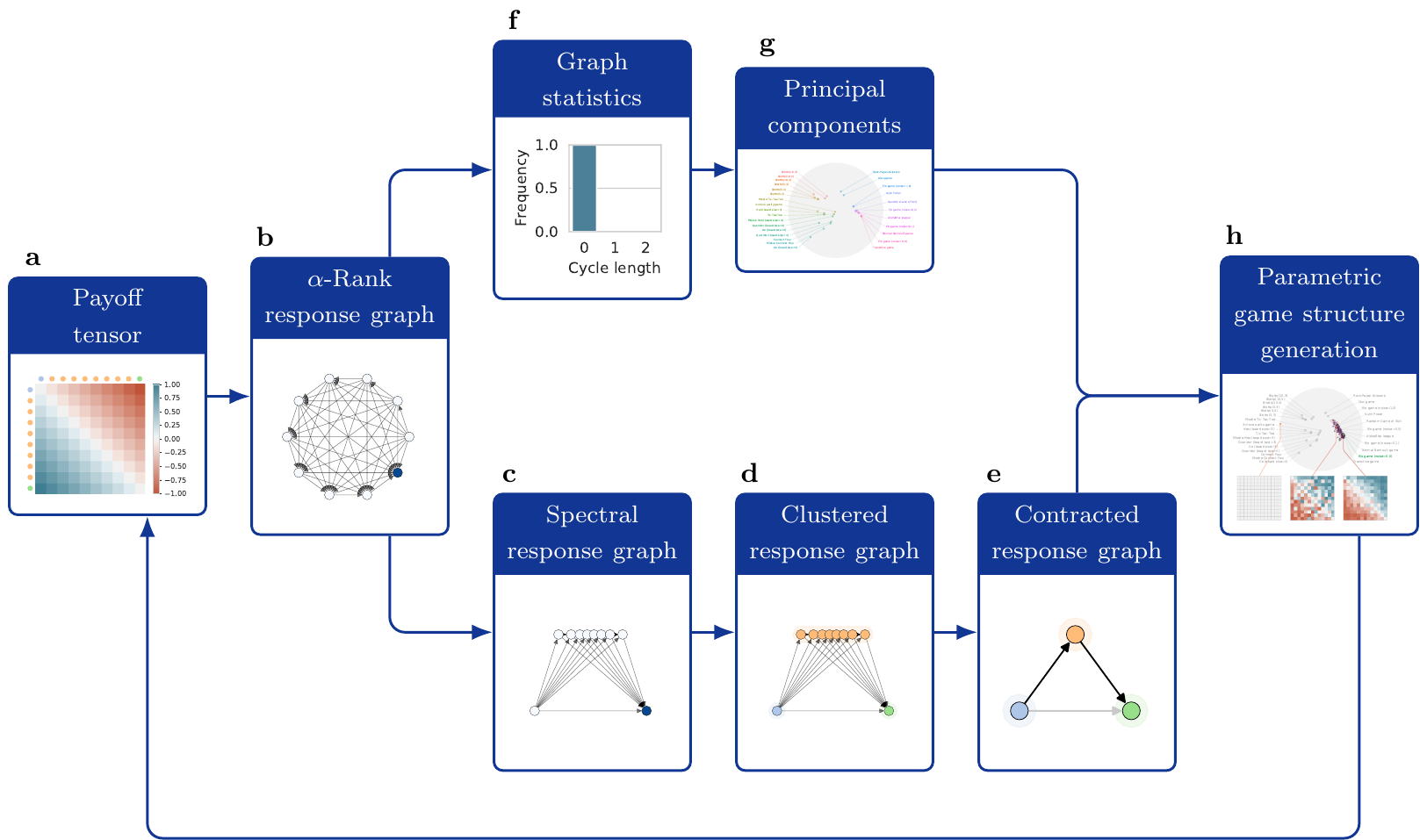}
    \caption{Method workflow, with  accompanying transitive game results.
    Given the game payoffs \textbf{a}, the so-called \alpharank response graph of the game is visualized in \textbf{b}.
    In \textbf{c}, reprojecting the response graph by using the top eigenvectors of the graph Laplacian yields the {spectral response graph}, wherein similar strategies are placed close to one another.
    In \textbf{d}, taking this one step further, one can cluster the spectral response graph, yielding the {clustered response graph}, which exposes three classes of strategies in this particular example.
    In \textbf{e}, contracting the clustered graph by fusing nodes within each cluster yields the high-level characterization of transitive games.
    In \textbf{f}, the lack of cycles in the particular class of transitive games becomes evident.
    Finally, in \textbf{g} and \textbf{h}, one can extract the principal components of various response graph statistics and establish a feedback loop to a procedural game structure generation scheme to yield new games. 
    }
    \label{fig:method_workflow}
\end{figure}

\begin{figure}[t]
    \centering{
        \phantomsubcaption\label{fig:cyclical_plot_payoffs}
        \phantomsubcaption\label{fig:cyclical_plot_networkx}
        \phantomsubcaption\label{fig:cyclical_plot_cycles_histogram}
        \phantomsubcaption\label{fig:cyclical_plot_networkx_spectral}
        \phantomsubcaption\label{fig:cyclical_plot_networkx_spectral_clustered}
        \phantomsubcaption\label{fig:cyclical_plot_networkx_spectral_clustered_contracted}
    }
    \includegraphics[width=\linewidth]{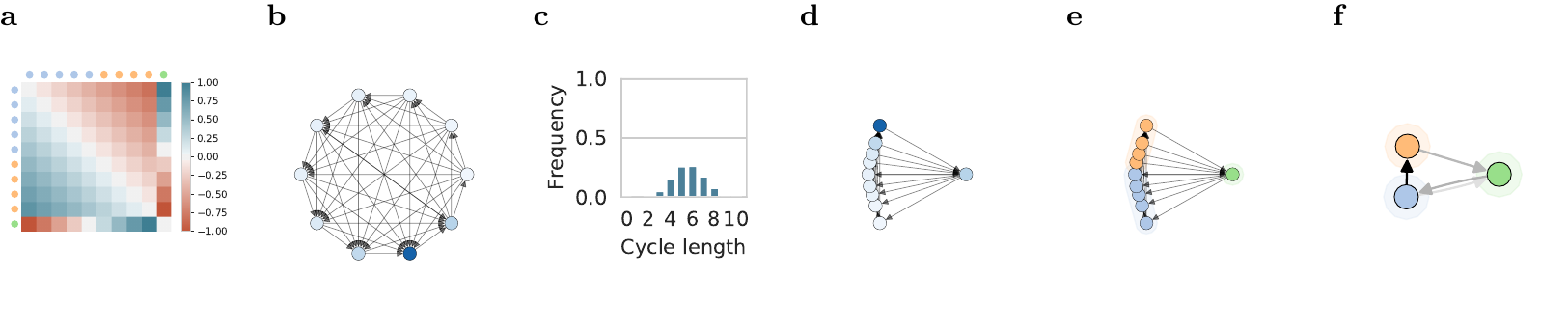}
    \caption{Cyclical game results. 
    \subref{fig:cyclical_plot_payoffs} game payoffs,
    \subref{fig:cyclical_plot_networkx} response graph,
    \subref{fig:cyclical_plot_cycles_histogram} cycles histogram,
    \subref{fig:cyclical_plot_networkx_spectral} spectral response graph,
    \subref{fig:cyclical_plot_networkx_spectral_clustered} clustered response graph,
    \subref{fig:cyclical_plot_networkx_spectral_clustered_contracted} contracted response graph.
    }
    \label{fig:cyclical_complexity_overview}
\end{figure}

\subsection*{Workflow} 
Overall, the characterization of the topological structure of games is an important and nuanced problem.
To address this problem, we use graph theory to build an analytical toolkit automatically summarizing the high-level strategic interactions within a game, and providing useful complexity measures thereof.
Specifically, consider again our motivating transitive game, re-visualized using a collection of graph-based measures in \cref{fig:method_workflow}.
Each of these measures provides a different viewpoint on the underlying game, collectively characterizing it.
Specifically, given the game payoffs, \cref{fig:method_workflow}b visualizes the so-called \alpharank response graph of the game;
here, each node corresponds to a strategy (for either player, as this particular game's payoffs are symmetric).
Transition probabilities between nodes are informed by a precise evolutionary model (detailed in \nameref{sec:methods} and \citet{omidshafiei2019alpha});
roughly speaking, a directed edge from one strategy to another indicates the players having a higher preference for the latter strategy, in comparison to the former.
The response graph, thus, visualizes all preferential interactions between strategies in the game.
Moreover, the color intensity of each node indicates its so-called \alpharank, which measures the long-term preference of the players for that particular strategy, as dictated by the transition model mentioned above;
specifically, darker colors here indicate more preferable strategies. 

This representation of a game as a graph enables a variety of useful insights into its underlying structure and complexity. 
For instance, consider the distribution of cycles in the graph, which play an important role in multiagent evaluation and training schemes~\citep{singh2000nash,tuyls2018generalised,balduzzi2018re,omidshafiei2019alpha} and, as later shown, are correlated to the computational complexity of solving two-player zero-sum games (e.g., via Double Oracle).
\Cref{fig:method_workflow}f makes evident the lack of cycles in the particular class of transitive games;
while this is clearly apparent in the underlying (fully ordered) payoff visualization of \cref{fig:method_workflow}a, it is less so in the unordered variants visualized in \cref{fig:elo_game_variations_transitive}.
Even so, the high-level relational structure between the strategies becomes more evident by conducting a spectral analysis of the underlying game response graph.
Full technical details of this procedure are provided in the \nameref{sec:methods} section.
At a high level, the so-called Laplacian spectrum (i.e., eigenvalues) of a graph, along with associated eigenvectors, captures important information regarding it (e.g., number of spanning trees, algebraic connectivity, and numerous related properties~\citep{mohar1991laplacian}).
Reprojecting the response graph by using the top eigenvectors yields the {spectral response graph} visualized in \cref{fig:method_workflow}c, wherein similar strategies are placed close to one another.
Moreover, one can cluster the spectral response graph, yielding the {clustered response graph}, which exposes three classes of strategies in \cref{fig:method_workflow}d: 
a fully dominated strategy with only outgoing edges (a singleton cluster, on the bottom left of the graph), a transient cluster of strategies with both incoming and outgoing edges (top cluster), and a dominant strategy with all incoming edges (bottom right cluster).
Finally, contracting the clustered graph by fusing nodes within each cluster yields the high-level characterization of transitive games shown in \cref{fig:method_workflow}e.

We can also conduct this analysis for instances of our other motivating games, such as the cyclical game visualized in \cref{fig:cyclical_plot_payoffs}.
Note here the distinct differences with the earlier transitive game example;
in the cyclical game, the \alpharank distribution in the response graph (\cref{fig:cyclical_plot_networkx}) has higher entropy (indicating preference for many strategies, rather than one, due to the presence of cycles). 
Moreover, the spectral reprojection in \cref{fig:cyclical_plot_networkx_spectral} reveals a clear set of transitive nodes (left side of visualization) and a singleton cluster of a cycle-inducing node (right side).
Contracting this response graph reveals the fundamentally cyclical nature of this game (\cref{fig:cyclical_plot_networkx_spectral_clustered_contracted}).
Finally, we label each strategy (i.e., each row and column) of the original payoff table \cref{fig:cyclical_plot_payoffs} based on this clustering analysis.
Specifically, the color-coded labels on the far left (respectively, top) of each row (respectively, column) in \cref{fig:cyclical_plot_payoffs} correspond to the clustered strategy colors in  \cref{fig:cyclical_plot_networkx_spectral}.
This color-coding helps clearly identify the final strategy (i.e., bottom row of the payoff table) as the outlier enforcing the cyclical relationships in the game.
Note that while there is no single graphical structure that summarizes the particular class of random games visualized earlier in \cref{fig:elo_game_variations_random_unpermuted}, we include this analysis for several instances of such games in Supplementary Note 2.

Crucially, a key benefit of this analysis is that the game structure exposed is identical for all instances of the transitive and cyclical games  visualized earlier in \cref{fig:elo_game_variations_transitive,fig:elo_game_variations_cyclical}, making it significantly easier to characterize games with related structure, in contrast to analysis of raw payoffs.
Our later case studies further exemplify this, exposing related underlying structures for several classes of more complex games.

\subsection*{Analysis of Canonical and Real-World Games}
The insights afforded by our graph-theoretic approach apply to both small canonical games and larger empirical games (where strategies are synonymous with trained AI agents).

\begin{figure}[t!]
    \gameRowPhantomSubcaptions{redundant_rock_paper_scissors}
    \gameRowPhantomSubcaptions{11_20}
    \includegraphics[width=\linewidth]{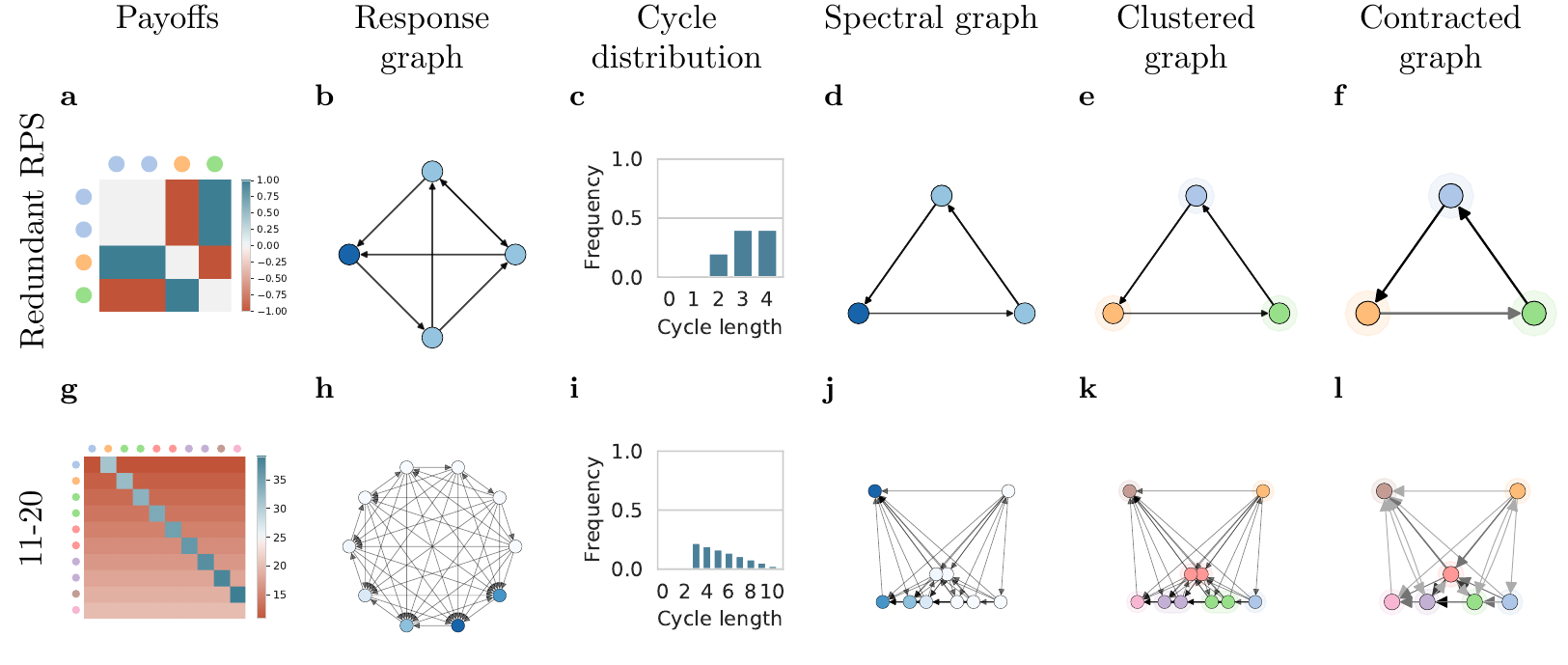}
    \caption{Results for Redundant Rock--Paper--Scissors (RPS) and 11-20 game. In Redundant RPS, the redundant copy of the first strategy (Rock) is clustered in the spectral response graph.
    In 11-20, seven clusters of strategies are revealed, exposing the cyclical nature of this game.}
    \label{fig:redundant_rps_and_11_20_complexity_overview}
\end{figure}

\begin{figure}[t!]
    \gameRowPhantomSubcaptions{alphago}
    \gameRowPhantomSubcaptions{soccer}
    \includegraphics[width=\linewidth]{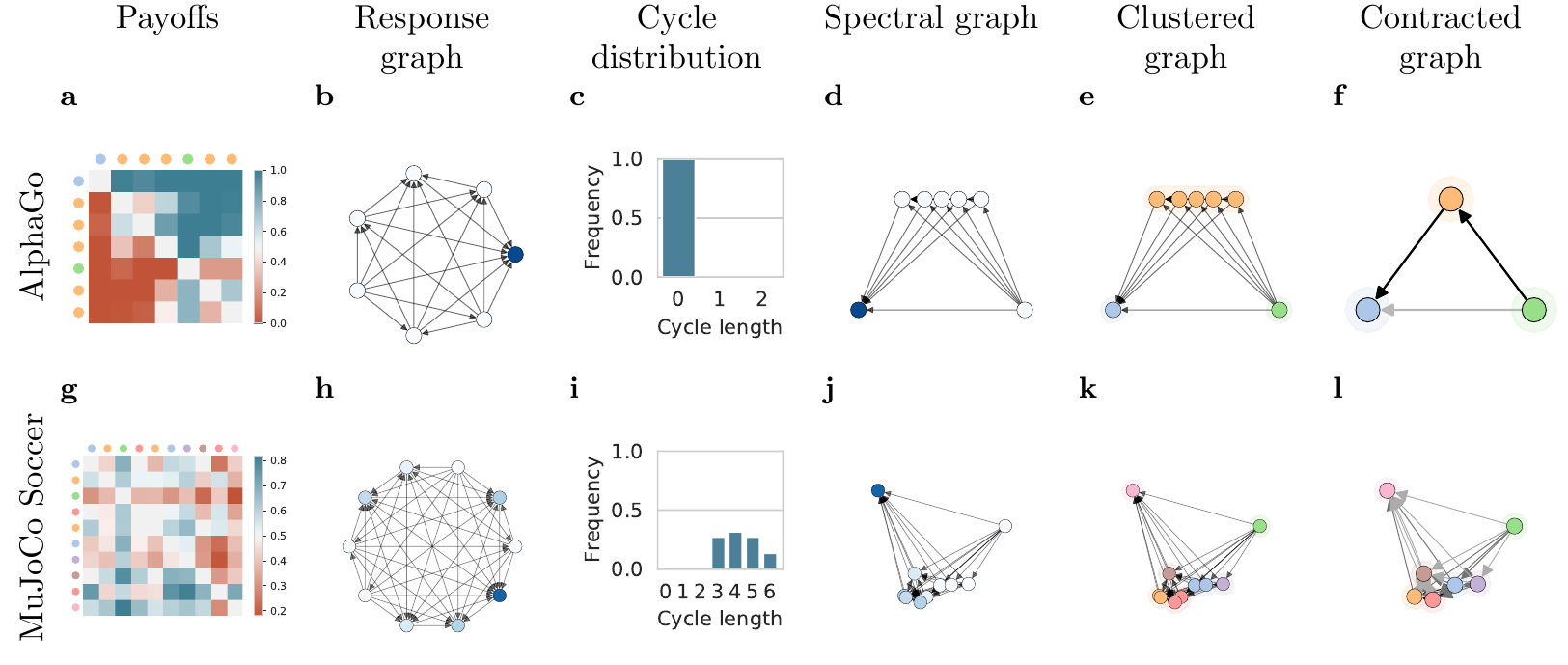}
    \caption{Results for empirical games of AlphaGo and MuJoCo soccer dataset.
    Note that as these are empirical games, strategies here correspond to trained AI agents.
    In AlphaGo, the strong transitive relationship between agents is revealed via our analysis.
    In MuJoCo soccer, more complex relations between similarly-performing agents are revealed in the clusters produced.
    }
    \label{fig:alphago_soccer_complexity_overview}
\end{figure}

Consider the canonical Rock--Paper--Scissors game, involving a cycle among the three strategies (wherein Rock loses to Paper, which loses to Scissors, which loses to Rock).
\cref{fig:redundant_rock_paper_scissors_plot_payoffs} visualizes a variant of this game involving a copy of the first strategy, Rock, which introduces a redundant cycle and thus affects the distribution of cycles in the game.
Despite this, the spectral response graph (\cref{fig:redundant_rock_paper_scissors_plot_networkx_spectral}) reveals that the redundant game topologically remains the same as the original Rock--Paper--Scissors game, thus reducing to the original game under spectral clustering.

This graph-based analysis also extends to general-sum games.
As an example, consider the slightly more complex game of 11-20, wherein two players each request an integer amount of money between 11 to 20 units (inclusive).
Each player receives the amount requested, though a bonus of 20 units is allotted to one player if they request exactly 1 unit less than the other player.
The payoffs and response graph of this game are visualized, respectively, in \cref{fig:11_20_plot_payoffs,fig:11_20_plot_networkx}, where strategies, from top-to-bottom and left-to-right in the payoff table, correspond to increasing units of money being requested.
This game, first introduced by \citet{arad201211}, is structurally designed to analyze so-called $k$-level reasoning, wherein a level-$0$ player is naive (i.e., here simply requests $20$ units), and any level-$k$ player responds to an assumed level-($k-1$) opponent;
e.g., here a level-$1$ player best responds to an assumed level-$0$ opponent, thus requesting 19 units to ensure receiving the bonus units.

The spectral response graph here (\cref{fig:11_20_plot_networkx_spectral}) indicates a more complex mix of transitive and intransitive relations between strategies.
Notably, the contracted response graph (\cref{fig:11_20_plot_networkx_spectral_clustered_contracted}) reveals 7 clusters of strategies.
Referring back to the rows of payoffs in \cref{fig:11_20_plot_payoffs}, relabeled to match cluster colors, demonstrates that our technique effectively pinpoints the sets of strategies that define the rules of the game:
weak strategies (11 or 12 units, first two rows of the payoff table, and evident in the far-right of the clustered response graph), followed by a set of intermediate strategies with higher payoffs (clustered pairwise, near the lower-center of the clustered response graph), and finally the two key strategies that establish the cyclical relationship within the game through $k$-level reasoning (19 and 20 units, corresponding to level-$0$ and level-$1$ players, in the far-left of the clustered response graph).

\begin{figure}[t]
    \gameRowFiletypePhantomSubcaptions{AlphaStar}
    \gameRowFiletypePhantomSubcaptions{AlphaStarTopK}
    \includegraphics[width=\linewidth]{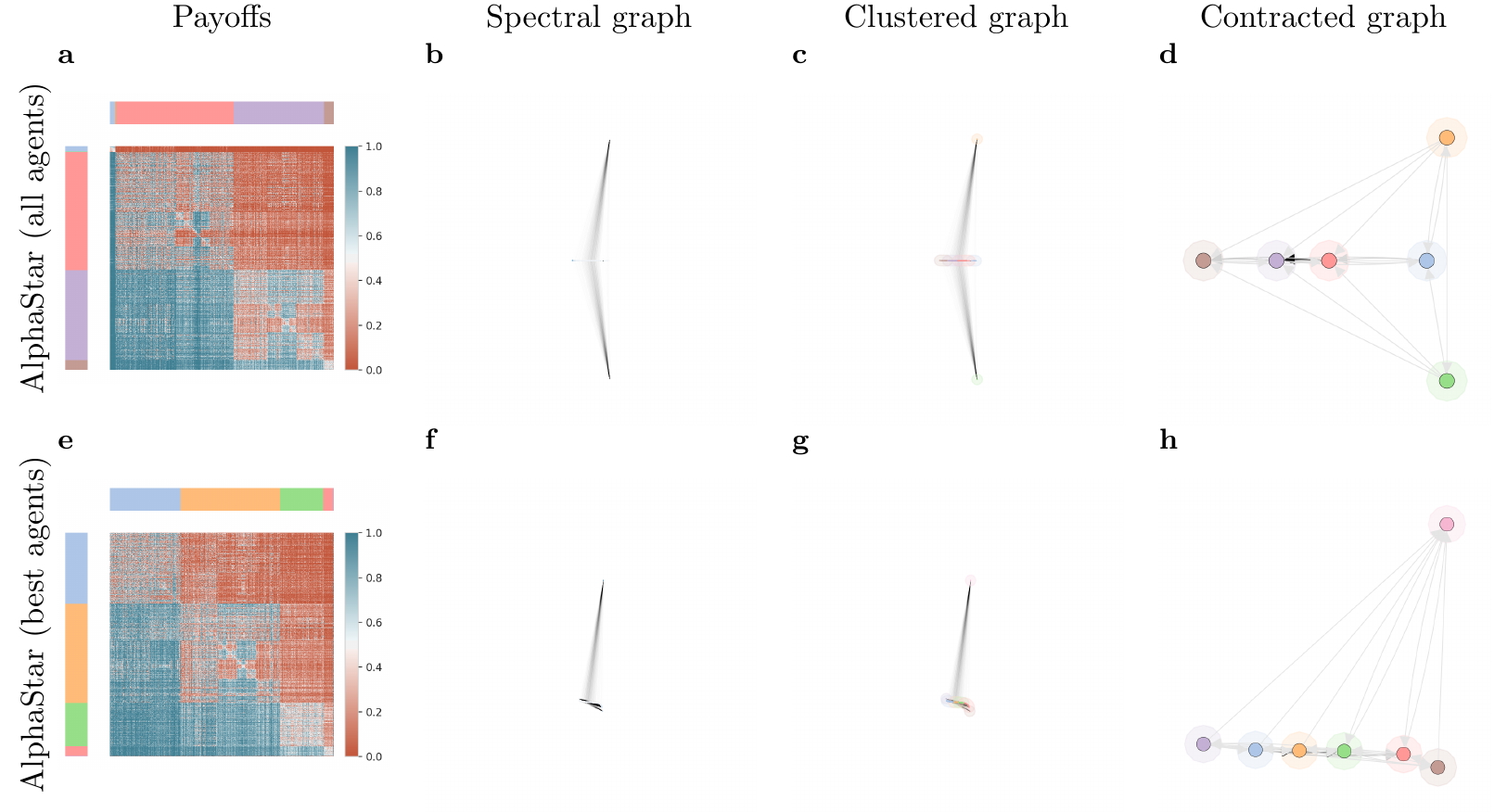}
    \caption{AlphaStar results, with both the full league and the league with best agents visualized.
    Spectral analysis of the empirical AlphaStar League game reveals that several key subsets of closely-performing agents, illustrated in \subref{fig:AlphaStar_plot_networkx_spectral_clustered_contracted}.
    Closer inspection of the agents used to construct this empirical payoff table reveals the following insights, with agent types corresponding to those detailed in~\citet{vinyals19}: 
    i) the blue, orange, and green clusters are composed of agents in the initial phases of training, which are generally weakest (as observed in \subref{fig:AlphaStar_plot_networkx_spectral_clustered_contracted}, and also visible as the narrow band of low payoffs in the top region of the payoff table \subref{fig:AlphaStar_plot_payoffs}); 
    ii) the red cluster consists primarily of various, specialized exploiter agents; 
    iii) the purple and brown clusters are primarily composed of the league exploiters and main agents, with the latter being generally higher strength than the former.
    }
    \label{fig:alphastar_results}
\end{figure}

This analysis extends to more complex instances of {empirical games}, which involve trained AI agents, as next exemplified.
Consider first the game of Go, as played by 7 AlphaGo variants: $AG(r)$, $AG(p)$, $AG(v)$, $AG(rv)$, $AG(rp)$, $AG(vp)$, and $AG(rvp)$, where each variant uses the specified combination of rollouts $r$, value networks $v$, and/or policy networks $p$. 
We analyze the empirical game where each strategy corresponds to one of these agents, and payoffs (\cref{fig:alphago_plot_payoffs}) correspond to the win rates of these agents when paired against each other (as detailed by \citet[Table 9]{Silver2016}).
The \alpharank distribution indicated by the node (i.e., strategy) color intensities in \cref{fig:alphago_plot_networkx} reveals $AG(rvp)$ as a dominant strategy, and the cycle distribution graph \cref{fig:alphago_plot_cycles_histogram} reveals a lack of cycles here.
The spectral response graph, however, goes further, revealing a fully transitive structure (\cref{fig:alphago_plot_networkx_spectral,fig:alphago_plot_networkx_spectral_clustered}), as in the motivating transitive games discussed earlier.
The spectral analysis on this particular empirical game, therefore, reveals its simple underlying transitive structure (\cref{fig:alphago_plot_networkx_spectral_clustered_contracted}). 

\begin{figure}[t]   
    \centering
    \gameRowFiletypePhantomSubcaptions{5_3_Blotto}
    \gameRowFiletypePhantomSubcaptions{10_3_Blotto}
    \gameRowFiletypePhantomSubcaptions{go_board_size_3_komi_6_5_}
    \includegraphics[width=\linewidth]{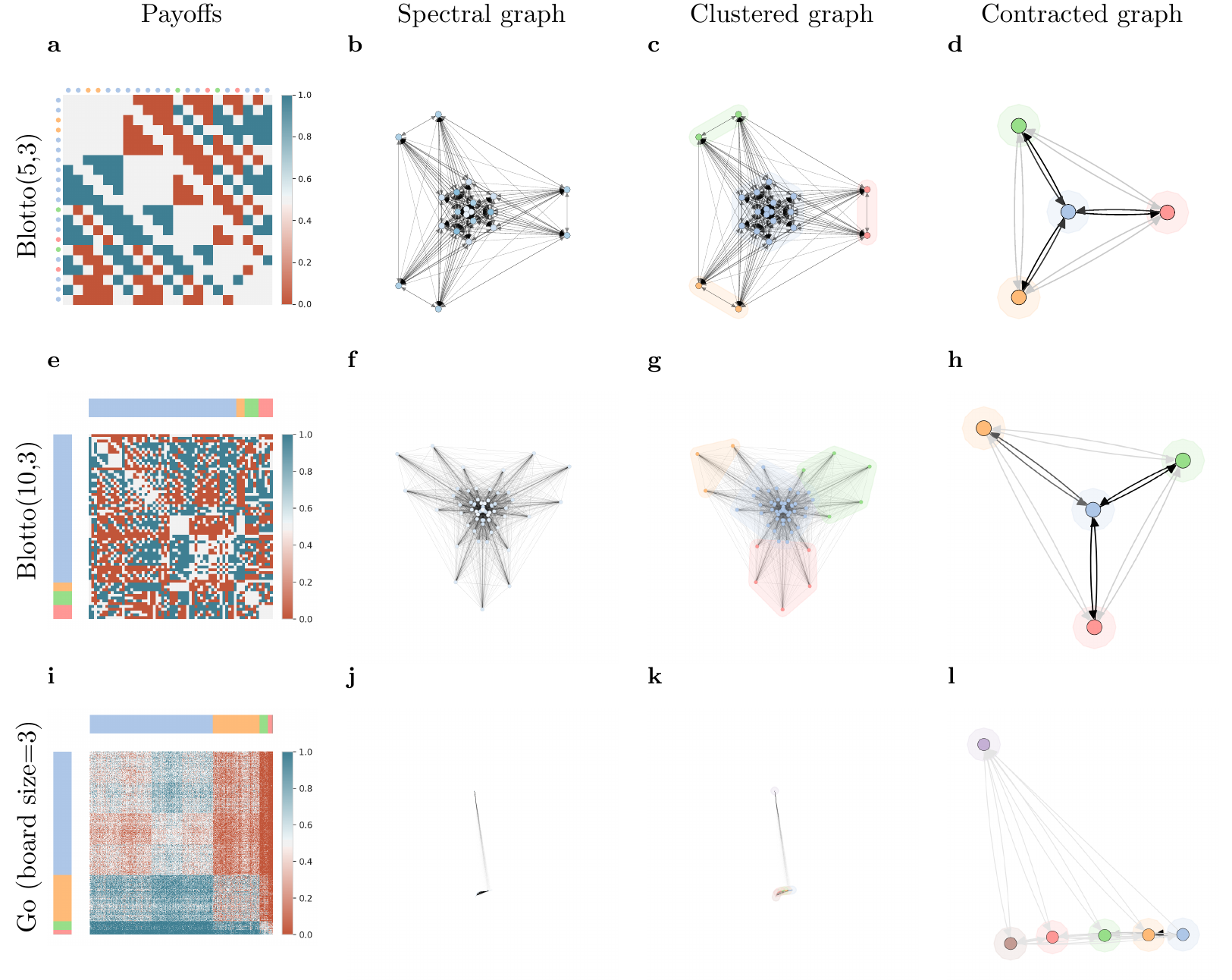}
    \caption{Results for Blotto(5,3), Blotto(10,3), and Go (board size=3). 
    Despite the significant difference in sizes, both instances of Blotto yield a remarkably similar contracted response graph.
    Moreover, the contracted response graph for Go is notably different from AlphaGo results, due to the latter being an empirical game constructed from trained AI agents rather than a representative set of sampled policies.
    }
    \label{fig:blotto_go_complexity_overview}
\end{figure}

Consider a more interesting empirical game, wherein agents are trained to play soccer in the continuous control domain of MuJoCo, exemplified in \cref{fig:alphago_soccer_complexity_overview} (second row).
Each agent in this empirical game is generated using a distinct set of training parameters (e.g., feedforward vs. recurrent policies, reward shaping enabled and disabled, etc.), with full agent specifications and payoffs detailed by \citet{liu2018emergent}.
The spectral response graph (\cref{fig:soccer_plot_networkx_spectral_clustered}) reveals two outlier agents: a strictly dominated agent (node in the top-right), and a strong (yet not strictly dominant) agent (node in the top-left). 
Several agents here are clustered pairwise, revealing their closely-related interactions with respect to the other agents;
such information could, for example, be used to discard or fuse such redundant agents during training to save computational costs.

Consider next a significantly larger-scale empirical game, consisting of 888 StarCraft~II agents from the AlphaStar Final league of \citet{vinyals19}.
StarCraft~II is a notable example, involving a choice of 3 races per player and realtime gameplay, making a wide array of behaviors possible in the game itself.
The empirical game considered is visualized in \cref{fig:AlphaStar_plot_payoffs}, and is representative of a large number of agents with varying skill levels.
Despite its size, spectral analysis of this empirical game reveals that several key subsets of closely-performing agents exist here, illustrated in \cref{fig:AlphaStar_plot_networkx_spectral_clustered_contracted}.
Closer inspection of the agents used to construct this empirical payoff table reveals the following insights, with agent types corresponding to those detailed in~\citet{vinyals19}: 
i) the blue, orange, and green clusters are composed of agents in the initial phases of training, which are generally weakest (as observed in \cref{fig:AlphaStar_plot_networkx_spectral_clustered_contracted}, and also visible as the narrow band of low payoffs in the top of \cref{fig:AlphaStar_plot_payoffs}); 
ii) the red cluster consists primarily of various, specialized exploiter agents; 
iii) the purple and brown clusters are primarily composed of the league exploiters and main agents, with the latter being generally higher strength than the former.
To further ascertain the relationships between only the strongest agents, we remove the three clusters corresponding to the weakest agents, repeating the analysis in 
\cref{fig:alphastar_results} (bottom row).
Here, we observe the presence of a series of progressively stronger agents (bottom nodes in \cref{fig:AlphaStarTopK_plot_networkx_spectral_clustered_contracted}), as well as a single outlier agent which quite clearly bests several of these clusters (top node of \cref{fig:AlphaStarTopK_plot_networkx_spectral_clustered_contracted}).

An important caveat, as this stage, is that the agents in AlphaGo, MuJoCo soccer, and AlphaStar above were trained to maximize performance, rather than to explicitly reveal insights into their respective {underlying} games of Go, soccer, and StarCraft~II. 
Thus, this analysis focused on characterizing relationships between the agents from the Policy Problem perspective, rather than the underlying games themselves, which provide insights into the interestingness of the game (Problem Problem).
This latter investigation would require a significantly larger population of agents, which cover the policy space of the underlying game effectively, as exemplified next.

Naturally, characterization of the underlying game can be achieved in games small enough where all possible policies can be explicitly compared against one another.
For instance, consider Blotto($\tau,\rho$), a zero-sum two-player game wherein each player has $\tau$ tokens that they can distribute amongst $\rho$ regions~\citep{borel1921theorie}.
In each region, each player with the most tokens wins (see \citet{tuyls2018generalised} for additional details).
In the variant we analyze here, each player receives a payoff of $+1$, $0$, and $-1$ per region respectively won, drawn, and lost.
The permutations of each player's allocated tokens, in turn, induce strong cyclical relations between the possible policies in the game. 
While the strategy space for this game is of size $\binom{\tau+\rho-1}{\rho-1}$, payoffs matrices can be fully specified for small instances, as shown for Blotto(5,3) and Blotto(10,3) in \cref{fig:blotto_go_complexity_overview} (first and second row, respectively).
Despite the differences in strategy space sizes in these particular instances of Blotto, the contracted response graphs in \cref{fig:5_3_Blotto_plot_networkx_spectral_clustered_contracted,fig:10_3_Blotto_plot_networkx_spectral_clustered_contracted} capture the cyclical relations underlying both instances, revealing a remarkably similar structure.

For larger games, the cardinality of the pure policy space typically makes it infeasible to fully enumerate policies and construct a complete empirical payoff table, despite the pure policy space being finite in size. 
For example, even in games such as Tic--Tac--Toe, while the number of unique board configurations is reasonably small (9! = 362880), the number of pure, behaviorally unique policies is enormous ($\approx 10^{567}$, see \citet[Section J]{czarnecki2020} for details).
Thus, coming up with a principled definition of a scheme for sampling a relevant set of policies summarizing the strategic interactions possible within large games remains an important open problem.
In these instances, we rely on sampling policies in a manner that captures a set of representative policies, i.e., a set of policies of varying skill levels, which approximately capture variations of strategic interactions possible in the underlying game.
The policy sampling approach we use is motivated with the above discussions and open question in mind, in that it samples a set of policies with varying skill levels, leading to a diverse set of potential transitive and intransitive interactions between them.

Specifically, we use the policy sampling procedure proposed by \citet{czarnecki2020}, which also seeks a set of representative policies for a given game.
The details of this procedure are provided in Supplementary Methods 1, and at a high level involve three phases:
i) using a combination of tree search algorithms, Alpha-Beta~\citep{newell1976computer} and Monte Carlo Tree Search~\citep{coulom2006efficient}, with varying tree depth limits for the former and varying number of simulations allotted to the latter, thus yielding policies of varying transitive strengths;
ii) using a range of random seeds in each instantiation of the above algorithms, thus producing a range of policies for each level of transitive strength;
iii) repeating the same procedure with negated game payoffs, thus also covering the space of policies that actively seek to lose the original game.
While this sampling procedure is a heuristic, it produces a representative set of policies with varying degrees of transitive and intransitive relations, and thus provides an approximation of the underlying game that can be feasibly analyzed.

Let us revisit the example of Go, constructing our empirical game using the above policy sampling scheme, rather than the AlphaGo agents used earlier.
We analyze a variant of the game with board size $3 \times 3$, as shown in \cref{fig:blotto_go_complexity_overview} (third row).
Notably, the contracted response graph (\cref{fig:go_board_size_3_komi_6_5__plot_networkx_spectral_clustered_contracted}) reveals the presence of a strongly-cyclical structure in the underlying game, in contrast to the AlphaGo empirical game (\cref{fig:alphago_soccer_complexity_overview}).
Moreover, the presence of a reasonably strong agent (visible in the top of the contracted response graph) becomes evident here, though this agent also shares cyclical relations with several sets of other agents.
Overall, this analysis exemplifies the distinction between analyzing an underlying game (e.g., Go) versus analyzing the agent training process (e.g., AlphaGo).
Investigation of links between these two lines of analysis, we believe, makes for an interesting avenue for future work.

\begin{figure}[t]
    \centering{
        \phantomsubcaption\label{fig:double_oracle_game_results_alpharank_entropy}
        \phantomsubcaption\label{fig:double_oracle_game_results_num_3cycles}
        \phantomsubcaption\label{fig:double_oracle_game_results_indegree_stats_mean}
    }
    \includegraphics[width=\linewidth]{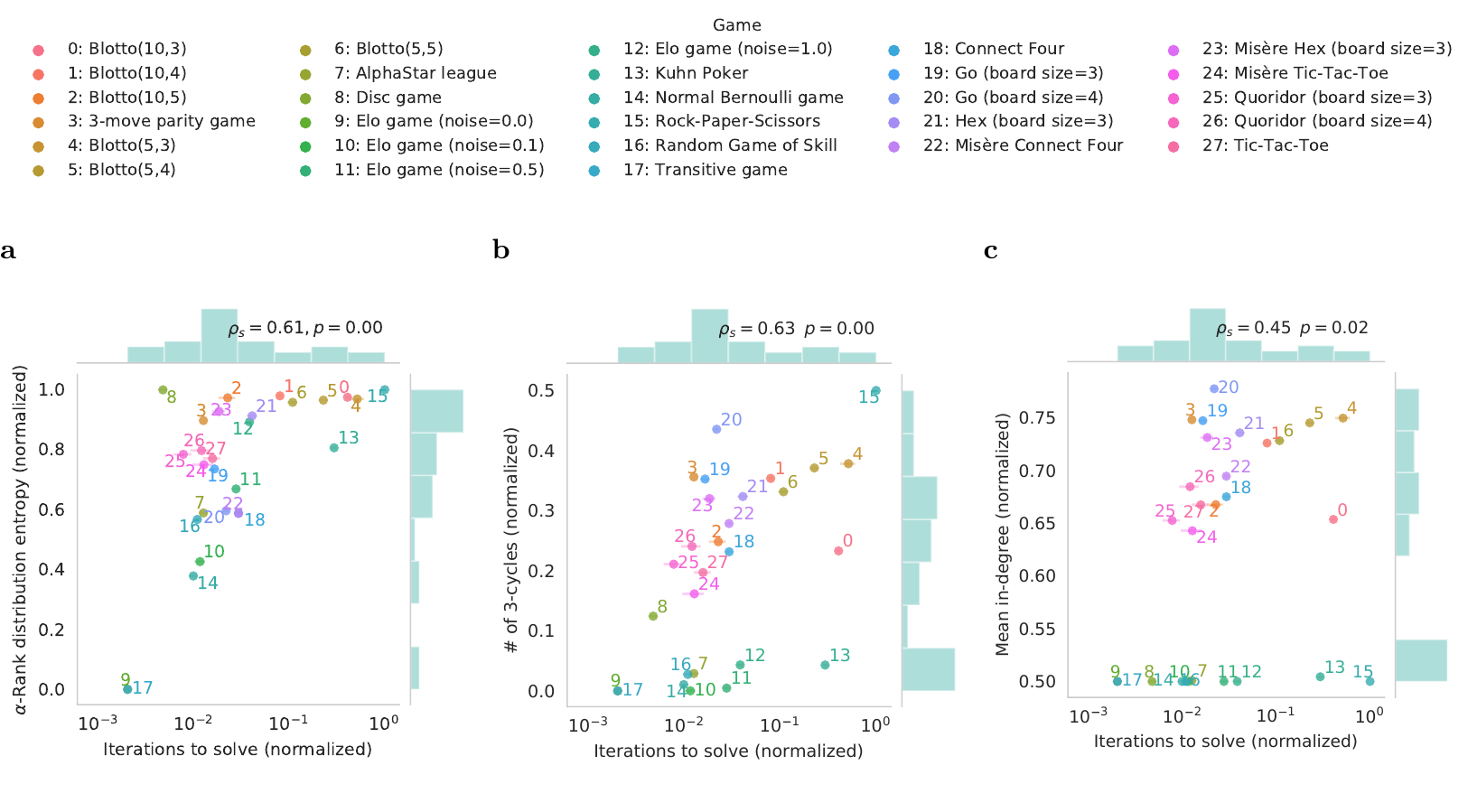}
    \caption{Response graph complexity vs. computational complexity of solving associated games. Each figure plots a respective measure of graph complexity against the normalized number of iterations needed to solve the associated game via the Double Oracle algorithm (with normalization done with respect to the total number of strategies in each underlying game). The Spearman correlation coefficient, $\rho_s$, is shown in each figure (with the reported two-sided p-value rounded to two decimals).
    }
    \label{fig:graph_vs_computational_complexity}
\end{figure}

\subsection*{Linking response graph and computational complexity}\label{sec:linking_response_graph_complexity}
A question that naturally arises is whether certain measures over response graphs are correlated with the computational complexity of solving their associated games.
We investigate these potential correlations here, while noting that these results are not intended to propose that a specific definition of computational complexity (e.g., with respect to Nash) is explicitly useful for defining a topology / classification over games.
In \cref{fig:graph_vs_computational_complexity}, we compare several response graph complexity measures against the number of iterations needed to solve a large collection of games using the Double Oracle algorithm~\citep{mcmahan2003planning}.
The results here consider specifically the \alpharank entropy, number of 3-cycles, and mean in-degree (with details in \nameref{sec:methods}, and results for additional measures included in Supplementary Note 2).
As in earlier experiments, solution of small-scale games is computed using payoffs over full enumeration of pure policies, whereas that of larger games is done using the empirical games over sampled policies.
Each graph complexity measure reported is normalized with respect to the maximum measure possible in a graph of the same size, and the number of iterations to solve is normalized with respect to the number of strategies in the respective game.
Thus, for each game, the normalized number of iterations to solve provides a measure of its relative computational complexity compared to games with the same strategy space size (for completeness, we include experiments testing the effects of normalization on these results in Supplementary Note 2).

Several trends can be observed in these results.
First, the entropy of the \alpharank distribution associated with each game correlates well with its computational complexity (see Spearman's correlation coefficient $\rho_s$ in the top-right of \cref{fig:double_oracle_game_results_alpharank_entropy}).
This matches intuition, as higher entropy \alpharank distributions indicate a larger support over the strategy space (i.e., strong strategies, with non-zero \alpharank mass), thus requiring additional iterations to solve.
Moreover, the number of 3-cycles in the response graph also correlates well with computational complexity, again matching intuition as the intransitivities introduced by cycles typically make it more difficult to traverse the strategy space~\citep{balduzzi2019open}.
Finally, the mean in-degree over all response graph nodes correlates less so with computational complexity (though degree-based measures still serve a useful role in characterizing and distinguishing graphs of differing sizes~\citep{berlingerio2012netsimile}).
Overall, these results indicate that response graph complexity provides a useful means of quantifying the computational complexity of games.

\subsection*{The Landscape of Games}
The results, thus far, have demonstrated that graph-theoretic analysis can simplify games (via spectral clustering), uncover their topological  structure (e.g., transitive structure of the AlphaGo empirical game), and yield measures correlated to the computational complexity of solving these games.
Overall, it is evident that the perspective offered by graph theory yields a useful characterization of games across multiple fronts.
Given this insight, we next consider whether this characterization can be used to compare a widely-diverse set of games.

To achieve this, we construct empirical payoff tables for a suite of games, using the policy sampling scheme described earlier for the larger instances (also see Supplementary Methods 1 for full details, including description of the games considered and analysis of the sensitivity of these results to the choice of empirical policies and mixtures thereof).
For each game, we compute the response graphs and several associated local and global complexity measures (e.g., \alpharank distribution entropy, number of 3-cycles, node-wise in- and out-degree statistics, and several other measures detailed in the \nameref{sec:methods} section), which constitute a feature vector capturing properties of interest.
Finally, a principal component analysis of these features yields the low-dimensional visualization of the landscape of games considered, shown in \cref{fig:rwg_embeddings}.

We make several key insights given this empirical landscape of games.
Notably, variations of games with related rules are well-clustered together, indicating strong similarity despite the widely-varying sizes of their policy spaces and empirical games used to construct them;
specifically, all considered instances of Blotto cluster together, with empirical game sizes ranging from $20 \times 20$ for Blotto(5,3) to $1000 \times 1000$ for Blotto(10,5).
Moreover, games with strong transitive components (e.g., variations of Elo games, AlphaStar League, Random Game of Skill, and Normal Bernoulli Game) are also notably separated from strongly cyclical games (Rock--Paper--Scissors, Disc game, and Blotto variations).
Closely-related real-world games (i.e., games often played by humans in the real world, such as Hex, Tic-Tac-Toe, Connect Four, and each of their respective Mis\`ere counterparts wherein players seek to lose) are also well-clustered.
Crucially, the strong alignment of this analysis with intuitions of the similarity of certain classes of games serves as an important validation of the graph-based analysis technique proposed in this work. 
Additionally, the analysis and corresponding landscape of games make clear that several games of interest for AI seem well-clustered together, which also holds for less interesting games (e.g., Transitive and Elo games). 

We note that the overall idea of generating such a landscape over games ties closely with prior works on taxonomization of multiplayer games~\citep{robinson2005topology,bruns2015names}.
Moreover, 2D visualization of the expressitivity (i.e., style and diversity) and the overall space of of procedurally-generated games features have been also investigated in closely related work~\citep{smith2010analyzing,shaker2013automatic}.  
A recent line of related inquiry also investigates the automatic identification, and subsequent visualization of core mechanics in single-player games~\citep{charity2020mech}.
Overall, we believe this type of investigation can be considered a method to taxonomize which future multiplayer games may be interesting, and which ones less so to train AI agents on.

While the primary focus of this paper is to establish a means of topologically studying games (and their similarities), a natural artifact of such a methodology is that it can enable investigation of interesting and non-interesting classes of games.
There exist numerous perspectives on what may make a game interesting, which varies as a function of the field of study or problem being solved.
These include (overlapping) paradigms that consider interestingness from: 
a human-centric perspective (e.g., level of social interactivity, cognitive learning and problem solving, enjoyment, adrenaline, inherent challenge, aesthetics, story-telling in the game, etc.) \citep{vygotsky1978interaction,deterding2015lens,koster2013theory,lazzaro2009we,prensky2001fun,wang2011game};
a curriculum learning perspective (e.g., games or tasks that provide enough learning signal to the human or artificial learner) \citep{baker2019emergent};
a procedural content generation or optimization perspective (in some instances with a focus on General Game Playing)\citep{genesereth2005general} that use a variety of fitness measures to generate new game instances (e.g., either direct measures related to the game structure, or indirect measures such as player win-rates) \citep{risi2019increasing,browne2010evolutionary,togelius2011search,nielsen2015towards,wang2019paired};
and game-theoretic, multiplayer, or player-vs.-task notions (e.g., game balance, level of competition, social equality or welfare of the optimal game solution, etc.) \citep{byde2003applying,hom2007automatic,yannakakis2004evolving}. 
Overall, it is complex (and, arguably, not very useful) to introduce a unifying definition of interestingness that covers all of the above perspectives.

Thus, we focus here on a specialized notion of interestingess from the perspective of AI training, linked to the work of \citet{czarnecki2020}.
As mentioned earlier, \citet{czarnecki2020} introduce the notion of Games of Skill, which are of interest, in the sense of AI training, due to two axes of interactions between agents: 
a transitive axis enabling progression in terms of relative strength or skill, and a radial axis representing diverse intransitive/cyclical interactions between strategies of similar strength levels.  
The overall outlook provided by the above paper is that in these games of interest there exist many average-strength strategies with intransitive relations among them, whereas the level of intransitivity decreases as transitive strength moves towards an extremum (either very low, or very high strength);
the topology of strategies in a Game of Skill is, thus, noted to resemble a spinning top.

Through the spinning top paradigm, \citet{czarnecki2020} identifies several real-world games (e.g., Hex, Tic--Tac--Toe, Connect--Four, etc.) as Games of Skill.
Notably, the lower left cluster of games in \cref{fig:rwg_embeddings} highlights precisely these games.
Interestingly, while the Random Game of Skill, AlphaStar League, and Elo game (noise=0.5) are also noted as Games of Skill in \citet{czarnecki2020}, they are found in a distinct cluster in our landscape.
On closer inspection of these payoffs for this trio of games, there exists a strong correlation between the number of intransitive relations and the transitive strength of strategies, in contrast to other Games of Skill such as Hex.
Our landscape also seems to highlight non-interesting games. 
Specifically, variations of Blotto, Rock--Paper--Scissors, and the Disc Game are noted to not be Games of Skill in \citet{czarnecki2020}, and are also found to be in distinct clusters in \cref{fig:rwg_embeddings}.

Overall, these results highlight how topological analysis of multiplayer games can be used to not only study individual games, but also identify clusters of related (potentially interesting) games. 
For completeness, we also conduct additional studies in Supplementary Note 2, which compare our taxonomization of 2 $\times$ 2 normal-form games (and clustering into potential classes of interest) to that of \citet{bruns2015names}.

\subsection*{The Problem Problem and Procedural Game Structure Generation}\label{sec:procedural_game_gen}

\begin{figure}
    \centering{
        \phantomsubcaption\label{fig:problem_problem_game_gen_RPS}
        \phantomsubcaption\label{fig:problem_problem_game_gen_Elo_game}
        \phantomsubcaption\label{fig:problem_problem_game_gen_RPS_and_Elo_game}
        \phantomsubcaption\label{fig:problem_problem_game_gen_AlphaStar_and_go_board_size_3_komi_6_5_}
    }
    \includegraphics[width=\linewidth]{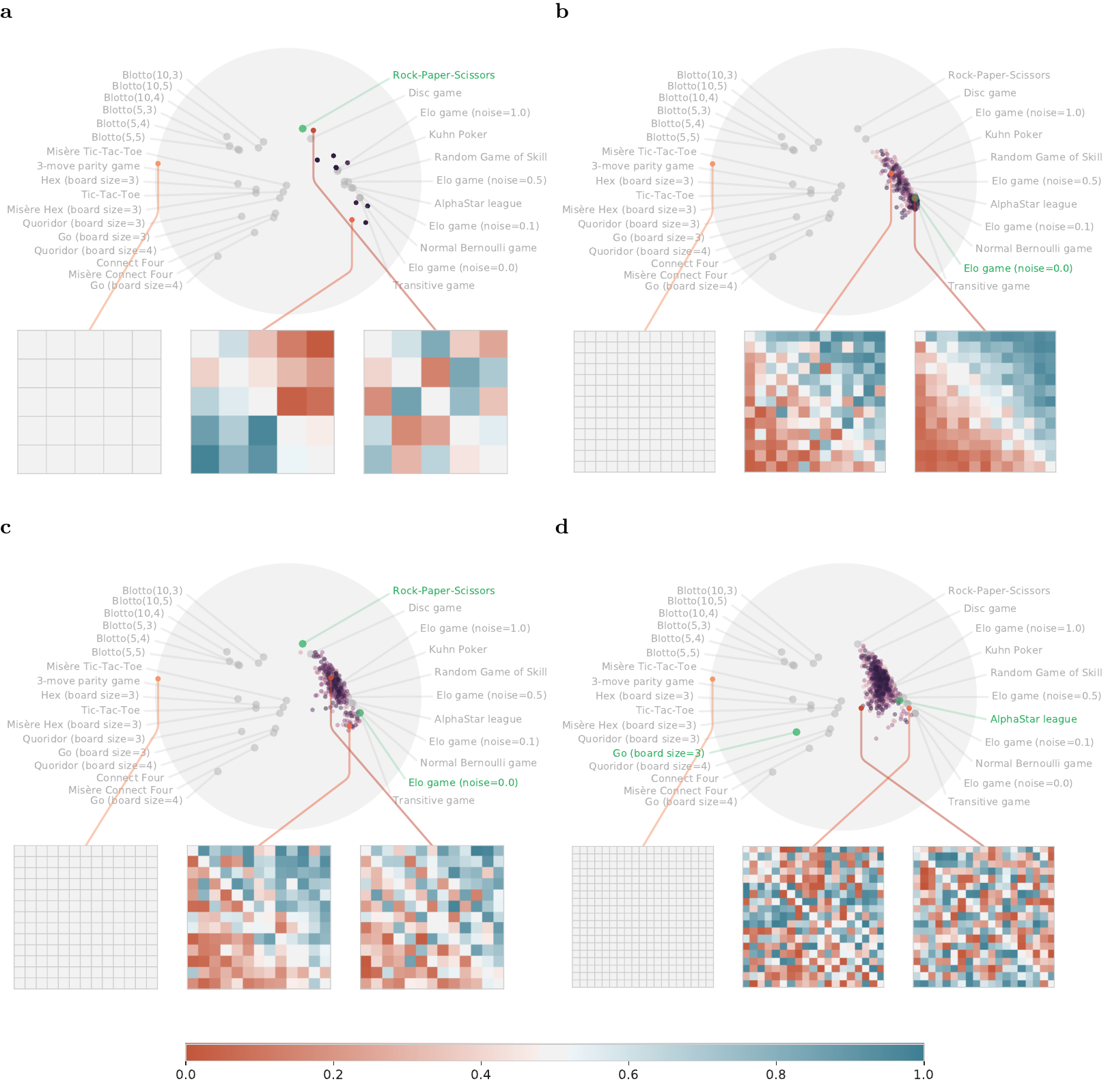}
    \caption{Visualization of procedural game structure generation projected in the games landscape. 
    Each figure visualizes the generation of a game of specified size, which targets a pre-defined game (or mixture of games) of a different size. The three payoffs in each respective figure, from left to right, correspond to the initial procedural game parameters, intermediate parameters, and final optimized parameters. 
    \subref{fig:problem_problem_game_gen_RPS} $5 \times 5$ generated game with the target game set to Rock--Paper--Scissors ($3 \times 3$).
    \subref{fig:problem_problem_game_gen_Elo_game} $13 \times 13$ generated game with the target game set to Elo ($1000 \times 1000$).
    \subref{fig:problem_problem_game_gen_RPS_and_Elo_game} $13 \times 13$ generated game with the target game set to the mixture of Rock--Paper--Scissors ($3 \times 3$) and Elo game ($1000 \times 1000$).
    \subref{fig:problem_problem_game_gen_AlphaStar_and_go_board_size_3_komi_6_5_} $19 \times 19$ generated game with the target game set to the mixture of AlphaStar League and Go (board size=3).
    Strategies are sorted by mean payoffs in \subref{fig:problem_problem_game_gen_Elo_game} and \subref{fig:problem_problem_game_gen_RPS_and_Elo_game} to more easily identify transitive structures expected from an Elo game.
    }
    \label{fig:problem_problem_game_gen}
\end{figure}

Having now established various graph-theoretic tools for characterizing games of interest, we revisit the so-called Problem Problem, which targets automatic generation of interesting environments.
Here we focus on the question of how we can leverage the topology discovered by our method to procedurally generate collections of new games (which can be subsequently analyzed, used for training, characterized as interesting or not per previous discussions, and so on). 

Full details of our game generation procedure are provided in the  \nameref{sec:methods} Section.
At a high level, we establish the feedback loop visualized in \cref{fig:method_workflow}, enabling automatic generation of games as driven by our graph-based analytical workflow.
We generate the payoff structure of a game (i.e., as opposed to the raw underlying game rules, e.g., as done by \citet{browne2010evolutionary}).
Thus, the generated payoffs can be considered either direct representations of normal form games, or empirical games indirectly representing underlying games with complex rules.
At a high level, given a parameterization of a generated game that specifies an associated payoff tensor, we synthesize its response graph and associated measures of interest (as done when generating the earlier landscape of games).
We use the multidimensional Elo parameterization for generating payoffs, due to its inherent ability to specify complex transitive and intransitive games~\citep{balduzzi2018re}. 
We then specify an objective function of interest to optimize over these graph-based measures. 
As only the evaluations of such graph-based measures (rather than their gradients) are typically available, we use a gradient-free approach to iteratively generate games optimizing these measures (CMA-ES~\citep{hansen2006cma} is used in our experiments). 
The overall generation procedure used can be classified as a Search-Based Procedural Content Generation technique (SBPCG) \citep{togelius2011search}.
Specifically, in accordance to the taxonomy defined by \citet{togelius2011search}, our work uses a direct encoding representation of the game (as the generated payoffs are represented as real-valued vectors of mElo parameters), with a theory-driven direct evaluation used for quantifying the game fitness/quality (as we rely on graph theory to derive the game features of interest, then directly minimize distances of principal components of generated and target games).

Naturally, we can maximize any individual game complexity measures, or a combination thereof, directly (e.g., entropy of the \alpharank distribution, number of 3-cycles, etc.).
More interestingly, however, we can leverage our low-dimensional landscape of games to directly drive the generation of new games towards existing ones with properties of interest.
Consider the instance of game generation shown in \cref{fig:problem_problem_game_gen_RPS}, which shows an overview of the above pipeline generating a $5 \times 5$ game minimizing Euclidean distance (within the low-dimensional complexity landscape) to the standard $3 \times 3$ Rock--Paper--Scissors game. 
Each point on this plot corresponds to a generated game instance.
The payoffs visualized, from left to right, respectively correspond to the initial procedural game parameters (which specify a game with constant payoffs), intermediate parameters, and final optimized parameters;
projections of the corresponding games within the games landscape are also indicated, with the targeted game of interest (Rock--Paper--Scissors here) highlighted in green. 
Notably, the final optimized game exactly captures the underlying rules that specify a general-size Rock--Paper--Scissors game, in that each strategy beats as many other strategies as it loses to.
In \cref{fig:problem_problem_game_gen_Elo_game}, we consider a larger $13 \times 13$ generated game, which seeks to minimize distance to a $1000 \times 1000$ Elo game (which is transitive in structure, as in our earlier motivating example in \cref{fig:elo_game_variations_transitive}).
Once again, the generated game captures the transitive structure associated with Elo games.

Next, we consider generation of games that exhibit properties of mixtures of several target games. 
For example, consider what happens if $3 \times 3$ Rock--Paper--Scissors were to be combined with the $1000 \times 1000$ Elo game above;
one might expect a mixture of transitive and cyclical properties in the payoffs, though the means of generating such mixed payoffs directly is not obvious due to the inherent differences in sizes of the targeted games.
Using our workflow, which conducts this optimization in the low-dimensional graph-based landscape, we demonstrate a sequence of generated games targeting exactly this mixture in \cref{fig:problem_problem_game_gen_RPS_and_Elo_game}.
Here, the game generation objective is to minimize Euclidean distance to the mixed principal components of the two target games (weighted equally).
The payoffs of the final generated game exhibit exactly the properties intuited above, with predominantly positive (blue) upper-triangle of payoff entries establishing a transitive structure, and the more sporadic positive entries in the lower-triangle establishing cycles. 

Naturally, this approach opens the door to an important avenue for further investigation, targeting generation of yet more interesting combinations of games of different sizes and rule-sets (e.g., as in \cref{fig:problem_problem_game_gen_AlphaStar_and_go_board_size_3_komi_6_5_}, which generates games targeting a mixture of Go (board size=3) and the AlphaStar League), and subsequent training of AI agents using such a curriculum of generated games.
Moreover, we can observe interesting trends when analyzing the Nash equilibria associated with the class of normal-form games considered here (as detailed in Supplementary Note 2).
Overall, these examples illustrate a key benefit of the proposed graph-theoretic measures in that it captures the underlying {structure} of various classes of games.
The characterization of games achieved by our approach directly enables the navigation of the associated games landscape to generate never-before-seen instances of games with fundamentally related structure.

\section*{Discussion}

In 1965, mathematician Alexander Kronrod stated that ``chess is the Drosophila of artificial intelligence"~\citep{mccarthy1997ai}, referring to the {genus} of flies used extensively for genetics research.
This parallel drawn to biology invites the question of whether a {family}, {order}, or, more concretely, shared structures linking various games can be identified. 
Our work demonstrated a means of revealing this topological structure, extending beyond related works investigating this question for small classes of games (e.g., $2 \times 2$ games~\citep{rapoport1966taxonomy,robinson2005topology,crandall2018cooperating}). 
We believe that such a topological landscape of games can help to identify and generate related games of interest for AI agents to tackle, as targeted by the Problem Problem, hopefully significantly extending the reach of AI system capabilities. 
As such, this paper presented a comprehensive study of games under the lens of graph theory and empirical game theory, operating on the response graph of any game of interest.
The proposed approach applies to general-sum, many-player games, enabling richer understanding of the inherent relationships between strategies (or agents), contraction to a representative (and smaller) underlying game, and identification of a game's inherent topology. 
We highlighted insights offered by this approach when applied to a large suite of games, including canonical games, empirical games consisting of trained agent policies, and real-world games consisting of representative sampled policies, extending well beyond typical characterizations of games using raw payoff visualizations, cardinal measures such as strategy or game tree sizes, or strategy rankings.
We demonstrated that complexity measures associated with the response graphs analyzed correlate well to the computational complexity of solving these games, and importantly enable the visualization of the landscape of games in relation to one another (as in \cref{fig:rwg_embeddings}).
The games landscape exposed here was then leveraged to procedurally generate games, providing a principled means of better understanding  and facilitating the solution of the so-called Problem Problem.

While the classes of games generated in this paper were restricted to the normal-form (e.g., generalized variants of Rock--Paper--Scissors), they served as an important validation of the proposed approach.
Specifically, this work provides a foundational layer for generating games that are of interest in a richer context of domains.
In contrast to some of the prior works in taxonomization of games (e.g., that of \citet{rapoport1966taxonomy,liebrand1983classification,bruns2015names}), a key strength of our approach is that it does not rely on human expertise, or manual isolation of patterns in payoffs or equilibria to compute a taxonomy over games.
We demonstrate an example of this in Supplementary Note 2, by highlighting similarities and differences of our taxonomization with those of \citet{bruns2015names} over a set of 144 2 $\times$ 2 games.
Importantly, these comparisons highlight that our response graph-based approach does not preclude more classical equilibrium-centric analysis from being conducted following clustering, while also avoiding the need for a human-in-the-loop analysis of equilibria for classifying the games themselves.

While our graph-based game {analysis} approach (i.e., the spectral analysis and clustering technique) applies to general-sum, many-player games, the procedural game structure {generation} approach used in our experiments is limited to zero-sum games (due to our use of the mElo parameterization).
However, any other general-sum payoff parameterization approaches (e.g., even direct generation of the payoff entries) can also be used to avoid the zero-sum constraint.
The study of normal-form games continues to play a prominent role in the game theory and machine learning literature \citep{BloembergenTHK15,IJCAI20-ishand,LiW20a,WrightWW19,WittFFTBW19,leibo2017multi,SpoonerS20} and as such the procedural generation of normal-form games can play an important role in the research community. 
An important line of future work will involve investigating means of generalizing this approach to generation of more complex classes of games.
Specifically, one way to generate more complex underlying games would be to parameterize core mechanisms of such a games class (either explicitly, or via mechanism discovery, e.g., using a technique such as that described by~\citet{charity2020mech}). Subsequently, one could train AI agents on a population of such games, constructing a corresponding empirical game, and using the response graph-based techniques used here to analyze the space of such games (e.g., in a manner reminiscent of~\citet{wang2019paired}, though under our graph-theoretic lens).
The connection to the SBPCG literature mentioned in \nameref{sec:procedural_game_gen} section also offers an avenue of alternative investigations into game structure generation, as the variations of fitness measures and representations previously explored in that literature \citep{togelius2011search} may be considered in lieu of the approach used here. 

Moreover, as the principal contribution of this paper was to establish a graph-theoretic approach for investigating the landscape of games, we focused our investigation on empirical analysis of a large suite of games.
As such, for larger games, our analysis relied on sampling of a representative set of policies to characterize them.
An important limitation here is that the empirical game-theoretic results are subject to the policies used to generate them.
While our sensitivity analysis (presented in Supplementary Note 2) seems to indicate that the combination of our policy sampling scheme and analysis pipeline produce fairly robust results, this is an important factor to revisit in significantly larger games.
Specifically, such a policy sampling scheme can be inherently expensive for extremely large games, making it important to further investigate alternative sampling schemes and associated sensitivities.
Consideration of an expanded set of such policies (e.g., those that balance the odds for players by ensuring a near-equal win probability) and correlations between the empirical game complexity and the complexity of the underlying policy representations (e.g., deep versus shallow neural networks or, whenever possible, Boolean measures of strategic complexity~\citep{feldman2000minimization,santos2018social}) also seem interesting to investigate.
Moreover, formal study of the propagation of empirical payoff variance to the topological analysis results is another avenue of future interest, potentially using techniques similar to~\citet{rowland2019multiagent}. 
Another direction of research for future work is to analyze agent-vs.-task games (e.g., those considered in~\citet{balduzzi2018re}) from a graph-theoretic lens. 
Finally, our approach is general enough to be applicable to other areas in social and life sciences\cite{smith1982evolution,miller2009complex,sigmund2010calculus,scheffer2009critical,may2008ecology}, characterized by complex ecologies often involving a large number of strategies or traits. In particular, these processes may be modeled through the use of large-scale response graphs or invasion diagrams\cite{fudenberg2006imitation,vasconcelos2017stochastic,imhof2005evolutionary,van2012emergence,donahue2020evolving}, whose overall complexity (and how it may vary with the inclusion or removal of species, strategies, and conflicts) is often hard to infer.

Overall, we believe that this work paves the way for related investigations of theoretical properties of graph-based games analysis, for further scientific progress on the Problem Problem and task theory, and further links to related works investigating the geometry and structure of games~\citep{rapoport1966taxonomy,huertas2003cartography,robinson2005topology,crandall2018cooperating,balduzzi2019open,czarnecki2020}.

\section*{Methods}\label{sec:methods}

\subsection*{Games}\label{sec:methods_games}
Our work applies to $K$-player, general-sum games, wherein each player $k \in [K]$ has a finite set $S^k$ of pure strategies. 
The space of pure strategy profiles is denoted $S = \prod_k S^k$, where a specific pure strategy profile instance is denoted $s = (s^1, \ldots, s^K) \in S$.
For a give profile $s \in S$, the payoffs vector is denoted $\mM(s) = (\mM^1(s),\ldots,\mM^K(s)) \in \mathbb{R}^K$, where $\mM^k(s)$ is the payoff for each player $k \in [k]$.  
We denote by $s^{-k}$ the profile of strategies used by all but the $k$-th player. 
A game is said to be zero-sum if $\sum_{k} \mM^k(s) = 0$ for all $s \in S$.
A game is said to be symmetric if all players have the same strategy set, and $\mM^k(s^1,\ldots,s^K) = \mM^{\rho(k)}(s^{\rho(1)}, \ldots, s^{\rho(K)})$ for all permutations $\rho$, strategy profiles $s$, and player indices $k \in [K]$.

\subsection*{Empirical games}
For the real-world games considered (e.g., Go, Tic--Tac--Toe, etc.), we conduct our analysis using an empirical game-theoretic approach~\citep{Walsh02,PhelpsPM04,wellman2006methods,Wellman13,TuylsPLHELSG20}.
Specifically, rather than consider the space of all pure strategies in the game (which can be enormous, even in the case of, e.g., Tic--Tac--Toe), we construct an {empirical game} over {meta-strategies}, which can be considered higher-level strategies over atomic actions.
In empirical games, a meta-strategy $s^k$ for each player $k$ corresponds to a sampled policy (e.g., in the case of our real-world games examples), or an AI agent (e.g., in our study of AlphaGo, where each meta-strategy was a specific variant of AlphaGo). 
Empirical game payoffs are calculated according to the win/loss ratio of these meta-strategies against one another, over many trials of the underlying games.
From a practical perspective, game-theoretic analysis applies to empirical games (over agents) in the same manner as standard games (over strategies); thus, we consider {strategies} and {agents} as synonymous in this work.
Overall, empirical games provide a useful abstraction of the underlying game that enables the study of significantly larger and more complex interactions.

\subsection*{Finite population models and \alpharank}\label{sec:methods_alpharank}
In game theory~\citep{nash:50,Weibull,Hofbauer98,Gintis09,Sandholm10}, one often seeks algorithms or models for evaluating and training strategies with respect to one another (i.e., models that produce a score or ranking over strategy profiles, or an equilibrium over them).
As a specific example, the Double Oracle algorithm~\citep{mcmahan2003planning}, which is used to quantify the computational complexity of solving games in some of our experiments, converges to Nash equilibria, albeit only in two-player zero-sum games. 
More recently, a line of research has introduced and applied the \alpharank algorithm~\citep{omidshafiei2019alpha,rowland2019multiagent,Muller2020A} for evaluation of strategies in {general-sum}, {n-player many-strategy} games.
\alpharank leverages notions from stochastic evolutionary dynamics in finite populations~\citep{nowak2004emergence,taylor2004evolutionary,imhof2005evolutionary,Nowak793,Traulsen06a,Traulsen06b} in the limit of rare mutations~\citep{fudenberg2006imitation,vasconcelos2017stochastic,veller2016finite}, which are subsequently analyzed to produce these scalar ratings (one per strategy or agent). 
At a high level, \alpharank models the probability of a population transitioning from a given strategy to a new strategy, by considering the additional payoff the population would receive via such a deviation.
These evolutionary relations are considered between all strategies in the game, and are summarized in its so-called {response graph}.
\alpharank then uses the stationary distribution over this response graph to quantify the long-term propensity of playing each of the strategies, assigning a scalar score to each. 

Overall, \alpharank yields a useful representation of the limiting behaviors of the players, providing a summary of the characteristics of the underlying game--albeit a 1-dimensional one (a scalar rating per strategy profile). 
In our work, we exploit the higher-dimensional structural properties of the \alpharank response graph, to make more informed characterizations of the underlying game, rather than compute scalar rankings.

\subsection*{Response graphs}\label{sec:methods_response_graphs}
The \alpharank response graph~\citep{omidshafiei2019alpha} provides the mathematical model that underpins our analysis. 
It constitutes an analogue (yet, not equivalent) model of the invasion graphs used to describe the evolution dynamics in finite populations in the limit when mutations are rare (see, e.g., \citet{fudenberg2006imitation,hauert2007via,vasconcelos2017stochastic,van2012emergence}). 
In this small-mutation approximation, directed edges stand for the fixation probability~\citep{nowak2004emergence} of a single mutant in a monomorphic population of resident individuals (the vertices), such that all transitions are computed through a processes involving only two strategies at a time. 
Here, we use a similar approach.
Let us consider a pure strategy profile $s = (s^1, \ldots, s^K)$. Consider a unilateral deviation (corresponding to a mutation) of player $k$ from playing $s^k \in S^k$ to a new strategy $\sigma^{k} \in S^k$, thus resulting in a new profile $\sigma=(\sigma^k, s^{-k})$.
The response graph associated with the game considers all such deviations, defining transition probabilities between all pairs of strategy profiles involving a unilateral deviation.
Specifically, let $\mE_{s, \sigma}$ denote the transition probability from $s$ to $\sigma$ (where the latter involves a unilateral deviation), defined as
\begin{align}\label{eq:alpharanktransition1}
    \mE_{s, \sigma} = 
    \begin{cases}
        \eta \frac{1- \exp\left( -\alpha  \left( \mM^k(\sigma) - \mM^k(s) \right) \right) }{1- \exp\left( -\alpha m \left( \mM^k(\sigma) - \mM^k(s) \right) \right)}  & \text{if } \mM^k(\sigma) \not= \mM^k(s)\\
        \frac{\eta}{m} & \text{otherwise}\, ,
    \end{cases}
\end{align}
where $\eta$ is a normalizing factor denoting the reciprocal of the total number of unilateral deviations from a given strategy profile, i.e., $\eta = (\sum_{l=1}^K (|S^l| - 1))^{-1}$.
Furthermore, $\alpha \geq 0$ and $m \in \mathbb{N}$ are parameters of the underlying evolutionary model considered and denote, respectively, to the so-called {selection pressure} and {population size}.

To further simplify the model and avoid sweeps over these parameters, we consider here the limit of infinite-$\alpha$ introduced by \citet{omidshafiei2019alpha}, which specifies transitions from lower-payoff profiles to higher-payoff ones with probability $\eta(1-\varepsilon)$, the reverse transition with probability $\eta\varepsilon$, and transition between strategies of equal payoff with probability $\nicefrac{\eta}{2}$, where $0<\varepsilon \ll 1$ is a small perturbation factor.
We use $\varepsilon = 1\mathrm{e}-10$ in our experiments, and found low sensitivity of results to this choice given a sufficiently small value.
For further theoretical exposition of \alpharank under this infinite-$\alpha$ regime, see \citet{rowland2019multiagent}.
Given the pairwise strategy transitions defined as such, the self-transition probability of $s$ is subsequently defined as,
\begin{align}
    \mE_{s, s} = 1 - \mkern-32mu \sum_{\substack{k \in [K] \\ \sigma | \sigma^k \in S^k \setminus \{s^k\} } } \mkern-30mu \mE_{s , \sigma} \, .
\end{align}
As mentioned earlier, if two strategy profiles $s$ and $\sigma$ do not correspond to a unilateral deviation (i.e., differ in more than one player's strategy), no transition occurs between them under this model (i.e., $\mE_{s,\sigma} = 0$). 

The transition structure above is informed by particular models in evolutionary dynamics as explained in detail in \citet{omidshafiei2019alpha}. 
The introduction of the perturbation term $\varepsilon$ effectively ensures the ergodicity of the associated Markov chain with row-stochastic transition matrix $\mE$.
This transition structure then enables definition of the \alpharank response graph of a game.
\begin{definition}[Response graph]
    The response graph of a game is a weighted directed graph (digraph) $G = (S, \mE)$ where each node corresponds to a pure strategy profile $s \in S$, and each weighted edge $\mE_{s,\sigma}$ quantifies the probability of transitioning from profile $s$ to $\sigma$.
\end{definition}
For example, the response graph associated with a transitive game is visualized in \cref{fig:method_workflow}b, where each node corresponds to a strategy $s$, and directed edges indicate transition probabilities between nodes.
\citet{omidshafiei2019alpha} define \alpharank $\vpi \in \Delta^{|S|-1}$ as a probability distribution over the strategy profiles $S$, by ordering the masses of the stationary distribution of $\mE$ (i.e., solution of the eigenvalue problem $\vpi^T\mE = \vpi^T$).
Effectively, the \alpharank distribution quantifies the average amount of time spent by the players in each profile $s \in S$ under the associated discrete-time evolutionary population model~\citep{fudenberg2006imitation}. 
Our proposed methodology uses the \alpharank response graphs in a more refined manner, quantifying the structural properties defining the underlying game, as detailed in the workflow outlined in \cref{fig:method_workflow} and, in more detail, below.

\subsection*{Spectral, clustered, and contracted response graphs}\label{sec:methods_workflow_spectral_rg}
This section details the workflow used to for spectral analysis of games' response graphs (i.e., the steps visualized in \cref{fig:method_workflow}c to \cref{fig:method_workflow}e).
Response graphs are processed in two stages: i) symmetrization (i.e., transformation of the directed response graphs to an associated undirected graph), and ii) subsequent spectral analysis.
This two-phase approach is a standard technique for analysis of directed graphs, which has proved effective in a large body of prior works (see \citet{malliaros2013clustering,van2015spectral} for comprehensive surveys).
Additionally, spectral analysis of the response graph is closely-associated with the eigenvalue analysis required when solving for the \alpharank distribution, establishing a shared formalism of our techniques with those of prior works.

Let $\mA$ denote the adjacency matrix of the response graph $G$, where $\mA = \mE$ as $G$ is a directed weighted graph.
We seek a transformation such that response graph strategies with similar relationships to neighboring strategies tend to have higher adjacency with one another.
Bibliometric symmetrization~\citep{satuluri2011symmetrizations} provides a useful means to do so in application to directed graphs, whereby the symmetrized adjacency matrix is defined $\widetilde{\mA} = \mA\mA^T + \mA^T\mA$.
Intuitively, in the first term, $\mA\mA^T$, the $(s,\sigma)$-th entry captures the weighted number of other strategies that both $s$ and $\sigma$ would deviate to in the response graph $G$; 
the same entry in the second term, $\mA^T\mA$, captures the weighted number of other strategies that would deviate to both $s$ and $\sigma$.
Hence, this symmetrization captures the relationship of each pair of response graph nodes $(s,\sigma)$ with respect to all other nodes, ensuring high values of weighted adjacency when these strategies have similar relational roles with respect to all other strategies in the game. 
More intuitively, this ensures that in games such as Redundant Rock--Paper--Scissors (see \cref{fig:redundant_rps_and_11_20_complexity_overview}, first row), sets of redundant strategies are considered to be highly adjacent to each other.

Following bibliometric symmetrization of the response graph, clustering proceeds as follows.
Specifically, for any partitioning of the strategy profiles $S$ into sets $S_1 \subset S$ and $\bar{S}_1 = S \setminus S_1$, define $w(S_1, \bar{S}_1) = \sum_{s \in S_1, \sigma \in \bar{S}_1} \mE_{s,\sigma}$.
Let the sets of disjoint strategy profiles $\{S_k\}_{k \in [K]}$ partition $S$ (i.e., $\bigcup_{k \in [K]} S_k = S$).
Define the $K$-cut of graph $G$ under partitions $\{S_k\}_{k\in [K]}$ as
\begin{align}
   cut(\{S_k\}) = \sum_k w(S_k,\bar{S}_k) \,,  \label{eq:cut}
\end{align}
which, roughly speaking, measures the connectedness of points in each cluster;
i.e., a low $cut$ indicates that points across distinct clusters are not well-connected.
A standard technique for cluster analysis of graphs is to choose the set of $K$ partitions, $\{S_k\}_{k \in [K]}$, which minimizes \cref{eq:cut}.
In certain situations, balanced clusters (i.e., clusters with similar numbers of nodes) may be desirable;
here, a more suitable metric is the so-called {normalized} $K$-cut, or $Ncut$, of graph $G$ under partitions $\{S_k\}_{k\in [K]}$,
\begin{align}
  Ncut(\{S_k\}) = \sum_k \frac{w(S_k,\bar{S}_k)}{w(S_k,S)} \,.  \label{eq:ncut}
\end{align}
Unfortunately, the minimization problem associated with \cref{eq:ncut} is NP-hard even when $K=2$ (see \citet{shi2000normalized}).
A typical approach is to consider a spectral relaxation of this minimization problem, which corresponds to a generalized eigenvalue problem (i.e., efficiently solved via standard linear algebra);
interested readers are referred to \citet{shi2000normalized,van2015spectral} for further exposition.
Define the Laplacian matrix $\mL = \mD - \widetilde{\mA}$ (respectively, $\mL = \mI - \mD^{-\nicefrac{1}{2}}\widetilde{\mA}\mD^{-\nicefrac{1}{2}}$), where degree matrix $\mD$ has diagonal entries $D_{i,i} = \sum_{j} \widetilde{\mA}_{i,j}$, and zeroes elsewhere.
Then the eigenvectors associated with the lowest nonzero eigenvalues of $\mL$ provide the desired spectral projection of the datapoints (i.e., spectral response graph), with the desired number of projection dimensions corresponding to the number of eigenvectors kept.
We found that using the unnormalized graph Laplacian $\mL = \mD - \widetilde{\mA}$ yielded intuitive projections in our experiments, which we visualize 2-dimensionally in our results (e.g., see \cref{fig:method_workflow}c).

The relaxed clustering problem detailed above is subsequently solved by application of a standard clustering algorithm to the spectral-projected graph nodes.
Specifically, we use agglomerative average-linkage clustering in our experiments (see \citet[Chapter 15]{rokach2005clustering} for details).
For determining the appropriate number of clusters, we use the approach introduced by \citet{pham2005selection}, which we found to yield more intuitive clusterings than the gap statistic~\citep{tibshirani2001estimating} for the games considered.

Following computation of clustered response graphs (e.g., \cref{fig:method_workflow}d), we contract clustered nodes (summing edge probabilities accordingly), as in \cref{fig:method_workflow}e. 
Note that for clarity, our visualizations only show edges corresponding to transitions from lower-payoff to higher-payoff strategies in the standard, spectral, and clustered response graphs, as these bear the majority of transition mass between nodes;
reverse edges (from higher- to lower-payoff nodes) and self-transitions are not visually indicated, despite being used in the underlying spectral clustering.
The exception is for contracted response graphs, where we do visualize weighted edges from higher- to lower-payoff nodes; 
this is due to the node contraction process potentially yielding edges with non-negligible weight in both directions. 

\subsection*{Low-dimensional landscape \& game generation}\label{sec:methods_workflow_game_gen}
We next detail the approach used to compute the games landscape and to procedurally generate games, as respectively visualized in \cref{fig:rwg_embeddings,fig:problem_problem_game_gen}.

To compute the low-dimensional games landscape, we use principal component analysis (PCA) of key features associated with the games' response graphs;
we use a collection of features we found to correlate well with the underlying computational complexity of solving these games (as detailed in the Results section).

Specifically, using the response graph $G$ of each game, we compute in- and out-degrees for all nodes, the entropy of the \alpharank distribution $\vpi$, and the total number of 3-cycles. 
We normalize each of these measures as follows:
dividing node-wise in- and out-degrees via the maximum possible degrees for a response graph of the same size;
dividing the \alpharank distribution entropy via the entropy of the uniform distribution of the same size;
finally, dividing the number of 3-cycles via the same measure for a fully connected directed graph. 

We subsequently construct a feature vector consisting of the normalized \alpharank distribution entropy, the normalized number of 3-cycles, and statistics related to normalized in- and out-degrees.
Specifically, we consider the mean, median, standard deviation, skew, and kurtosis of the in- and out-degrees across all response graph nodes, similar to the NetSimile~\citep{berlingerio2012netsimile} approach, which characterized undirected graphs.
This yields a feature vector of fixed size for all games.
We subsequently conduct a PCA analysis of the resulting feature vectors, visualizing the landscape in \cref{fig:rwg_embeddings} via projection of the feature vectors onto the top two principal components, yielding a low-dimensional embedding $\vv_{g}$ for each game $g$.

\subsection*{Procedural game structure generation}
The games generated in \cref{fig:elo_game_variations,fig:problem_problem_game_gen} use the multidimensional Elo (mElo) parametric structure~\citep{balduzzi2018re}, an extension of the classical Elo~\citep{elo78} rating system used in Chess and other games.
In mElo games, each strategy $i$ is characterized by two sets of parameters: i) a scalar rating $r_i \in \reals$ capturing the strategy's transitive strength, and ii) a $2k$-dimensional vector $\vc_i$ capturing the strategy's intransitive relations to other strategies. 
The payoff a strategy $i$ receives when played against a strategy $j$ in a mElo game is defined by $\emM(i,j) = \sigma(r_i - r_j + \vc_i^T \mOmega \vc_j)$
where $\sigma(z) = (1+\exp(-z))^{-1}$, $\mOmega = \sum_{i=1}^{k} \left( \ve_{2i-1} \ve_{2i}^T - \ve_{2i} \ve_{2i-1}^T \right)$, and $\ve_{i}$ is the unit vector with coordinate $i$ equal to 1.
This parametric structure is particularly useful as it enables definition of a wide array of games, ranging from those with fully transitive strategic interactions (e.g., those with a single dominant strategy, as visualized in \cref{fig:elo_game_variations_transitive}), to intransitive interactions (e.g., those with cyclical relations, as visualized in \cref{fig:elo_game_variations_cyclical}), to a mix thereof.

The procedural game structure generation visualized in \cref{fig:problem_problem_game_gen} is conducted as follows.
First, we compute the low-dimensional game embeddings for the collection of games of interest, as detailed above. 
Next, for an initially randomly-generated mElo game of the specified size and rank $k$, we concatenate the associated mElo parameters $r_i$ and $\vc_i$ for all strategies, yielding a vector of length $|S|(1+2k)$ fully parameterizing the mElo game, and constituting the decision variables of the optimization problem used to generate new games.
We used a rank $5$ mElo parameterization for all game generation experiments.
For any such setting of mElo parameters, we compute the associated mElo payoff matrix $\emM$, then the associated response graph and features, and finally project these features onto the principal components previously computed for the collection of games of interest, yielding the projected mElo components $\vv_{mElo}$.

Subsequently, given a game $g$ of interest that we would like to structurally mimic via our generated mElo game, we use a gradient-free optimizer, CMA-ES~\citep{hansen2006cma}, to minimize $||\vv_{mElo}-\vv_{g}||_2^2$ by appropriately setting the mElo parameters.
For targeting mixtures of games (e.g., as in \cref{fig:problem_problem_game_gen_RPS_and_Elo_game,fig:problem_problem_game_gen_AlphaStar_and_go_board_size_3_komi_6_5_}), we simply use a weighted mixture of their principal components $\vv_g$ (with equal weights used in our experiments).
We found the open-source implementation of CMA-ES~\citep{hansen2019pycma} to converge to suitable parameters within 20 iterations for all experiments, with the exception of the larger game generation results visualized in \cref{fig:problem_problem_game_gen_AlphaStar_and_go_board_size_3_komi_6_5_}, which required 40 iterations. 

\subsection*{Statistics}
To generate the distributions of Double Oracle iterations needed to solve the motivating examples (\cref{fig:elo_game_variations}), we used 20 generated games per class (transitive, cyclical, random), showing four examples of each in \cref{fig:elo_game_variations_transitive,fig:elo_game_variations_cyclical,fig:elo_game_variations_random}.
For each of these 20 games, we used 10 random initializations of the Double Oracle algorithm, reporting the full distribution of iterations.
To generate the complexity results in \cref{fig:graph_vs_computational_complexity}, we likewise used 10 random initializations of Double Oracle per game, with standard deviations shown in the scatter plots (which may require zooming in).
For the Spearman correlation coefficients shown in each of \cref{fig:double_oracle_game_results_alpharank_entropy,fig:double_oracle_game_results_num_3cycles,fig:double_oracle_game_results_indegree_stats_mean}, the reported p-value is two-sided and rounded to two decimals.

\subsection*{Data Availability}
We use OpenSpiel~\citep{lanctot2019openspiel} (\url{https://github.com/deepmind/open_spiel}) as the backend providing many of the games and associated payoff datasets studied here (see Supplementary Methods 1 for details).
Payoff datasets for empirical games in the literature are referenced in the main text.

\subsection*{Code Availability}
We use OpenSpiel~\citep{lanctot2019openspiel} (\url{https://github.com/deepmind/open_spiel}) for the implementation of \alpharank and Double Oracle.

%% file: main_additional_info.tex
\section*{Acknowledgements}
The authors gratefully thank Marc Lanctot and Thore Graepel for insightful feedback on the paper. FCS acknowledges the support of FCT-Portugal through grants PTDC/MAT/STA/3358/2014 and UIDB/50021/2020.
The authors thank the three anonymous reviewers and editor for their constructive and insightful feedback, which helped to significantly improve the clarity of the proposed method, experimental results, and discussions thereof.

\section*{Author contributions}
SO, KT, and FCS conceptualized the graph-based framework used in the paper.
SO implemented and generated the experimental results (spectral analysis of response graphs, clusterings, landscape of games, procedural games generation).
WMC devised and implemented the policy sampling scheme for real world games, and generated the payoff tables for the games listed in \cref{fig:rwg_embeddings}.
SO, KT, WMC, FCS, MR, JC, DH, PM, JP, BDV, AG, and RM contributed actively to discussions, analysis, writing, and review of the paper. 

\section*{Competing interests}
The authors declare no competing financial or non-financial interests.

\section*{Additional information}
\paragraph{Materials \& correspondence.}
Material requests and correspondence should be requested to Shayegan Omidshafiei (\url{somidshafiei@google.com}).

%% file: supplementary_info_text.tex
\section*{Supplementary Methods 1}
This section provides additional discussions and exposition of methods.

\subsection*{Policy sampling scheme}
The policy sampling scheme we use for large, real-world games follows the policy space coverage procedure first specified by~\citet{czarnecki2020}, and is detailed as follows.
First, we use a tree search algorithm, Alpha-Beta~\citep{newell1976computer} with varying tree depth limits $d$, and seeds $s$. This involves running the Alpha-Beta algorithm to depth $d$, and using random actions with seed $s$ thereafter (i.e., if the game does not terminate).
This yields policies of varying transitive strengths (controlled by depth $d$), with a range of related policies per depth (controlled by seed $s$).
This also covers the case of a purely random policy when $d$ is set to $0$ (with seed $s$ controlling the randomness).
Second, we repeat the same procedure with negated game payoffs, thus also covering the space of policies that seek to lose the original game.
Third, to further expand the policy space, we also define an augmented Alpha-Beta search, which assumes that branches of the game tree with depth beyond $d$ have a value of $0$. We likewise run this variant with negated payoffs as well.
Finally, to cover the policy space for more difficult games, we further augment this sampling strategy by using MCTS~\citep{coulom2006efficient} on each game, with $k$ simulations and varying seeds $s$.

For each variant of Alpha-Beta detailed above, we use depth parameters $d \in \{1,\ldots,9\}$.
For MCTS, we use $k \in \{10, 100, 1000\}$ simulations.
For all algorithms, we also sweep over seeds $s \in \{1, \ldots, 50\}$.
While this sampling procedure is a heuristic, it produces a range of policies with varying degrees of transitive and intransitive relations, and thus provide a useful approximation of the underlying game. 

\subsection*{Description of games analyzed}
We provide an overview of the games analyzed (noting that we omit descriptions of 11-20, AlphaGo, MuJoCo soccer, Blotto, and AlphaStar League, as they are detailed in the main text). 
As our methods operate on \alpharank response graphs, they apply to many-player general-sum games.
For the specific instances of two-player zero-sum games analyzed here, we symmetrize payoffs and standardize them such that $\emM \in [-1, 1]$.

\paragraph{Redundant Rock--Paper--Scissors.}
We modify the standard Rock--Paper--Scissors payoffs, $\mM_{RPS}$, by duplicating the first strategy (Rock), yielding the payoffs for the redundant variant, $\mM_{RRPS}$,
\begin{align}
    \mM_{RPS} =  
    \begin{bmatrix}
    0& -1& 1\\
    1& 0& -1\\
    -1& 1& 0\\
    \end{bmatrix} 
    \qquad
    \mM_{RRPS} =  
    \begin{bmatrix}
    0& 0& -1& 1\\
    0& 0& -1& 1\\
    1& 1& 0& -1\\
    -1& -1& 1& 0\\ 
    \end{bmatrix} \,.
\end{align}

\paragraph{Disc game.}
The Disc game~ is a cyclical game, defined as a differentiable generalization of the standard game of Rock--Paper--Scissors\citep{balduzzi2019open}.
We construct the Disc game payoffs as in~\citet{czarnecki2020}, by first uniformly sampling 1000 points, $\{S_i\}_{i\in[1000]}$, in the unit circle, subsequently defining payoffs,
\begin{align}
    \emM(i,j) =  
    S_i^T \begin{bmatrix}
    0 & -1\\
    1& 0\\
    \end{bmatrix}S_j \, .
\end{align}

\paragraph{Elo games and noisy variants.}
The variety of Elo games essentially correspond to the multidimensional Elo model detailed in the main paper, with the intransitive components removed, and payoffs rescaled.
Noisy variants of these Elo games are generated via specifying a noise parameter $\sigma^2$, adding zero-mean normally distributed noise of the specified variance to the non-noisy Elo payoff table, thus yielding noisy payoffs $\mM_{\sigma^2}$.
Finally, we symmetrize these payoffs, yielding final payoffs $\mM = \mM_{\sigma^2}-\mM_{\sigma^2}^T$.

\paragraph{Games in motivating examples.}
The variants of transitively-structured games in the motivating examples are Elo games, as specified above.
The cyclical games are $N \times N$ mElo games of rank $k=1$, with all transitive $r$ set to $0$, and intransitive components specified as follows: $c_{0:N-2,0} = [0, 1, \ldots, N-2]$, $c_{0:N-2,1} = [N-2, N-1, \ldots, 0]$, and $c_{N-1,:} = [-1, 1]$ 
The random-structured games are mElo games with rank $k=3$, and transitive and intransitive parameters i.i.d. sampled from $\mathcal{N}(0,1)$.

\paragraph{Transitive game.}
The Transitive Game payoffs are simply set to $+1$ for the upper-triangle, $-1$ for the lower-triangle, and $0$ across the diagonal. 

\paragraph{Random Game of Skill.}
The Random Game of Skill is a $1000 \times 1000$ game, generated as defined by~\citet{czarnecki2020}. 
Specifically, payoffs are $\mM(i,j) = 0.5(W_{ij}-W_{ji})+S_i-S_j$, where $W_{ij}$, $W_{ji}$, $S_i$, and $S_j$ are sampled from $\mathcal{N}(0,1)$. 
In the Normal Bernoulli Game, parameters $S_i$ and $S_j$ are sampled likewise, while $W_{ij}$ and $W_{ji}$ are sampled from $\mathcal{U}(0,1)$ and payoffs are specified as $\mM(i,j) = W_{ij}-W_{ji}+S_i-S_j$. 

\paragraph{3-move parity game.}
The parity game is generated per the definition in~\citet{czarnecki2020}.

\paragraph{Real-world games.}
We use the OpenSpiel~\citep{lanctot2019openspiel} implementations of the following games: Tic--Tac--Toe, Hex (board size=3), Quoridor (board size=3), Quoridor (board size=4), Go (board size=3), Go (board size=4), 
Connect Four, Kuhn Poker.
For Go, we use a Komi (first-move advantage) of 6.5 points.
With the exception of Kuhn Poker (which has only 12 information states, and fully enumerable policy space), we use the policy sampling scheme detailed above for this collection of games. 

\section*{Supplementary Note 1: Related Work}

This paper largely focuses on topological analysis and taxonomization of multiplayer games, via study of the interactions possible within them.
As such, this work sits at the intersection of several different research disciplines, including game theory, machine learning, multiagent systems and network science.

A motivating application for such an approach is to classify or cluster games that present interesting challenges for artificial agents.
The central question revolving around the interestingness of environments for artificial learning agents has a long history in machine learning, task theory~\citep{thorisson2016artificial}, procedural content generation~\citep{togelius2011search}, and curriculum learning~\citep{bengio2009curriculum}.

An overview of the role of games in AI, including evaluation and generation of the games themselves, is provided in \citet{yannakakis2018artificial}.
\citet{thorisson2016artificial} emphasizes the need for a task theory in AI, a unifying and formal framework that enables comparison, abstraction, characterization, and decomposition of agent tasks (and the associated environments within which tasks are defined).
The decompositionality of tasks enabled by such a theory can prove useful for not only understanding the task-space, but also enabling more efficient agent training via curriculum-based teaching, as discussed in \citet{bieger2018task}.
An overview of recent AI benchmarks and platforms is provided in~\citet{hernandez2017new}, which also includes pointers to workshops and recent works investigating the potential taxonomization of such a wide suite of tasks and environments.  
Overall, our contribution can be broadly classified as falling under the task theory regime, as a step towards characterizing the particular set of tasks studied here (i.e., multiplayer games that can be characterized via empirical payoff tables) via graph- and game-theoretic techniques.

Procedural generation of game content has been used in commercially-available games for decades, spanning from the dungeon-crawler and space trading games Rogue and Elite, respectively released in 1980 and 1984 \citep{toy1980,braben1984elite}, to more recent expansive 3D open-world games such as No Man's Sky \citep{nomanssky2016}.
Significant research has also been conducted in this field, 
thorough overviews provided by \citet{togelius2011search,shaker2016procedural,risi2019increasing}.
Other notable examples of procedural content generation and automated game design include~\citet{smith2011answer,nelson2007towards,cook2011multi,cook2016angelina}.
The complexity of the environments generated by these and related techniques has been substantially increasing in recent years.
Recently, \citet{juliani2019obstacle} introduced a procedurally generated 3D tower environment for training AI agents, which increases complexity of associated skills required by the agent to solve the task. 
A technique for iterative procedural task generation and AI training was introduced recently by \citet{wang2019paired}, and further generalized in~\citet{wang2020enhanced}.

The literature on General Game Playing~\citep{genesereth2005general}, in particular, has close connections to the topic studied here.
Specifically, works such as~\citet{browne2010evolutionary} investigate generation of complete rule-sets for extensive-form games, culminating in design of a commercially-available game (Yavalath).
Related works on rule-based game generation~\citep{togelius2014characteristics} include examples generating pacman-like games~\citep{togelius2008experiment} and chess-like games~\citep{kowalski2016evolving}, with a useful survey provided in~\citet{nelson2016rules}.
A comparative evaluation of games and agents generated is provided in~\citet{perez2019general}.

Generation of games with attributes desirable by humans has also seen some attention, including  querying of human preferences (specifically, fun, challenge, and frustration) for personalized content generation in \citet{shaker2010towards}, creation of more believable environments \citep{camilleri2016platformer}, and creation of balanced board games \citep{hom2007automatic}.
Works have also taken an agent-vs.-task perspective for evaluating both agent performance and task/game quality \citep{nielsen2015towards,balduzzi2018re}.
A number of works have also considered specific notions of the interestingness of a game (or an evaluation of the fitness of a game from varying perspectives)~\citep{vygotsky1978interaction,deterding2015lens,lazzaro2009we,prensky2001fun,wang2011game,baker2019emergent,genesereth2005general,risi2019increasing,browne2010evolutionary,togelius2011search,nielsen2015towards,wang2019paired,byde2003applying,hom2007automatic,yannakakis2004evolving}; 
we refer readers to the Landscape of Games section of the main text for discussion of these works.

In curriculum learning, initial work concerned the use of curricula of tasks in training neural networks via supervised learning~\citep{elman1993learning,sanger1994neural,krueger2009flexible,bengio2009curriculum}. This continues to be an active field of research to this day, with many new methods for curriculum learning being developed to match the increasingly intricate architectures used within deep learning~\citep{graves2017automated,weinshall2018curriculum,hacohen2019power}, and the increasingly wide range of applications for deep neural networks, such as deep reinforcement learning~\citep{florensa2017reverse,cobbe2018quantifying,justesen2018illuminating,silver2017mastering,vinyals19}. Recent work has brought open-ended learning via multiagent interaction and procedural environment generation to the forefront of curriculum learning~\citep{justesen2018illuminating,leibo2019autocurricula,everett2019optimising,balduzzi2019open,wang2019paired,czarnecki2020}.

This work contributes to this line of inquiry by framing the intrinsic interestingness of games as a driving factor in open-ended multiagent learning. While equilibrium computation in many classes of games is PPAD-complete (thus generally considered intractable~\citep{daskalakis2009complexity,chen2009settling,daskalakis2013complexity}), and while the computational complexity results in this work focus on the two-player zero-sum case, it is worth mentioning that our methods also directly apply to many-player general-sum settings (as exemplified by the analysis of the general-sum 11-20 game in the main text, and of many-player variants of Kuhn Poker in \cref{fig:additional_rg_results_1}). 

Several prior works have analyzed game-theoretic response graphs from the perspective of agent evaluation~\citep{wang2003reinforcement,ungureanu2005nash,mirrokni2004convergence,apt2015classification,omidshafiei2019alpha,rowland2019multiagent,Muller2020A},
and more generally there is a wide literature concerning analysis of directed graphs across a range of domains and disciplines~\citep{page1999pagerank,kim2008complex,costa2007characterization,hausmann2014atlas}. Additionally, there is an established line of work in game theory that seeks a discrete classification of games based on payoff tables.
The space of 2x2 normal-form games have been classified into groups with restrictions, such as ordering the payouts~\citep{rapoport1966taxonomy,liebrand1983classification,bruns2015names}, and into a {periodic table} through topological and graph-theoretic techniques~\citep{robinson2005topology}. 
We provide a comparison against the approach of \citet{bruns2015names} below.
More recent work has introduced a framework for binary classification of agent behaviors (e.g., as strategic and non-strategic) in simultaneous-move normal form games~\citep{wright2018formal}.

Social dilemma games (such as Prisoners Dilemma, Chicken, or Hoarding), categorized by~\citet{dawes1980social, liebrand1983classification},
are a subset of games encapsulating the trade-offs found in social dilemmas.
Social dilemmas such as the tragedy of the commons show the potential conflict between individual and group self-interest~\citep{hardin1968tragedy, olson2012logic, platt1973social}.
Early work in social dilemmas showed how the dilemmas can arise from reinforced behavior~\citep{platt1973social}.
More recently, reinforcement learning (RL) has been applied to sequential social dilemmas~\citep{leibo2017multi} where strategic decisions to cooperate or defect were shown to coincide with the learning of policies. 
Agent behavior was also shown to depend on factors such as the relative difficulties of learning policies and in the case of cooperation, possibly non-occurring coordination sub-problems.
RL has also been used to train agents that cooperate with humans across a variety of two-player repeated games\citep{crandall2018cooperating}.

In addition to analysis of individual graphs, there has been much work on the analysis and clustering of collections of graphs in a variety of contexts~\citep{Sanfeliu83,berlingerio2012netsimile,koutra2013deltacon,tsitsulin2018netlsd}.
The graph-based analysis conducted herein draws on these works (e.g., using several local and global graph features and distributions to characterize our games, as in~\citet{berlingerio2012netsimile}).
However, our approach focuses specifically on games, exploiting several measures related to the complexity of games, enabling navigation of their underlying landscape and subsequent generation of new games of interest.

A wide-spread strategy comparison method in machine learning is the Elo rating~\citep{elo78} which ranks agents against one another in a purely transitive manner. \citet{balduzzi2018re} showed the importance of intransitive strategies when comparing agents, leading to the so called multidimensional Elo model, modeling more complex relations between agents. Finally, in a line of work related to training of agents, Policy-Space Response Oracles (PSRO)\citep{lanctot2017unified} and Rectified PSRO~\citep{balduzzi2019open} lead to learning a collection of possibly intransitive strategies, but not to directly analyze them, as done in this paper.

\section*{Supplementary Note 2: Additional results}
This section provides additional results, including response graphs and experiments conducted for generating the landscapes in the main text.

\subsection*{Taxonomization of 2 $\times$ 2 games}
Here we conduct a set of experiments comparing our approach against that of \citet{bruns2015names} (focusing on the ordinal games results presented in Figure 1 of their paper).

\begin{table}
    \centering
    \setlength{\tabcolsep}{3pt}
    \caption{$2 \times 2$ games and row player payoffs from \citet{bruns2015names}. Important note: corresponding column player payoffs are defined as a transpose along the {anti}-diagonal in \citet{bruns2015names} (i.e., not according to the typical convention of using standard transposes for symmetric games). This latter point has no impact on quantitative results, but is important for readers qualitatively analyzing the payoff structures and corresponding results below.}
    \resizebox{\textwidth}{!}{%
        \begin{tabular}{cccccccccccc}
        \toprule
        \textbf{Chicken} &
        \textbf{Battle} &
        \textbf{Hero} &
        \textbf{Compromise} &
        \textbf{Deadlock} &
        \textbf{Prisoner's dilemma} &
        \textbf{Stag hunt} &
        \textbf{Assurance} &
        \textbf{Coordination} &
        \textbf{Peace} &
        \textbf{Harmony} &
        \textbf{Concord} \\
        \textbf{(Ch)} &
        \textbf{(Ba)} &
        \textbf{(Hr)} &
        \textbf{(Cm)} &
        \textbf{(Dl)} &
        \textbf{(Pd)} &
        \textbf{(Sh)} &
        \textbf{(As)} &
        \textbf{(Co)} &
        \textbf{(Pc)} &
        \textbf{(Ha)} &
        \textbf{(Nc)} \\
        \midrule
        $\begin{bmatrix} 2 & 3 \\ 1 & 4 \end{bmatrix}$ &
        $\begin{bmatrix} 3 & 2 \\ 1 & 4 \end{bmatrix}$ &
        $\begin{bmatrix} 3 & 1 \\ 2 & 4 \end{bmatrix}$ &
        $\begin{bmatrix} 2 & 1 \\ 3 & 4 \end{bmatrix}$ &
        $\begin{bmatrix} 1 & 2 \\ 3 & 4 \end{bmatrix}$ &
        $\begin{bmatrix} 1 & 3 \\ 2 & 4 \end{bmatrix}$ &
        $\begin{bmatrix} 1 & 4 \\ 2 & 3 \end{bmatrix}$ &
        $\begin{bmatrix} 1 & 4 \\ 3 & 2 \end{bmatrix}$ &
        $\begin{bmatrix} 2 & 4 \\ 3 & 1 \end{bmatrix}$ &
        $\begin{bmatrix} 3 & 4 \\ 2 & 1 \end{bmatrix}$ &
        $\begin{bmatrix} 3 & 4 \\ 1 & 2 \end{bmatrix}$ &
        $\begin{bmatrix} 2 & 4 \\ 1 & 3 \end{bmatrix}$ \\
        \bottomrule
        \end{tabular}
    }
    \label{fig:2x2_payoffs_bruns}
\end{table}

\begin{figure}
    \begin{subfigure}{0.45\textwidth}
        \centering
        \caption{}
        \includegraphics[width=0.9\textwidth]{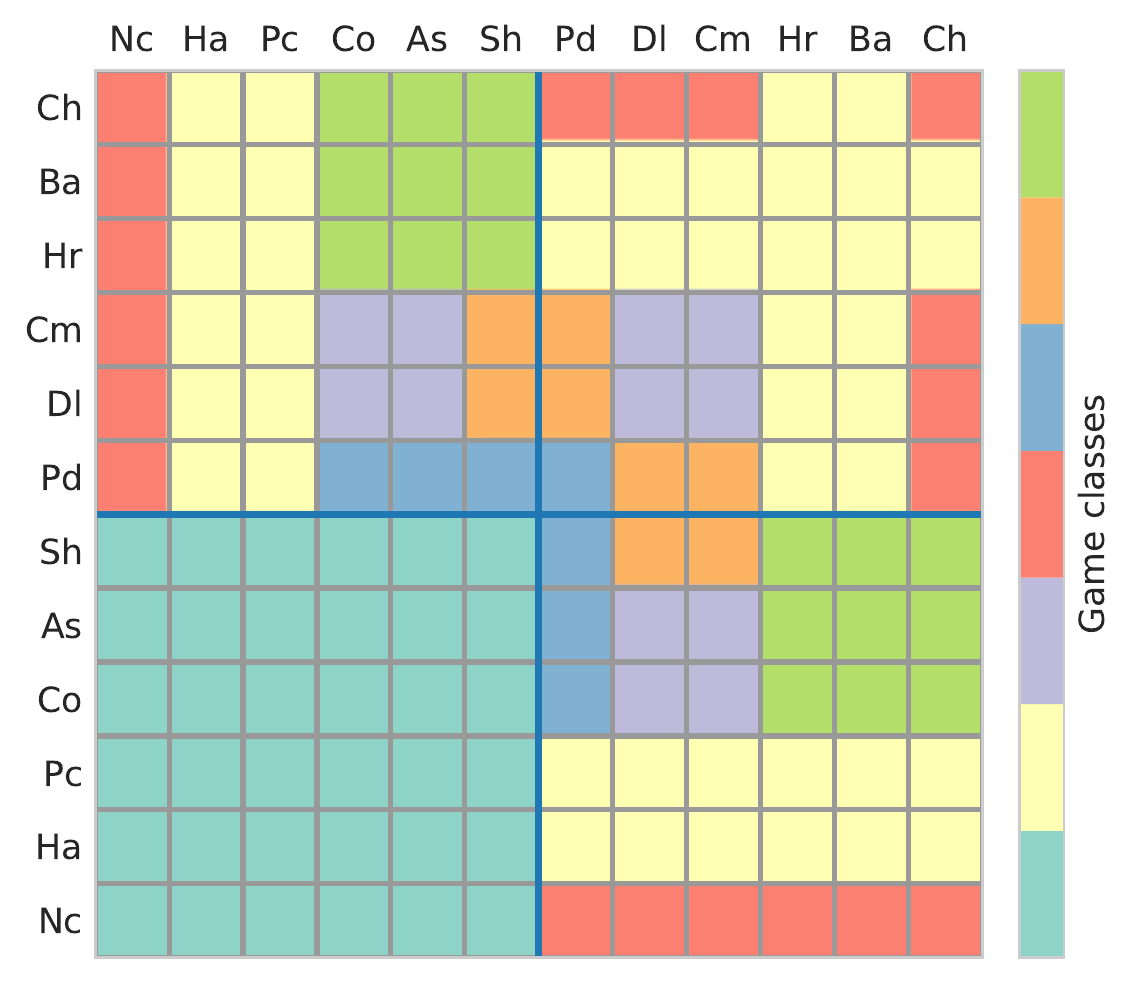}
        \label{fig:2x2_taxonomy_bruns}
    \end{subfigure}%
    \hfill%
    \begin{subfigure}{0.45\textwidth}
        \centering
        \caption{}
        \includegraphics[width=0.9\textwidth]{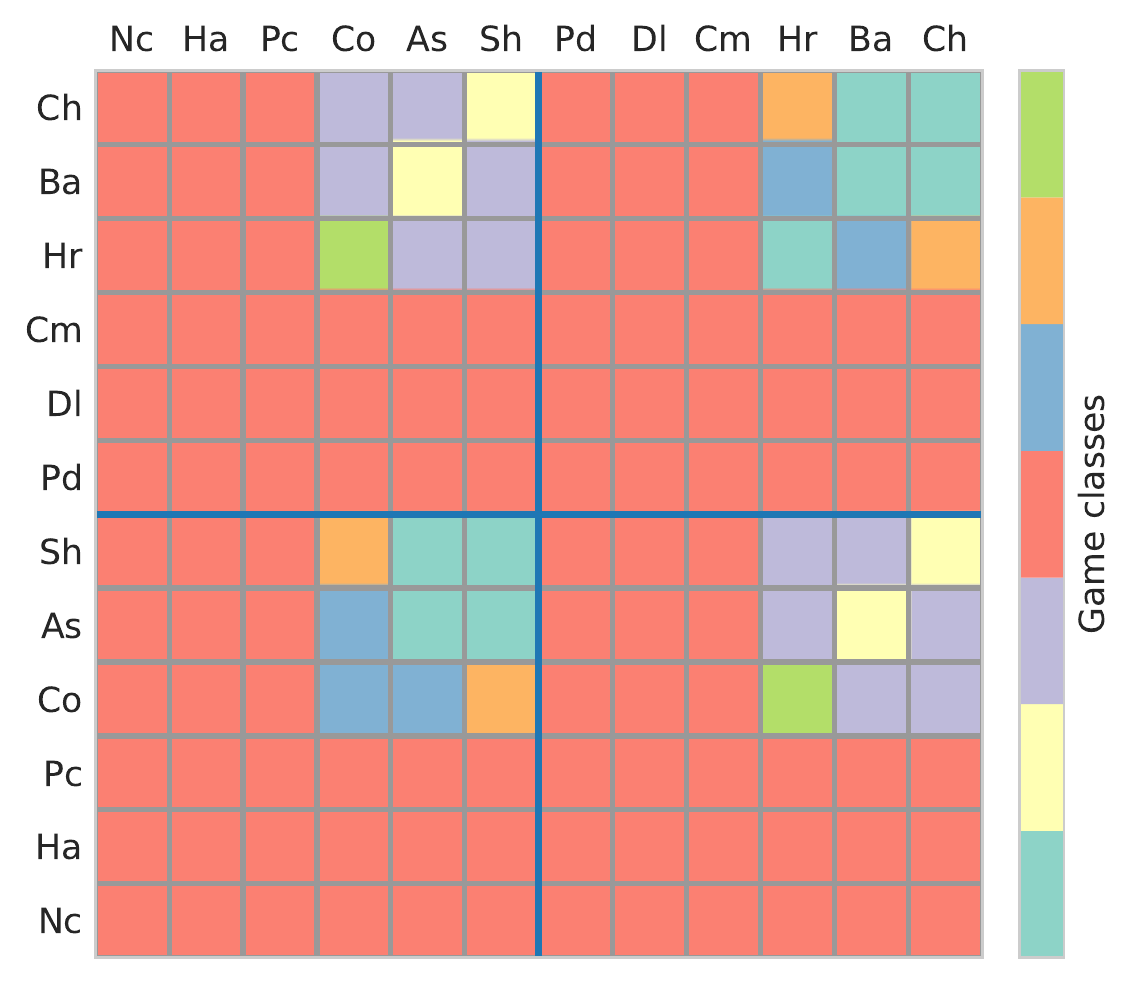}
        \label{fig:2x2_taxonomy_ours_alpha_0p2}
    \end{subfigure}\\
    \begin{subfigure}{0.45\textwidth}
        \centering
        \caption{}
        \includegraphics[width=0.9\textwidth]{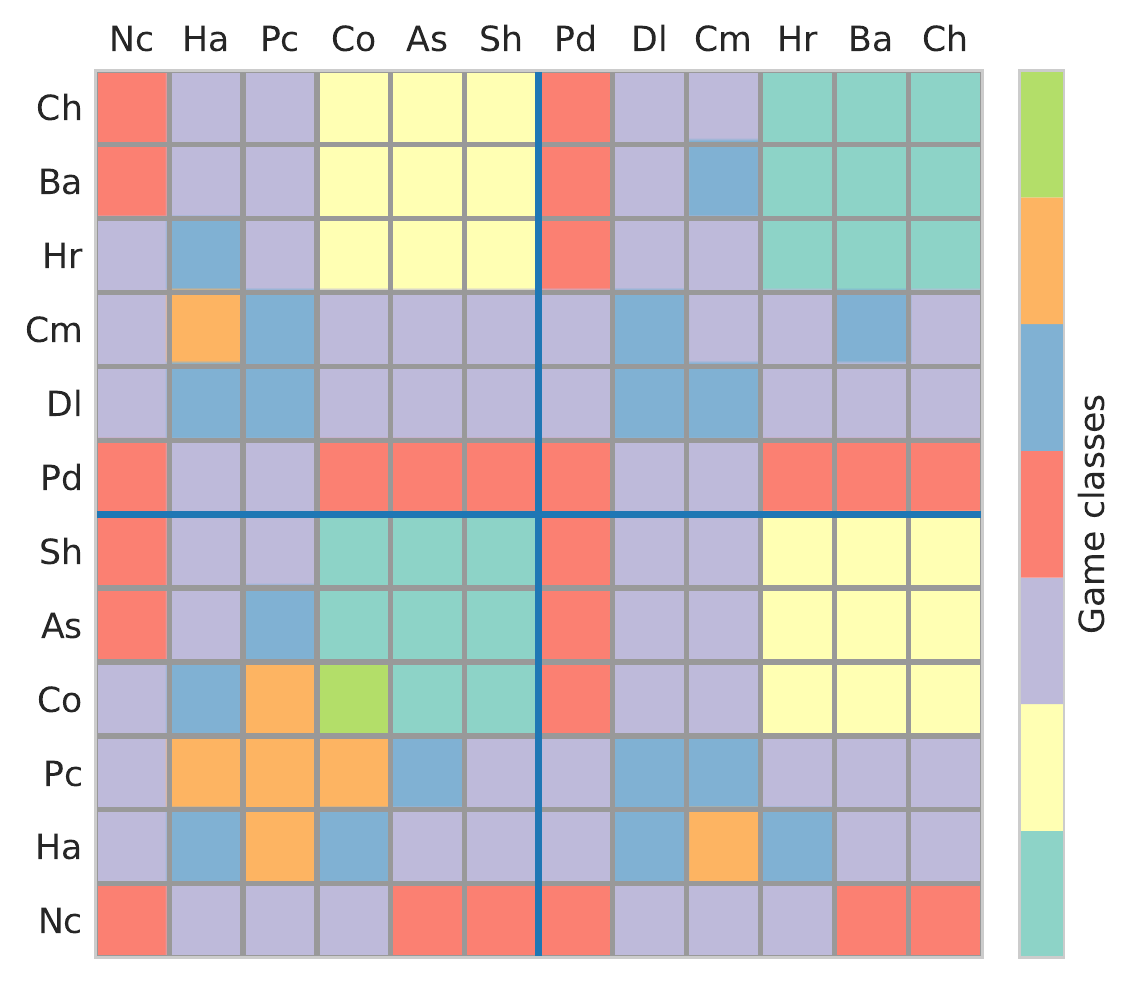}
        \label{fig:2x2_taxonomy_ours_alpha_0p01}
    \end{subfigure}%
    \hfill%
    \begin{subfigure}{0.45\textwidth}
        \centering
        \caption{}
        \includegraphics[width=0.9\textwidth]{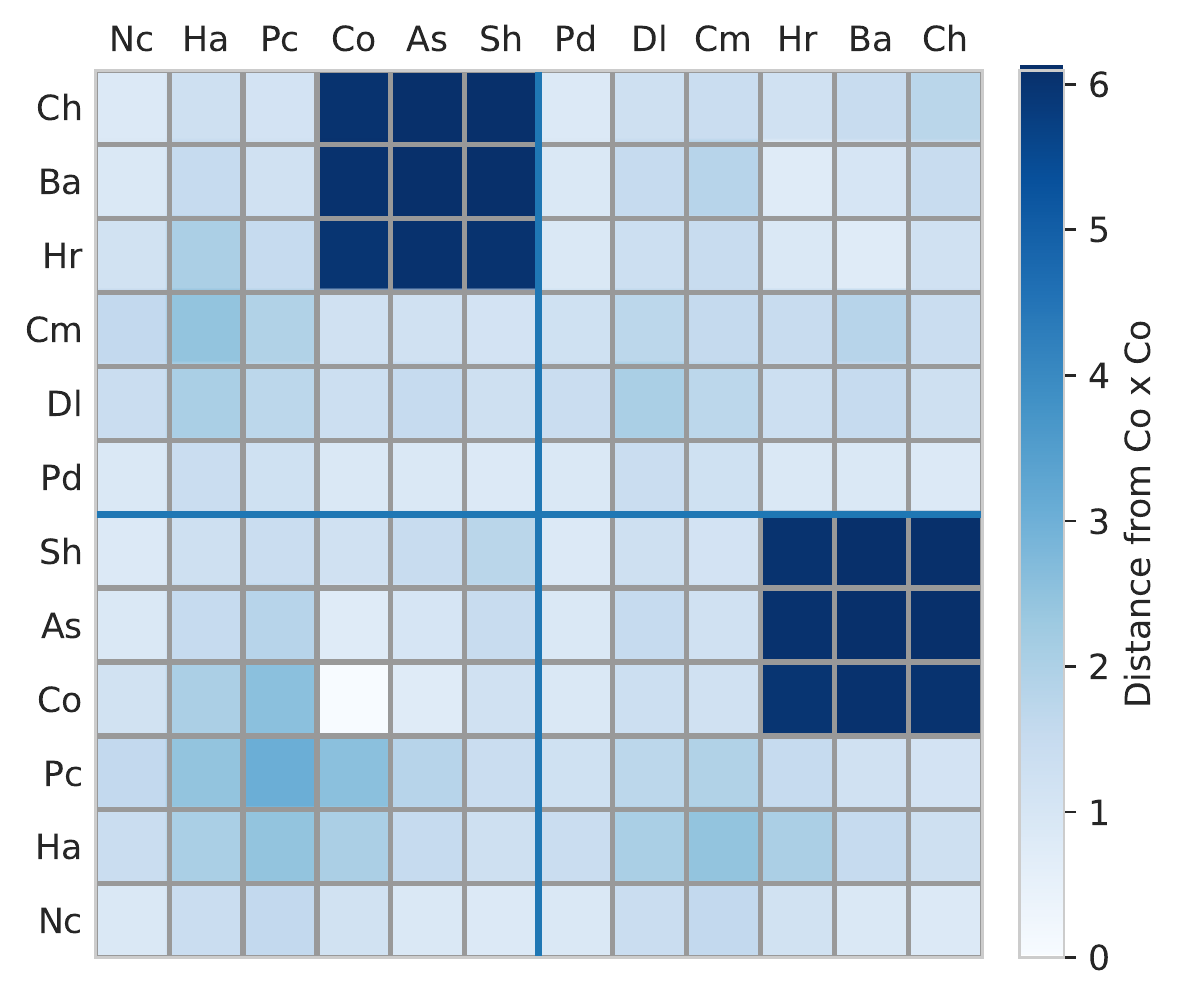}
        \label{fig:2x2_taxonomy_ours_alpha_0p01_distance_Co_Co}
    \end{subfigure}\\
    \begin{subfigure}{0.45\textwidth}
        \centering
        \caption{}
        \includegraphics[width=0.9\textwidth]{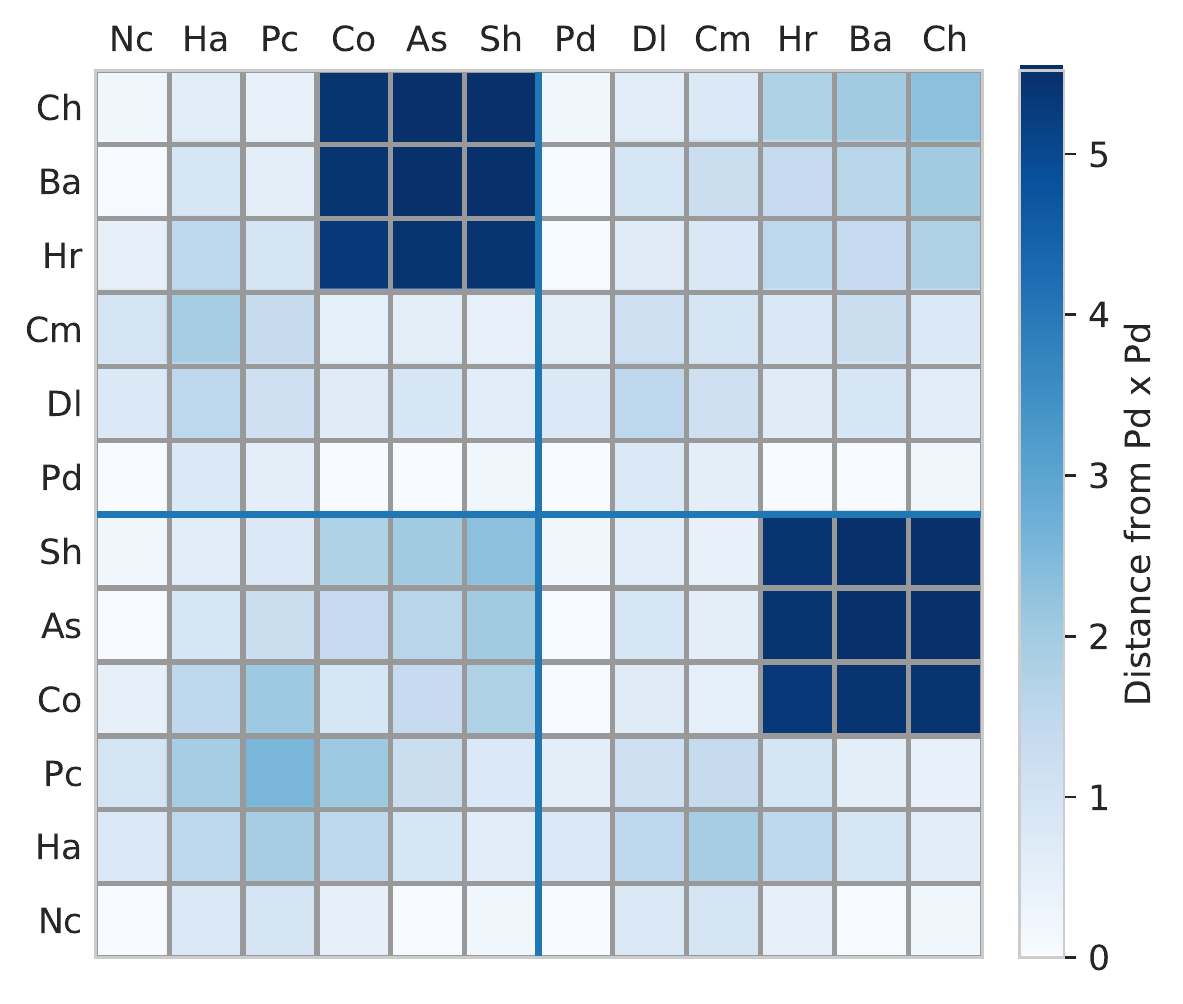}
        \label{fig:2x2_taxonomy_ours_alpha_0p01_distance_Pd_Pd}
    \end{subfigure}%
    \hfill%
    \begin{subfigure}{0.45\textwidth}
        \centering
        \caption{}
        \includegraphics[width=0.9\textwidth]{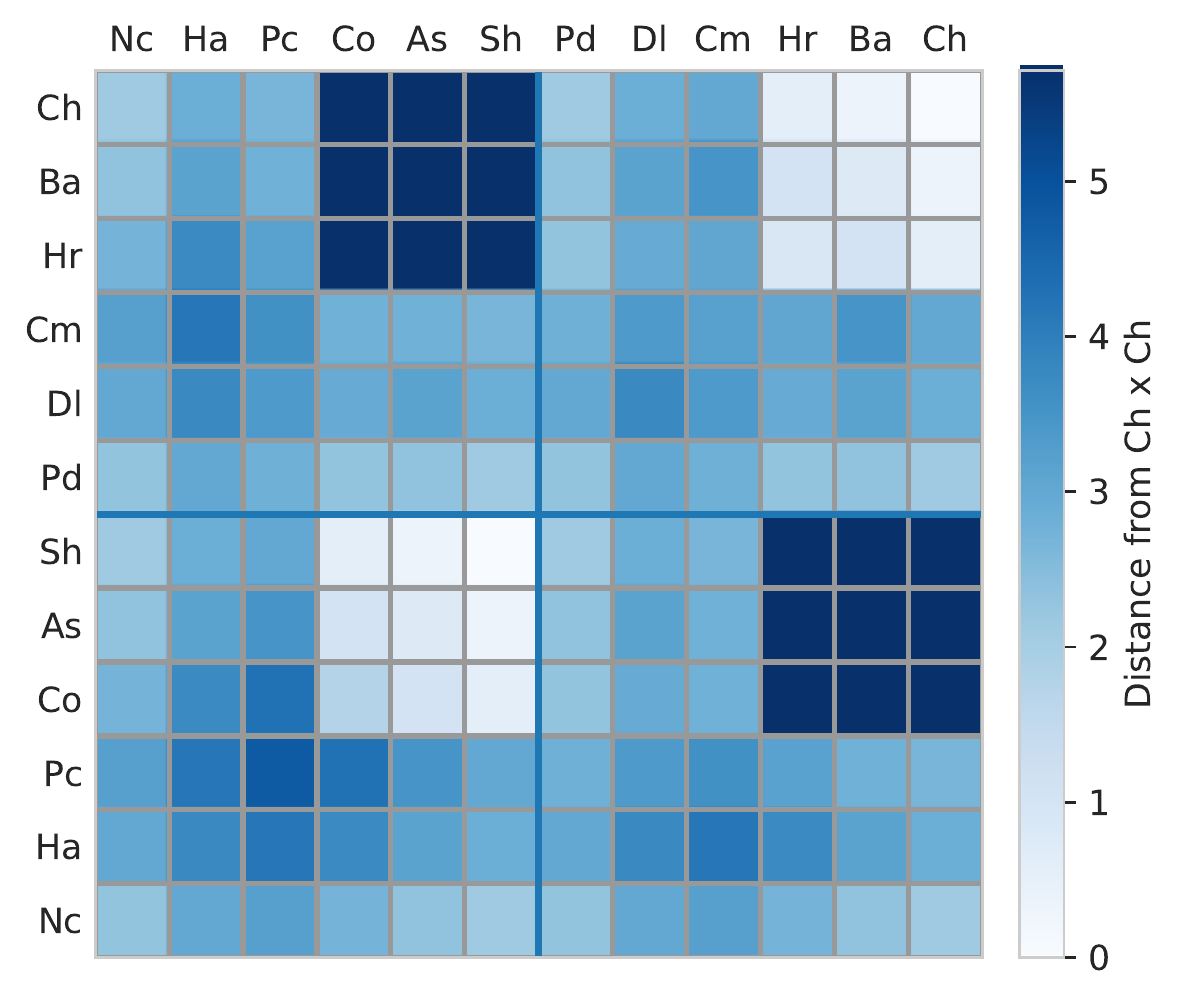}
        \label{fig:2x2_taxonomy_ours_alpha_0p01_distance_Ch_Ch}
    \end{subfigure}
    \caption{Comparison of our approach to that of \citet{bruns2015names} for $2 \times 2$ games.
    \subref{fig:2x2_taxonomy_bruns} Clusters (Bruns).
    \subref{fig:2x2_taxonomy_ours_alpha_0p2} Clusters (ours, high \alpharank $\alpha$, where $\alpha=0.2$).
    \subref{fig:2x2_taxonomy_ours_alpha_0p01} Clusters (ours, lower \alpharank $\alpha$, where $\alpha=0.01$).
    \subref{fig:2x2_taxonomy_ours_alpha_0p01_distance_Co_Co} Game distances (ours, to $Co \times Co$).
    \subref{fig:2x2_taxonomy_ours_alpha_0p01_distance_Pd_Pd} Game distances (ours, to $Pd \times Pd$).
    \subref{fig:2x2_taxonomy_ours_alpha_0p01_distance_Ch_Ch} Game distances (ours, to $Ch \times Ch$).
    }
    \label{fig:supp_2x2_games_bruns_comparison}
\end{figure}

\begin{figure}
    \centering
    \includegraphics[height=0.95\textheight]{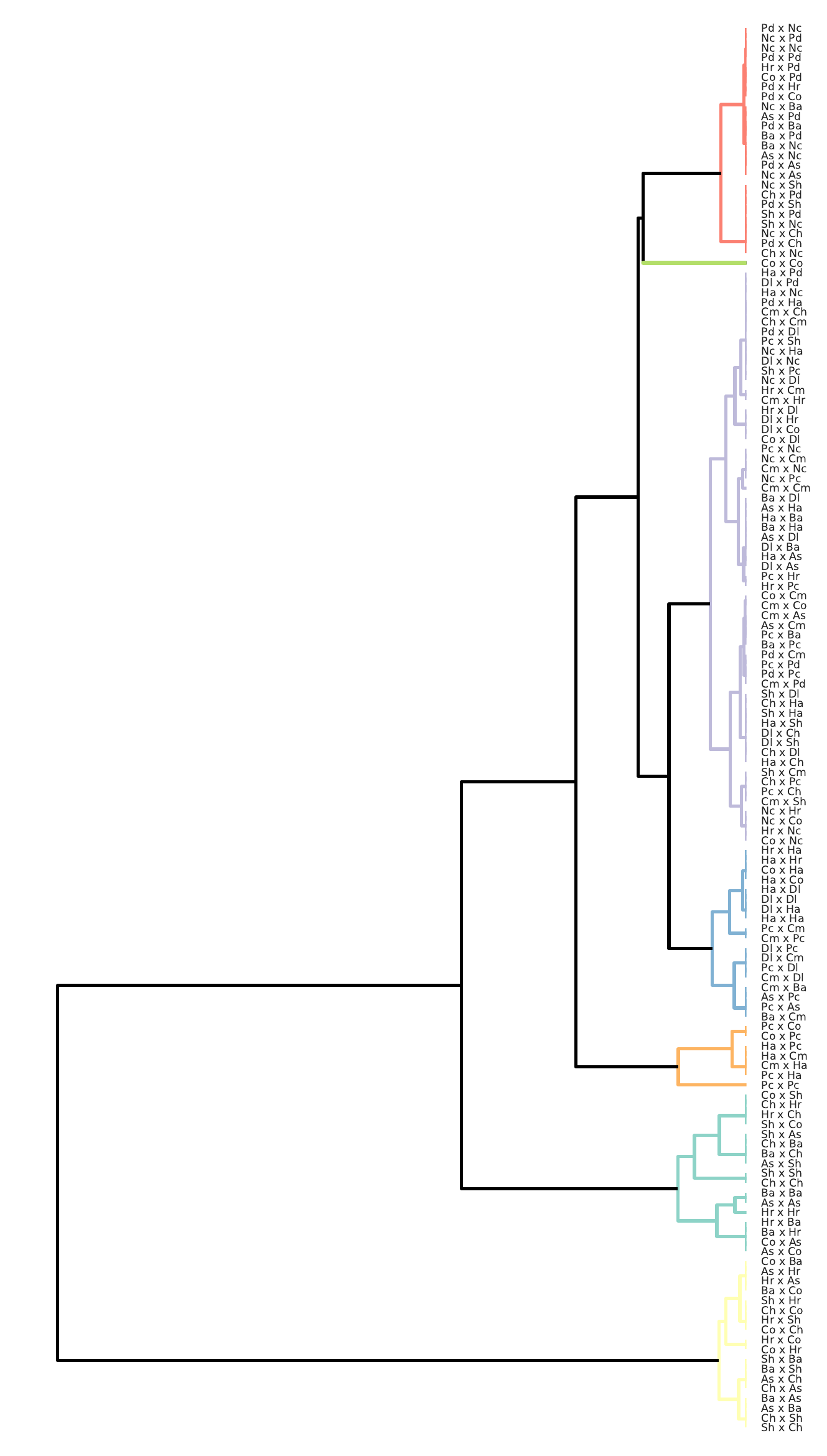}
    \caption{Dendrogram of $2 \times 2$ games, computed using our approach. All 144 games are visualized here, with classes colored in accordance to \cref{fig:2x2_taxonomy_ours_alpha_0p01}.
    }
    \label{fig:supp_2x2_games_phylo_tree}
\end{figure}

Let us first provide a summary of the approach of \citet{bruns2015names}, prior to comparing our method against it.
The taxonomy of $2 \times 2$ games proposed by \citet{bruns2015names} relies on two elements: 1) a topological ordering of said games, according to the patterns of payoffs received by each player; 2) classification of payoff families, based on the respective payoffs received by each player at the Nash equilibria. 
Based on the earlier work of \citet{robinson2005topology}, \citet{bruns2015names} identifies 12 core payoff patterns in ordinal $2 \times 2$ games (which we summarize for the row player in \cref{fig:2x2_payoffs_bruns}).
Column player payoffs are defined in \citet{bruns2015names} by transposing the row player payoffs along the {anti}-diagonal (i.e., not the conventional transpose). 
The combinations of row and column player payoffs yields a total of 144 $2 \times 2$ games of interest.
These games are subsequently categorized according to the relative payoffs received by the players in the Nash equilibria (e.g., win-win, cyclic, unfair outcomes, etc.), yielding a total of 7 classes (reproduced here in \cref{fig:2x2_taxonomy_bruns}).
Note that the 4 quadrants indicated by the thick blue lines in \cref{fig:2x2_taxonomy_bruns} indicate so-called layers of 36 games each, where games in each layer have the same highest-payoff relationship between the two players.
For example, games in the lower-left layer are all win-win (thus have the highest payoff for both players in the same payoff cell), whereas games in the top-right layer have the highest payoffs in diagonally-opposite payoff cells for the players.

To generate a clustering of these games using our approach, we run the graph-based analysis detailed in the paper (i.e., computing response graphs, collecting graph measures, and running PCA on the collection of 144 games). 
We subsequently cluster these games using their top-2 principal components, using 7 clusters (as in \citet{bruns2015names}). 
The resulting clusters are visualized in \cref{fig:2x2_taxonomy_ours_alpha_0p2}.
Note that as our clusters are automatically discovered, and not hand-specified using Nash equilibrium payoffs, only their structure (and not color) should be compared against Bruns' clusters.
While some similarities exist between these and Bruns' clusters (e.g., a visually-apparent division into the same 4 layers is present in our clustering, and certain games clusters such as ($Ch \times Co$, $Ch \times As$, $Ba \times Co$, $Ba \times Sh$, $Hr \times As$, $Hr \times Sh$) are shared between the two methods), notable differences are present.

To better understand these differences, consider again the means by which the Bruns clusters are derived: manual identification of patterns over the players' payoffs in the Nash equilibria of each game.
Our approach, by contrast, relies on aggregation of statistics over the response graph of the games.
As $2 \times 2$ games consist of 4 strategy profiles in total, the response graphs correspondingly have 4 nodes, implying that few variations are possible in their structure (in contrast to the much larger games targeted in our main results).
Despite this, we can better tune our approach for these extremely small games. 
Specifically, one of the statistics used for computing our clustering is the \alpharank distribution entropy associated with the response graph.
Typically, \alpharank is used with a high value of selection-intensity parameter, $\alpha$, to increase the distribution mass over strong strategies.
In these small $2 \times 2$ games, decreasing the value of \alpharank's selection-intensity parameter, $\alpha$, causes the \alpharank distribution to increase in entropy, and subsequently better capture the distribution of payoffs received over the profiles (akin to the Nash-based payoff comparisons used by Bruns to derive their clusters).
The results of this modification, which brings our approach closer to that of Bruns, is visualized in \cref{fig:2x2_taxonomy_ours_alpha_0p01}.
Notable similarities between our clusters and Bruns' are evident in this updated approach.
Namely, variations of prisoner's dilemma (Pd) are clustered in our approach and theirs: ($Pd \times \{Co, As, Sh, Pd\}$ (and their transposed analogs).
The cyclic game cluster of $\{Ch, Ba, Hr\} \times \{Co, As, Sh\}$ (and their transposed analogs) are identically clustered in the approaches.

Note also that due to the Nash-payoff based classification scheme of Bruns, variations in the lower-left quadrant (where all Nash equilibria are win-win) are not evident in their approach (\cref{fig:2x2_payoffs_bruns}).
By contrast, our approach highlights differences between these games.
For example, whereas Bruns classifies the symmetric Coordination Game ($Co \times Co$) in the same class as Stag Hunt ($Sh \times Sh$), ours places it in its own class altogether (better matching intuition, due to the anti-coordination outcomes of Stag Hunt not being present in the coordination game).

With these comparisons in place, we further note several benefits of our approach compared to that of \citet{bruns2015names} and related works on classification of $2 \times 2$ games.
First, our approach does not rely on a hand-crafted taxonomy or classification of games;
our games classes are identified automatically, and without need for human expertise in determination of patterns in payoffs.
Second, as the taxonomy introduced by Bruns' relies on enumeration of payoff orderings, it faces significant scalability issues if to be extended to larger games (involving more strategies and/or players).
By contrast, our approach applies directly to all normal form games, and we are not aware of similar automated classification schemes for larger, many-player games.
Moreover, whereas Bruns approach provides a hard classification based on hard-coded structural patterns in payoffs, ours provides a soft classification based on a similarity metric (specifically, based on the games' principal components).
Thus, our approach can be used to compute distances between games, in contrast to Bruns' and related approaches.
We visualize several examples of $2 \times 2$ such game distances in \cref{fig:2x2_taxonomy_ours_alpha_0p01_distance_Co_Co,fig:2x2_taxonomy_ours_alpha_0p01_distance_Ch_Ch,fig:2x2_taxonomy_ours_alpha_0p01_distance_Pd_Pd}.
A subsequent benefit of being able to compute distances between games is that it enables computation of a dendrogram over a collection of games.
We use our approach to visualize this tree for Bruns' $2 \times 2$ games in  \cref{fig:supp_2x2_games_phylo_tree}.
Finally, the downstream usefulness of our approach is that such a similarity metric can be optimized to generate new games (as illustrated in the main text), in contrast to the hard classification scheme of Bruns.

\subsection*{Sensitivity analysis}

\begin{figure}
    \centering
    \begin{subfigure}{\textwidth}
        \centering
        \caption{}
        \includegraphics[width=0.5\textwidth]{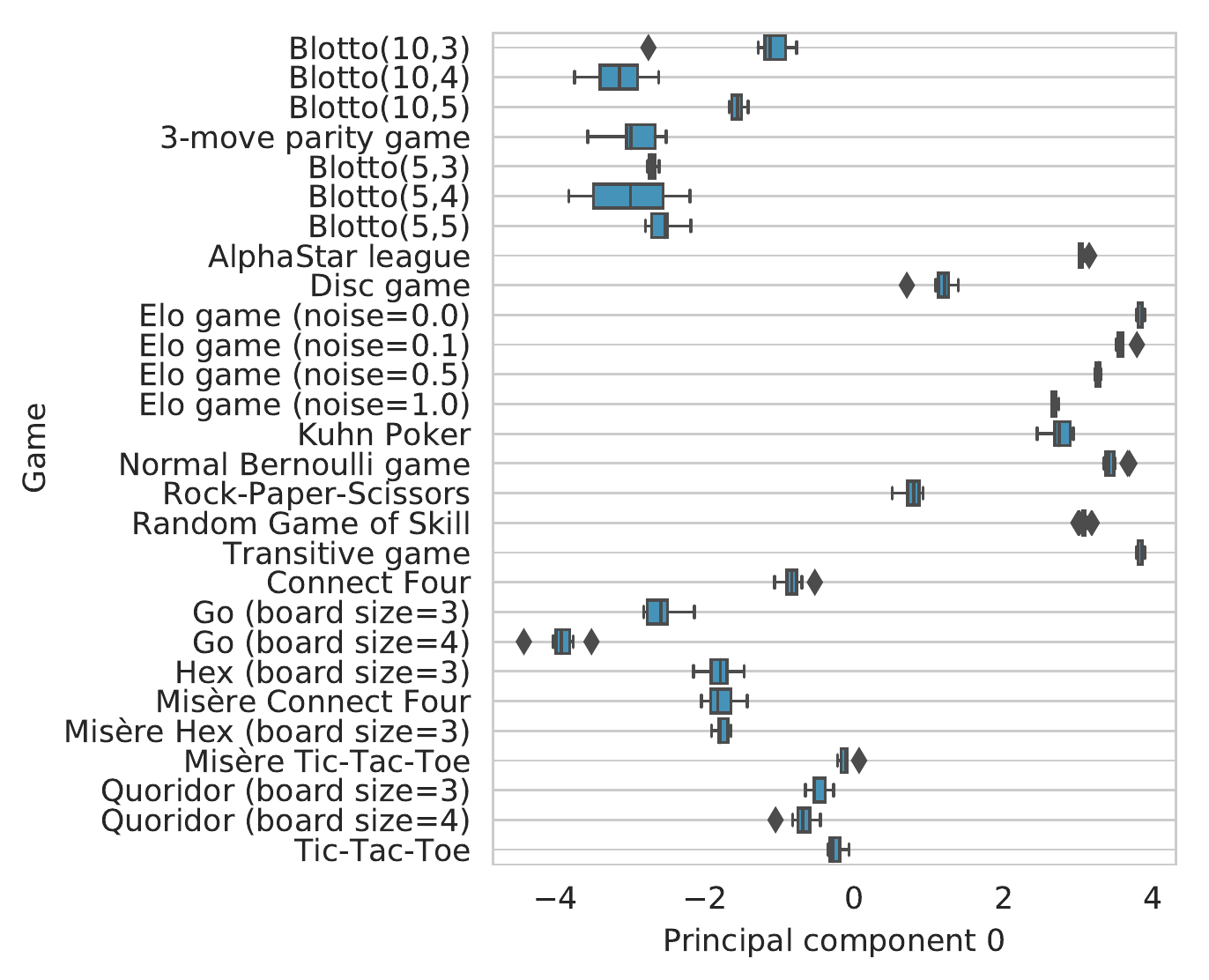}%
        \includegraphics[width=0.5\textwidth]{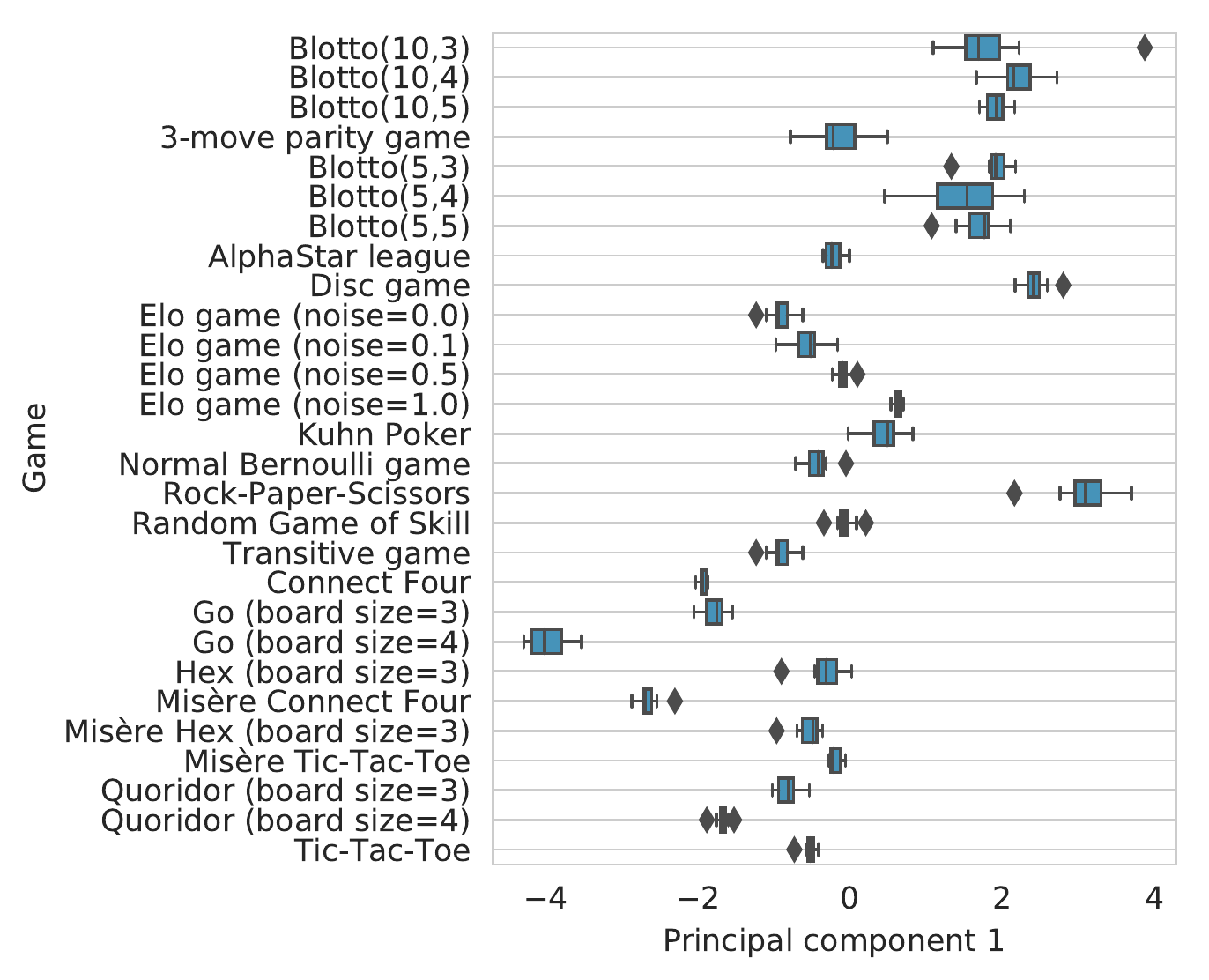}
        \label{fig:bootstrap_pc}
    \end{subfigure}
    \begin{subfigure}{0.5\textwidth}
        \centering
        \caption{}
        \includegraphics[width=\textwidth]{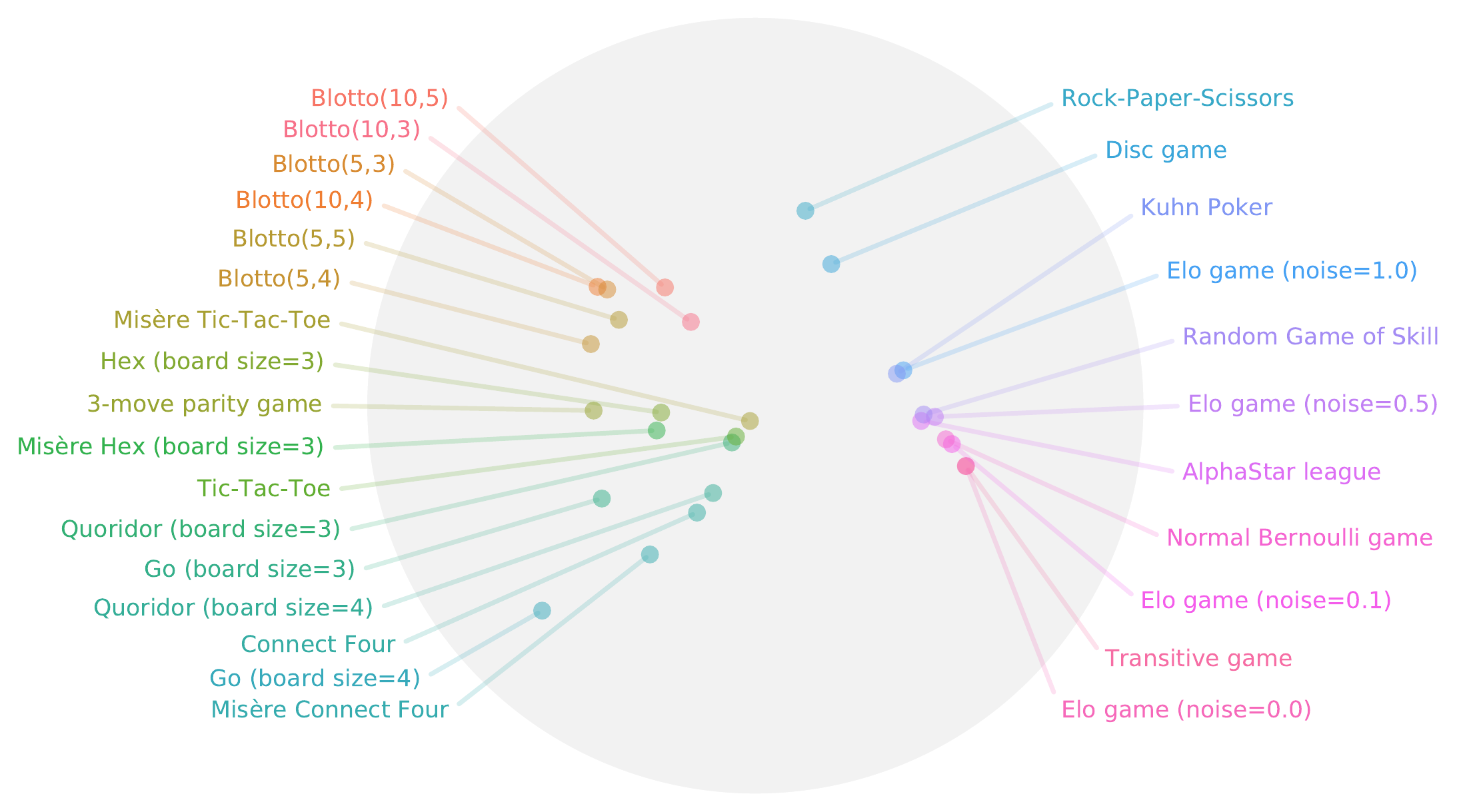}\\
        \label{fig:bootstrap_embeddings_trial_0}
    \end{subfigure}%
    \begin{subfigure}{0.5\textwidth}
        \centering
        \caption{}
        \includegraphics[width=\textwidth]{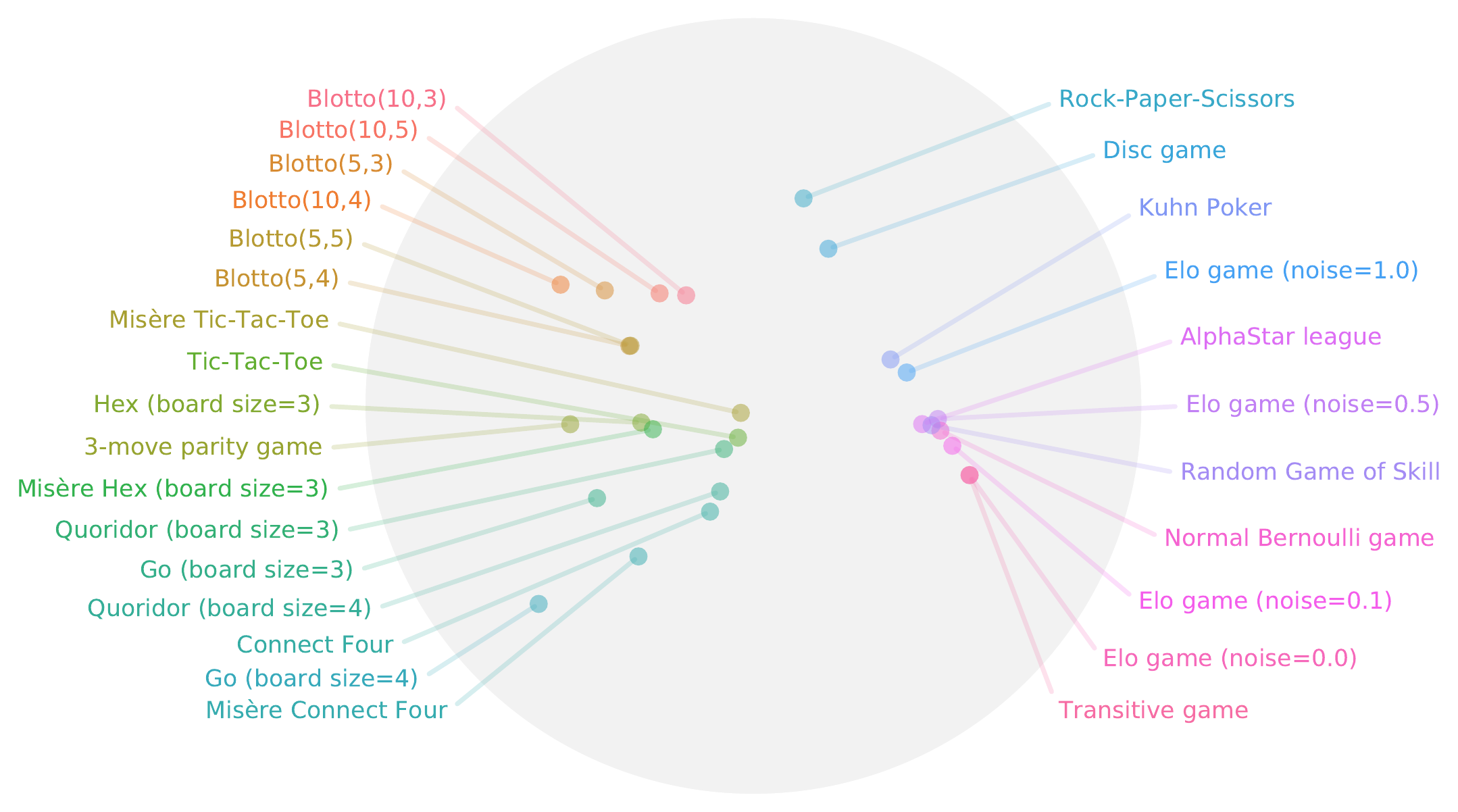}\\
        \label{fig:bootstrap_embeddings_trial_1}
    \end{subfigure}\\
    \begin{subfigure}{0.5\textwidth}
        \centering
        \caption{}
        \includegraphics[width=\textwidth]{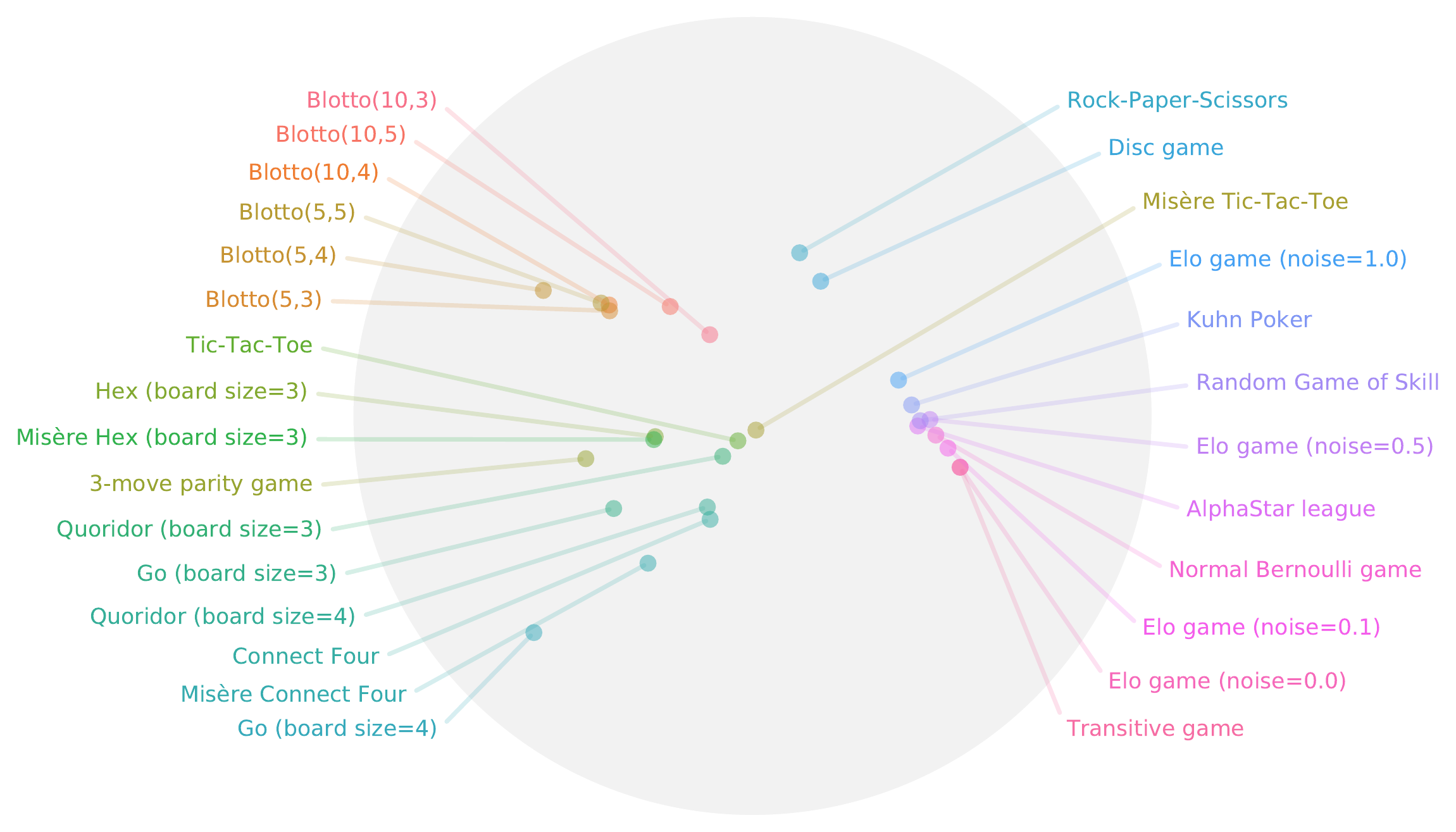}\\
        \label{fig:bootstrap_embeddings_trial_2}
    \end{subfigure}%
    \begin{subfigure}{0.5\textwidth}
        \centering
        \caption{}
        \includegraphics[width=\textwidth]{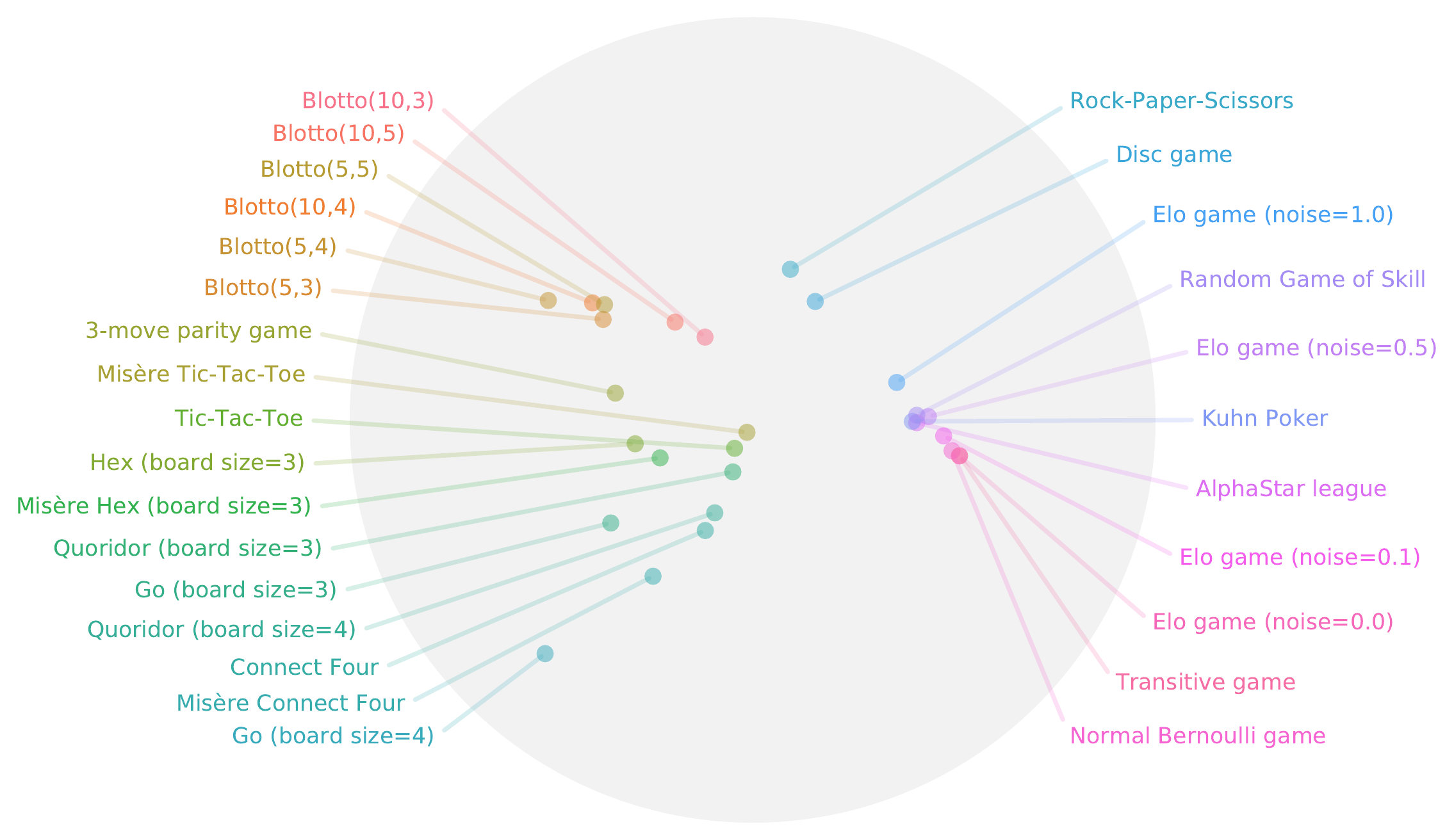}\\
        \label{fig:bootstrap_embeddings_trial_3}
    \end{subfigure}\\
    \caption{Sensitivity to choice of empirical strategies, via subsampling of policy space. For each empirical game, we randomly subsample 50\% of the policies (discarding the rest). We subsequently run the full analysis pipeline. 10 trials of subsampling are conducted per game (with the exception of Rock--Paper--Scissors, where we do not subsample strategies/policies due to it being a small canonical game). \subref{fig:bootstrap_pc} indicates the quartiles of the first two principal components per game, over all trials.
    Specifically, the distribution of games' first two principal components under 10 trials of policy subsampling are illustrated, showing sensitivity to policy space changes. Boxplot elements are defined as follows: the box visualizes quartiles; whisker bars show data variability beyond the quartiles; the median is indicated by the center line; diamonds indicate outliers.
    \subref{fig:bootstrap_embeddings_trial_0}--\subref{fig:bootstrap_embeddings_trial_3} shows the landscape of games for 4 example trials.
    Note that game colors are kept the same as the original landscape of games visual (\cref{fig:rwg_embeddings_normCoG_True}) for easier comparison.
    }
    \label{fig:sensitivity_analysis}
\end{figure}

As mentioned in the main text, the empirical game-theoretic results rely on sampling of a representative set of policies to characterize the underlying games. 
An important limitation here is that the the empirical game-theoretic results are subject to the policies used to generate them.
Here we conduct a set of experiments to test the sensitivity of the results to these sampled policies.

For each game (with the exception of Rock--Paper--Scissors, which is a canonical game and only 3 $\times$ 3 in size), we randomly subsample 50\% of the policies.
All other policies are discarded, thus yielding an empirical payoff matrix 4 times smaller in overall size.
We subsequently run the full analysis pipeline over 10 trials of subsampling per game (see \cref{fig:sensitivity_analysis} for results). 
\Cref{fig:bootstrap_pc} indicates the quartiles of the first two principal components per game, summarized over all trials.
Despite 50\% of the policies being randomly discarded, the overall statistics are relatively robust across the different trials.
There is some sensitivity of the principal components to the empirical policies sampled in the process, which on closer inspection seems to occur for highly cyclical games such as Blotto variations and the Disc game. 
We additionally recompute the Landscape of Games figure, independently for several of the subsampling trials, with 4 examples shown in \cref{fig:bootstrap_embeddings_trial_0,fig:bootstrap_embeddings_trial_1,fig:bootstrap_embeddings_trial_2,fig:bootstrap_embeddings_trial_3}.
Note that while there are some quantitative differences (in terms of relative positioning of these games in the projected space), the overall trends and clusters are quite robust to the subsampling (compared to Figure 1 of the main paper, which shows the landscape generated using all policies).

Overall, this analysis seems to indicate that the combination of our policy sampling scheme and analysis pipeline produce robust results, though as mentioned in the Discussions section of the main text, this will be a useful aspect of the method to revisit when experimenting on significantly larger games.

\subsection*{Impact of normalization on complexity results}\label{append:ablative}

\begin{figure}[t!]
    \centering
    \includegraphics[width=\textwidth]{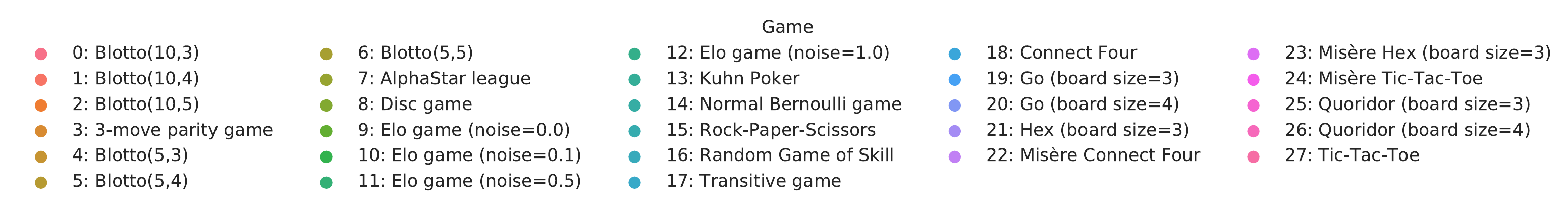}
    \begin{subfigure}[t]{\textwidth}
        \centering
        \caption{}
        \includegraphics[width=0.33\textwidth]{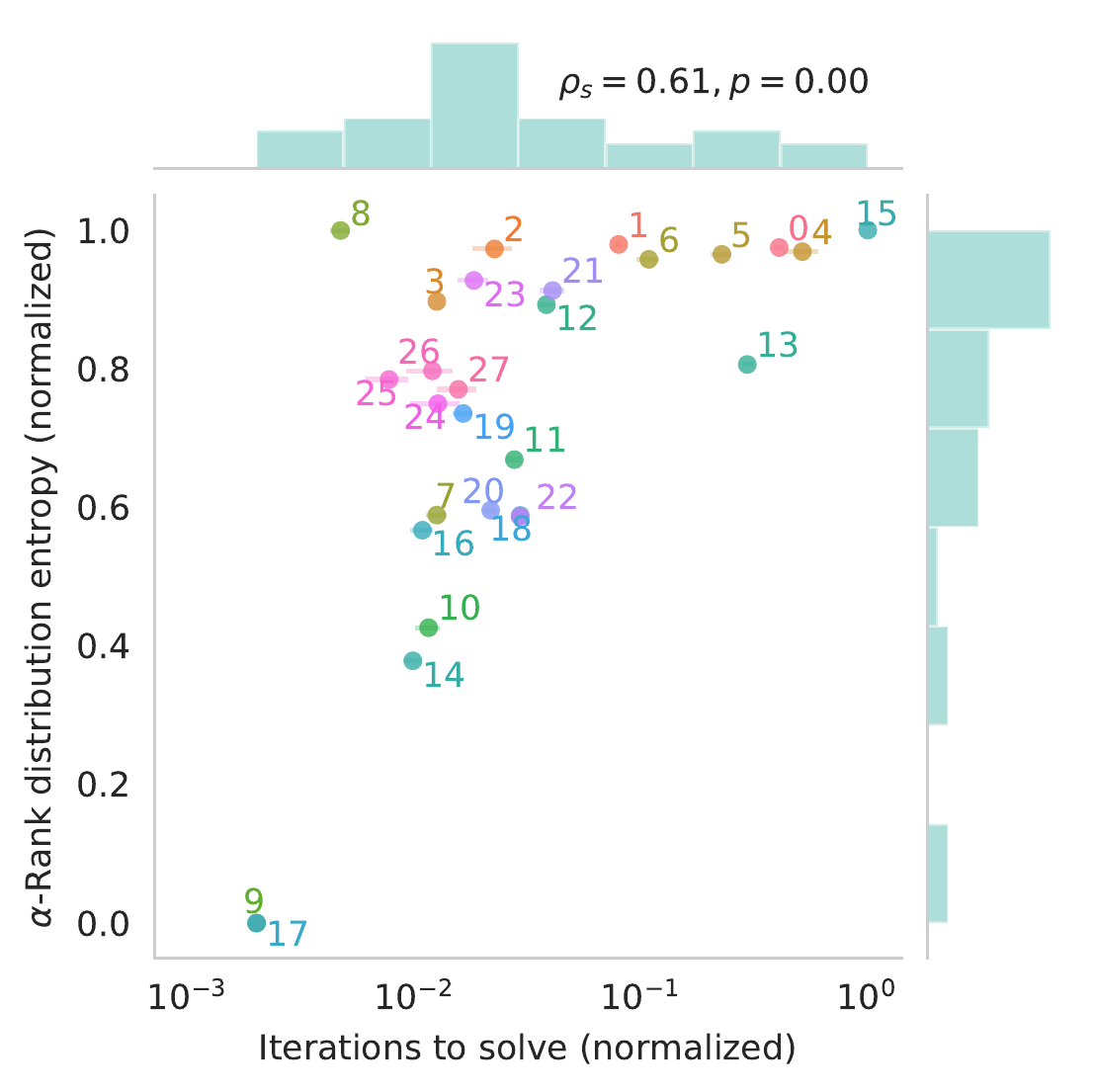}%
        \hfill%
        \includegraphics[width=0.33\textwidth]{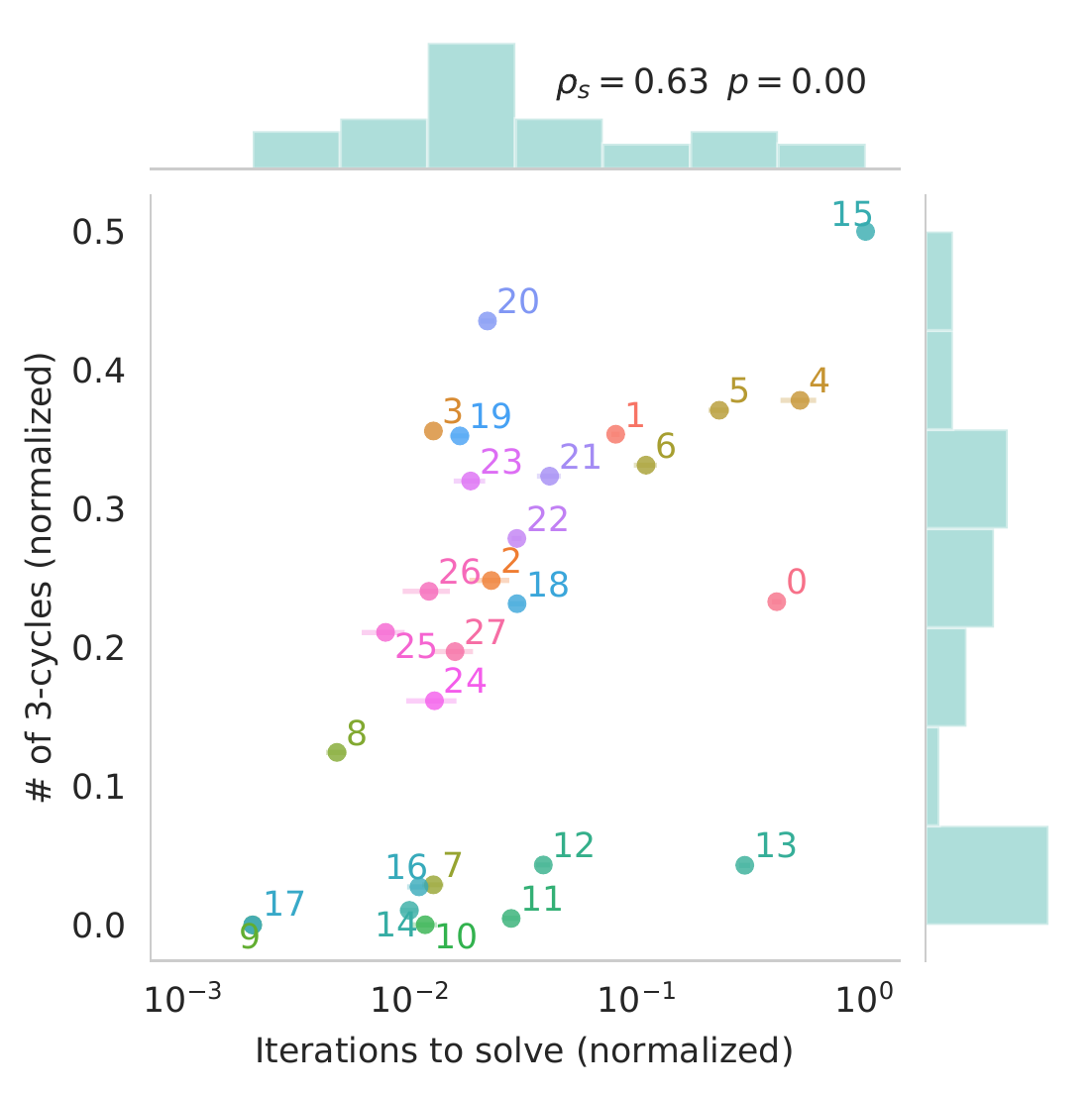}
        \hfill%
        \includegraphics[width=0.33\textwidth]{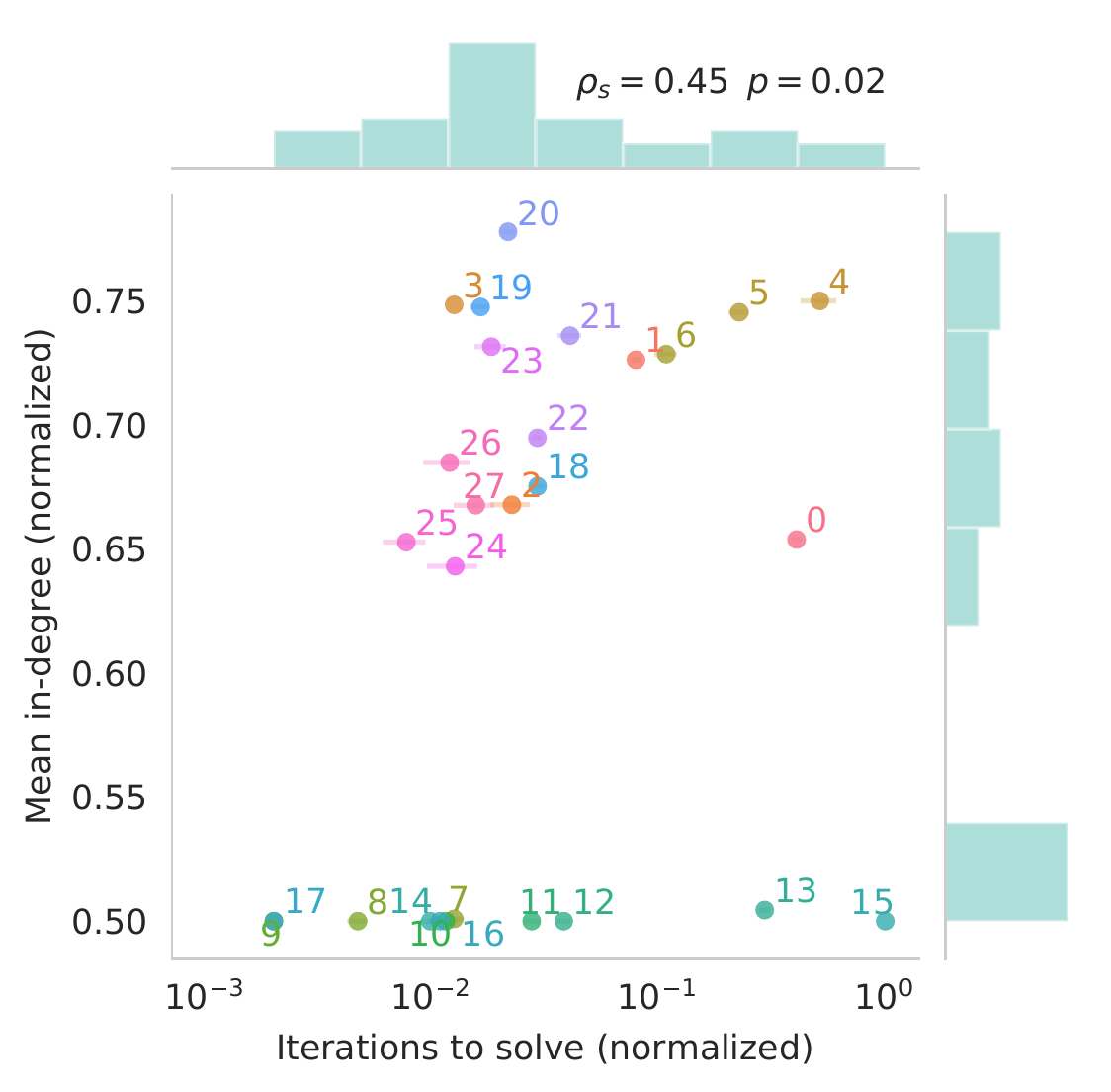}%
        \label{fig:double_oracle_normDO_True_normCoG_True}
    \end{subfigure}\\
    \begin{subfigure}[t]{\textwidth}
        \centering
        \caption{}
        \includegraphics[width=0.33\textwidth]{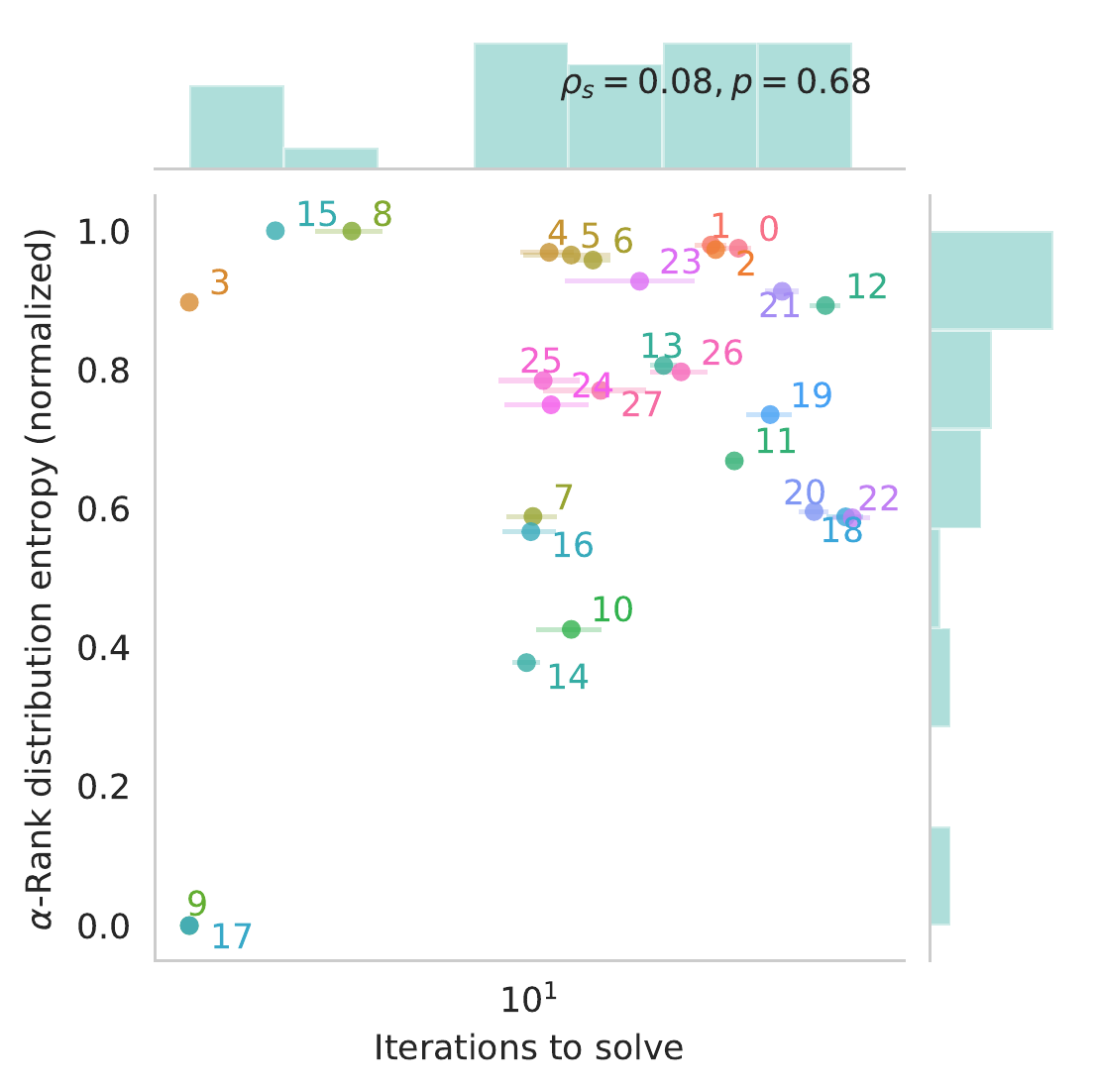}%
        \hfill%
        \includegraphics[width=0.33\textwidth]{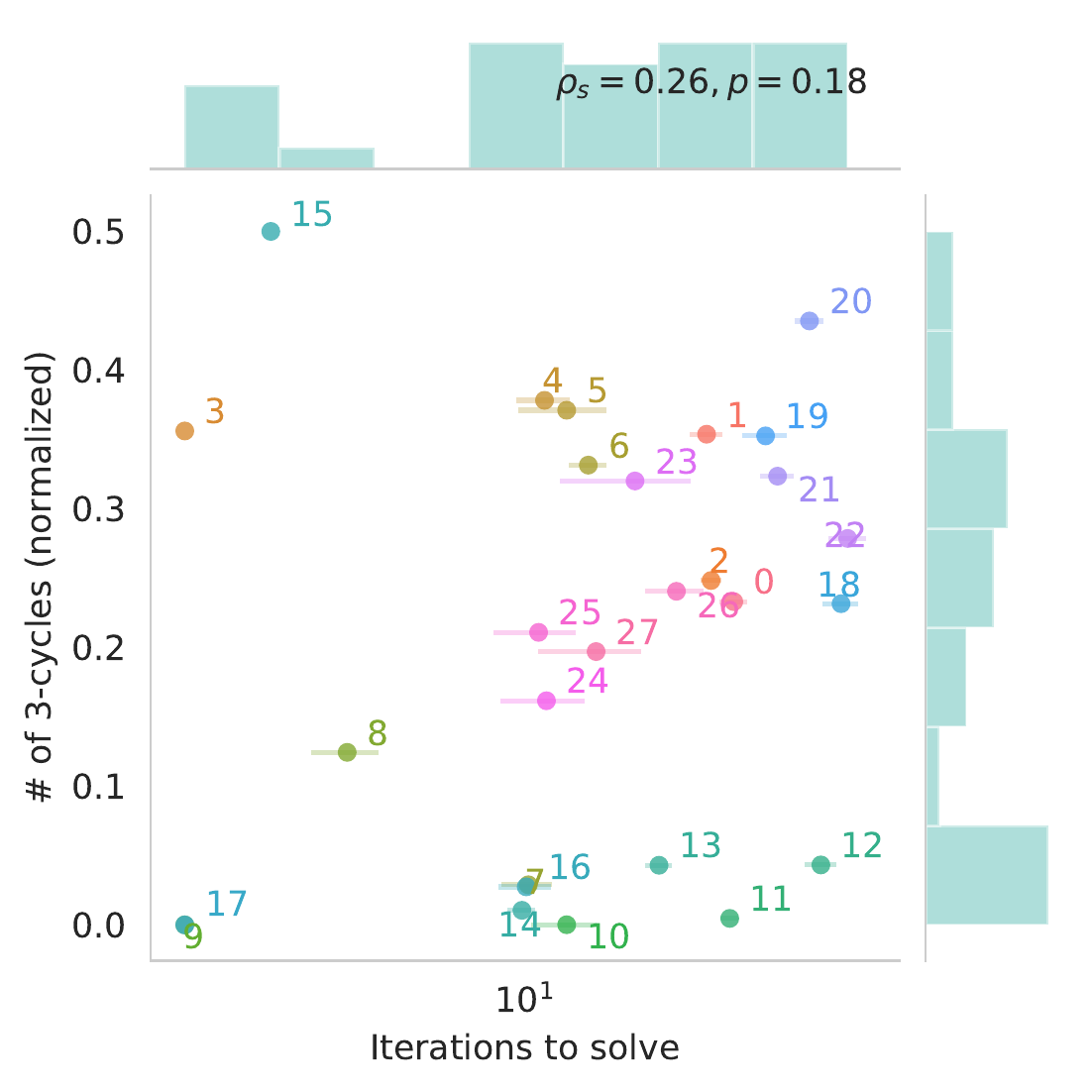}
        \hfill%
        \includegraphics[width=0.33\textwidth]{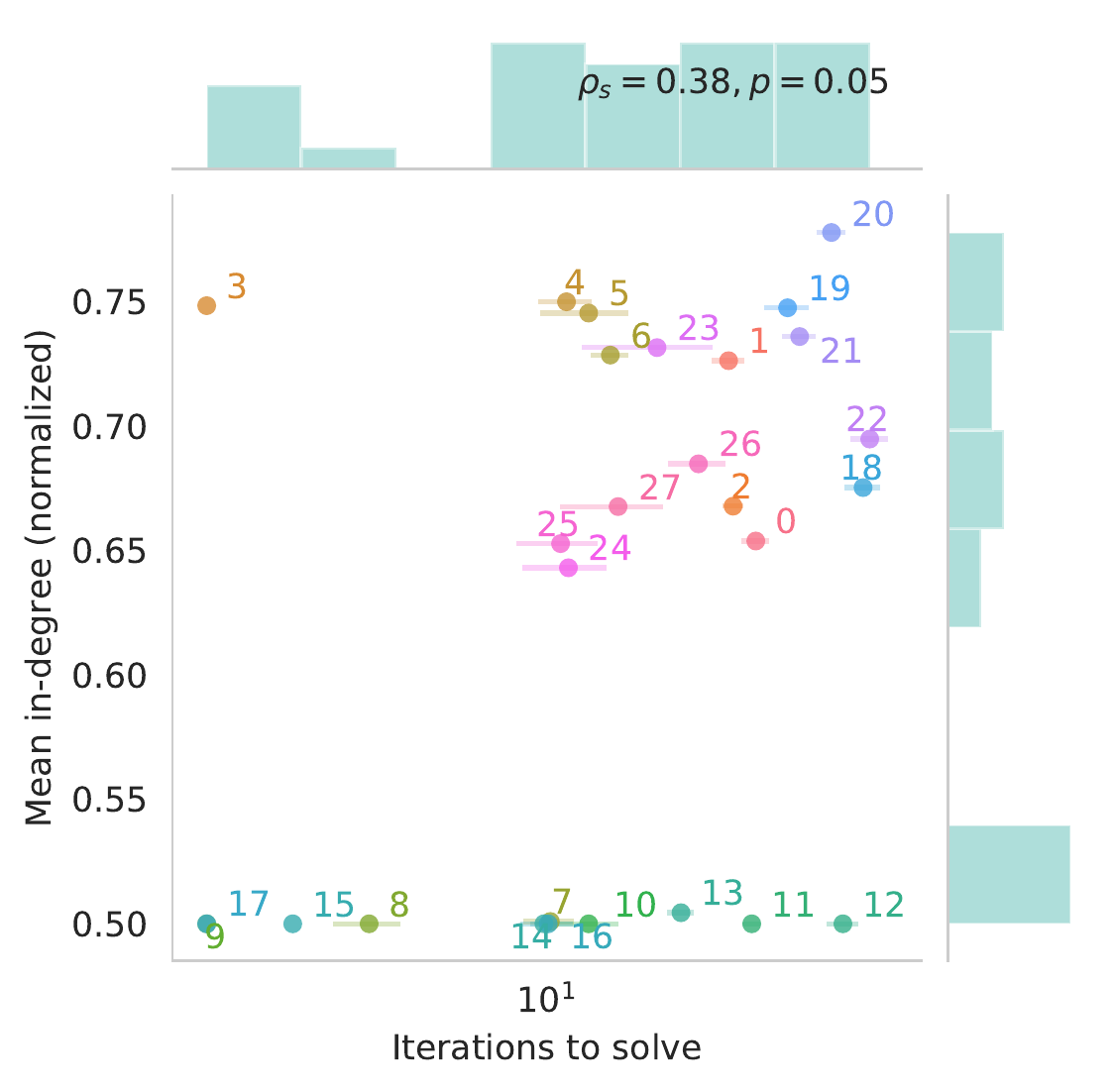}%
        \label{fig:double_oracle_normDO_False_normCoG_True}
    \end{subfigure}
    \caption{Response graph complexity vs. computational complexity of solving various games. Each column plots a respective measure of graph complexity against the number of iterations needed to solve the associated game via the double oracle algorithm. 
    In both rows, the graph complexity measures are normalized.
    \subref{fig:double_oracle_normDO_True_normCoG_True} shows results where the number of double oracle iterations is also normalized.
    \subref{fig:double_oracle_normDO_False_normCoG_True} shows results where the number of double oracle iterations is not normalized.
    }
    \label{fig:ablative_graph_vs_computational_complexity}
\end{figure}
    
\begin{figure}[t!]
    \centering
    \includegraphics[width=\textwidth]{figs/double_oracle/double_oracle_game_results_legend.pdf}
    \begin{subfigure}[t]{\textwidth}
        \centering
        \caption{}
        \includegraphics[width=0.33\textwidth]{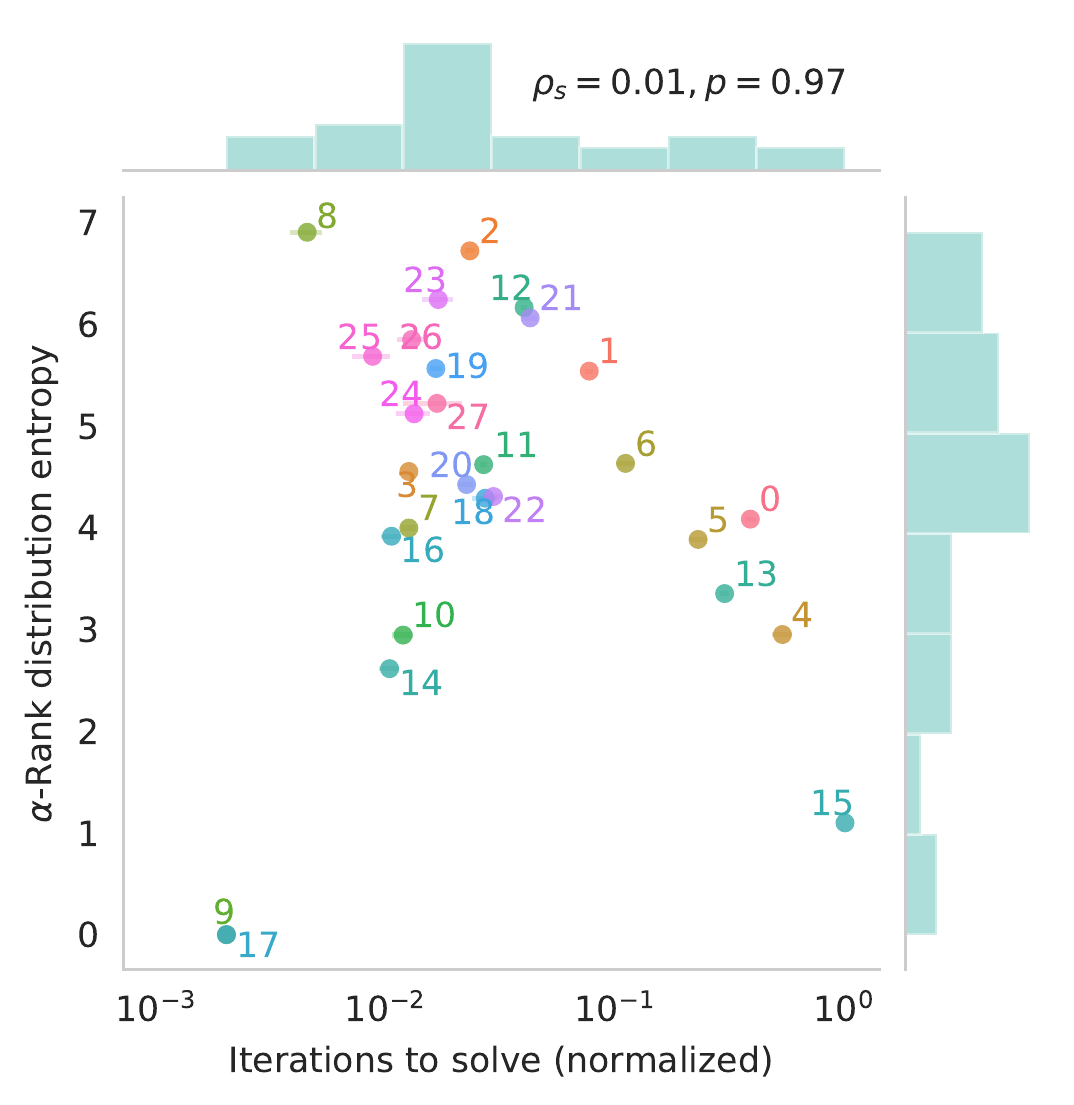}%
        \hfill%
        \includegraphics[width=0.33\textwidth]{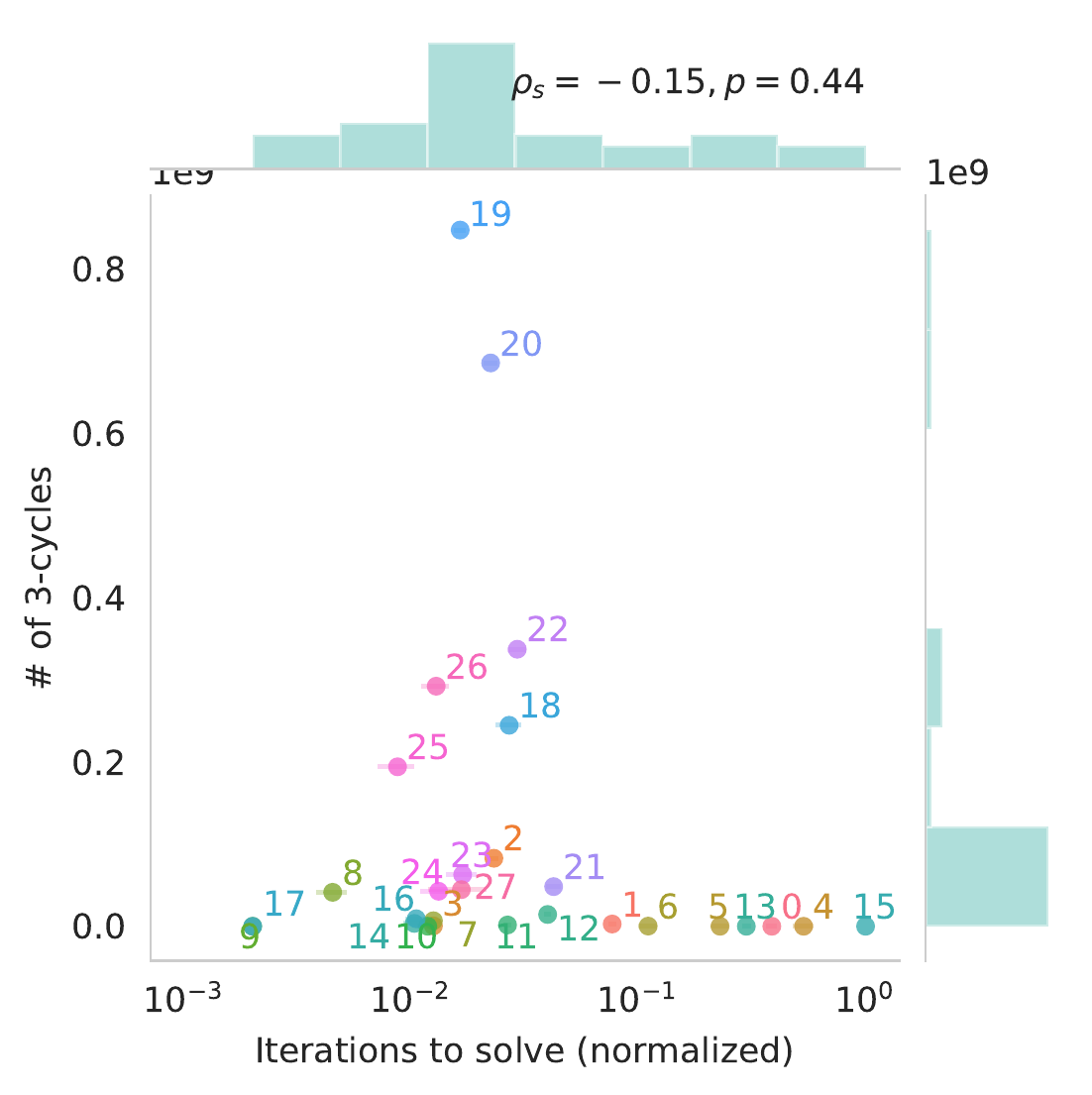}
        \hfill%
        \includegraphics[width=0.33\textwidth]{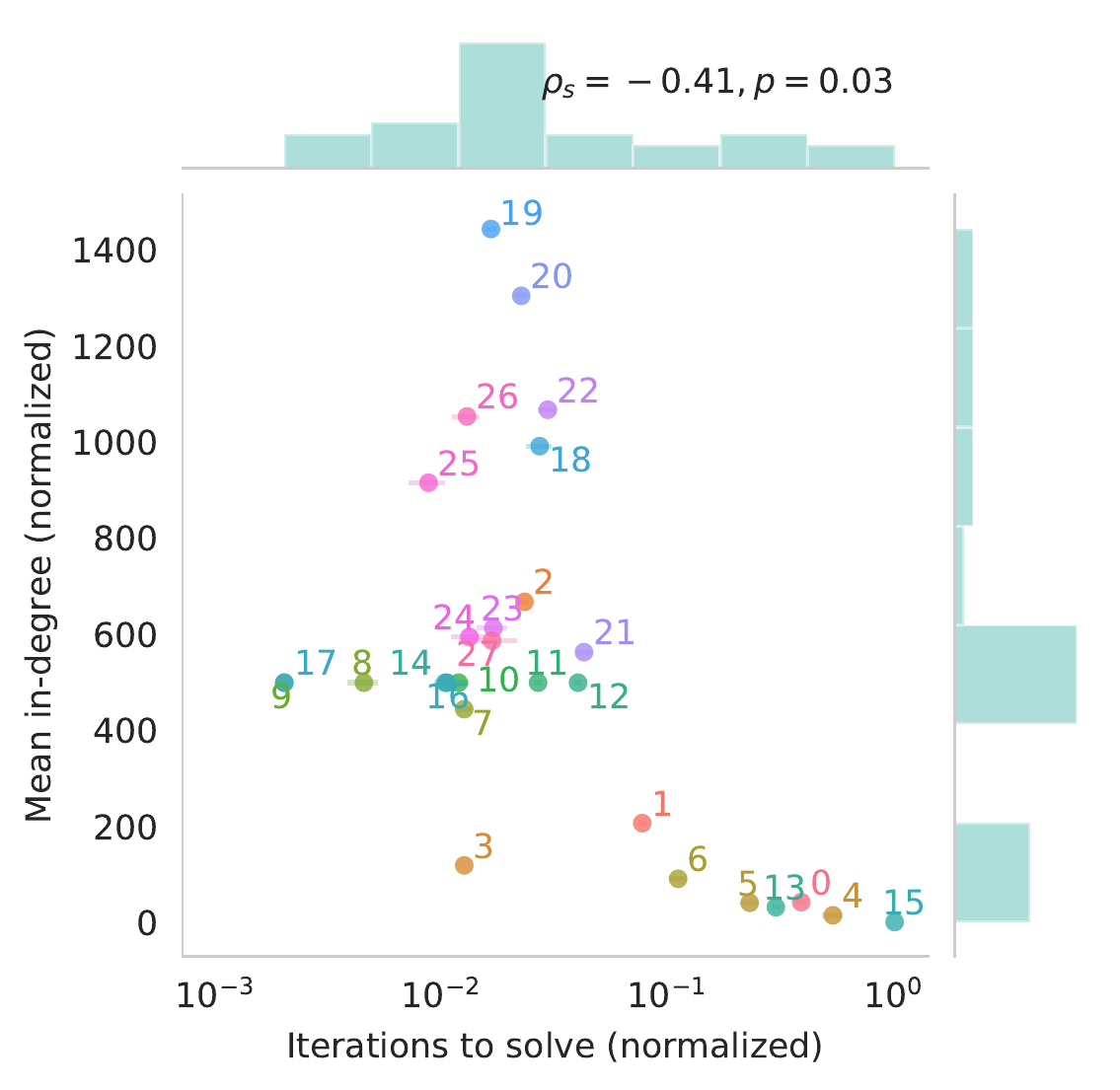}%
        \label{fig:double_oracle_normDO_True_normCoG_False}
    \end{subfigure}\\
    \begin{subfigure}[t]{\textwidth}
        \centering
        \caption{}
        \includegraphics[width=0.33\textwidth]{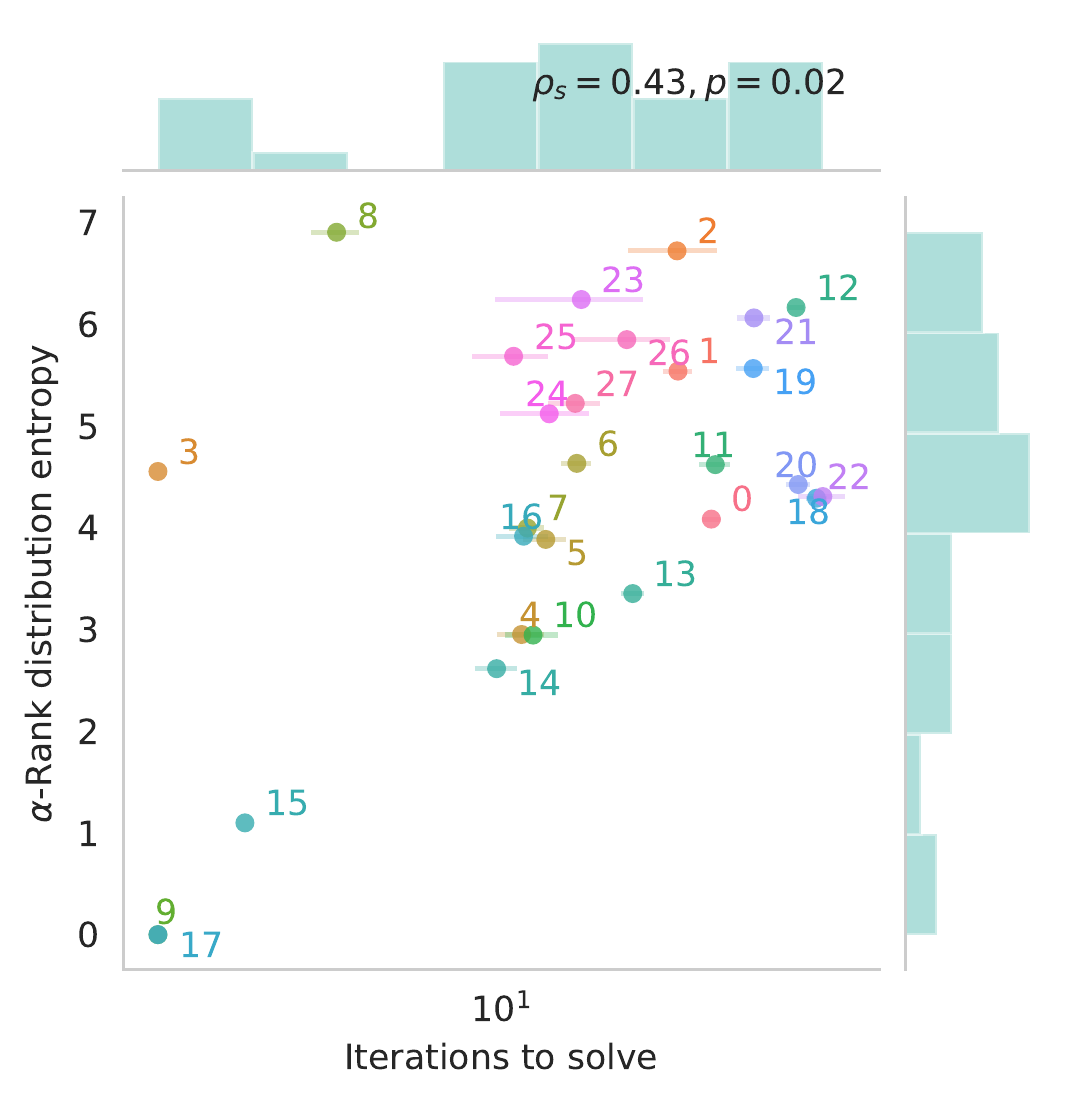}%
        \hfill%
        \includegraphics[width=0.33\textwidth]{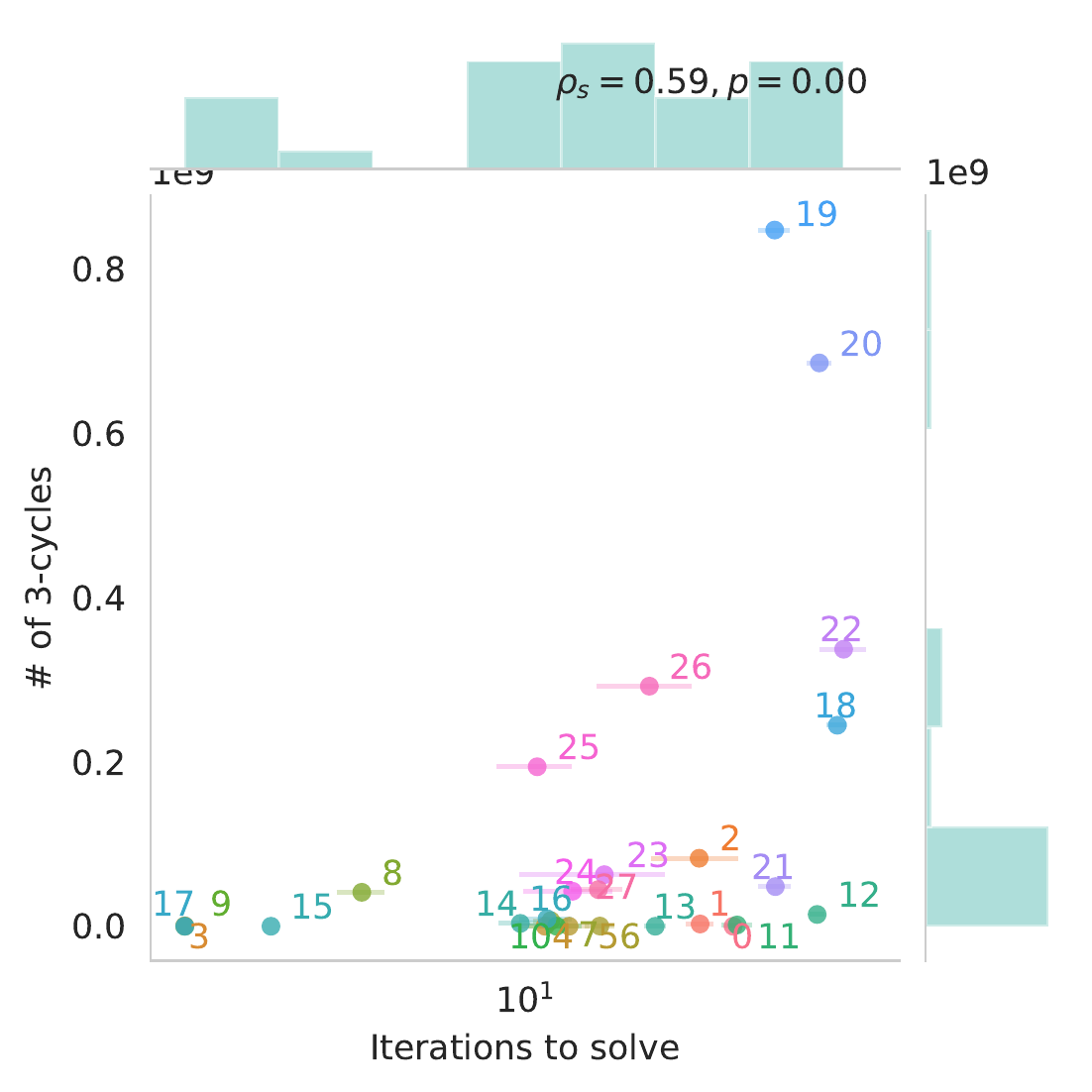}
        \hfill%
        \includegraphics[width=0.33\textwidth]{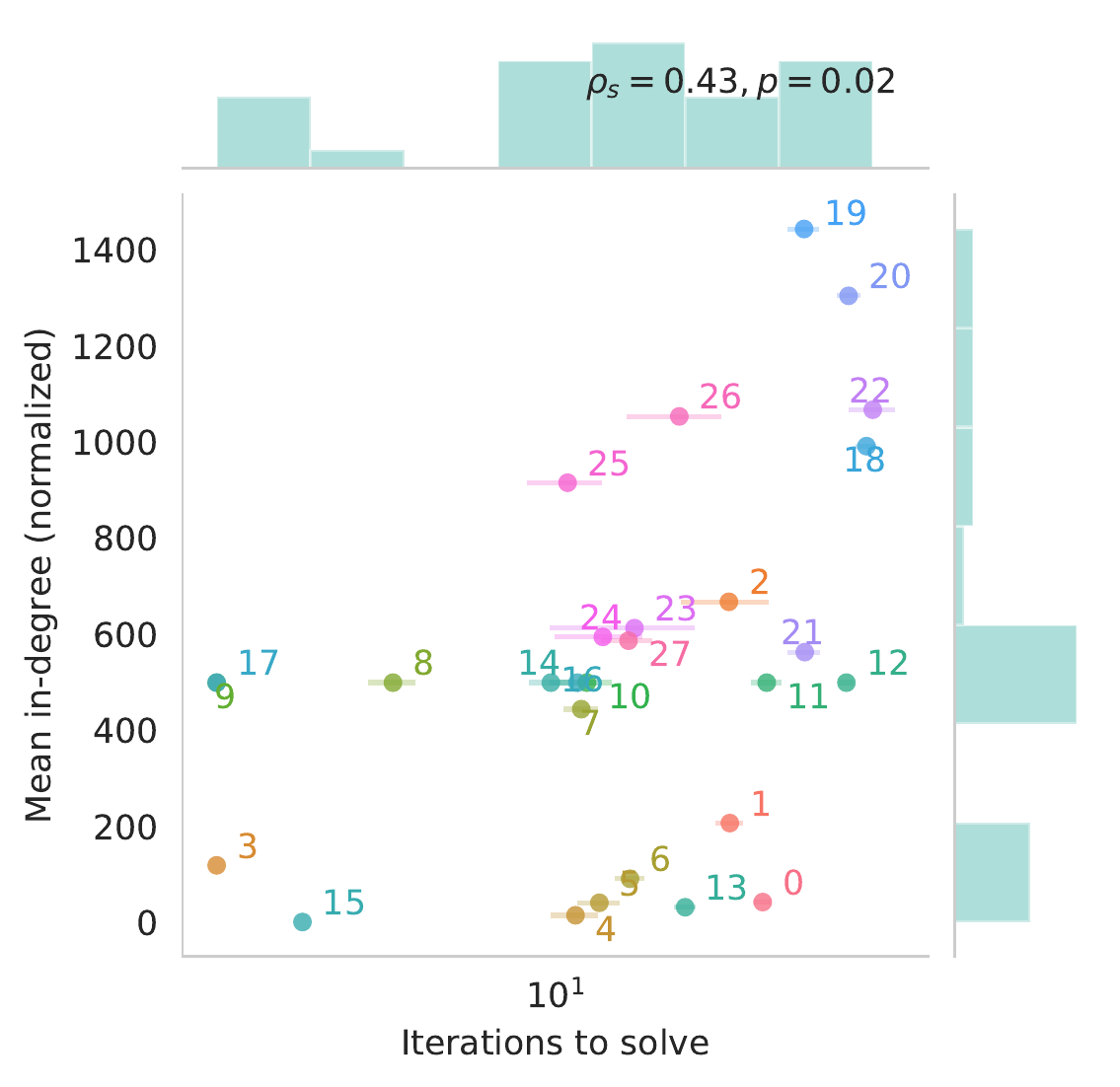}%
        \label{fig:double_oracle_normDO_False_normCoG_False}
    \end{subfigure}%
    \caption{Response graph complexity vs. computational complexity of solving various games. Each column plots a respective measure of graph complexity against the number of iterations needed to solve the associated game via the double oracle algorithm. 
    In both rows, the graph complexity measures are not normalized.
    \subref{fig:double_oracle_normDO_True_normCoG_False} shows results where the number of double oracle iterations is normalized.
    \subref{fig:double_oracle_normDO_False_normCoG_False} shows results where the number of double oracle iterations is not normalized.
    }
    \label{fig:ablative_graph_vs_computational_complexity_cog_unnorm}
\end{figure}

\begin{figure}[!htbp]
	\centering
    \begin{subfigure}{\textwidth}
        \centering
    	\caption{}
    	\includegraphics[width=0.8\textwidth]{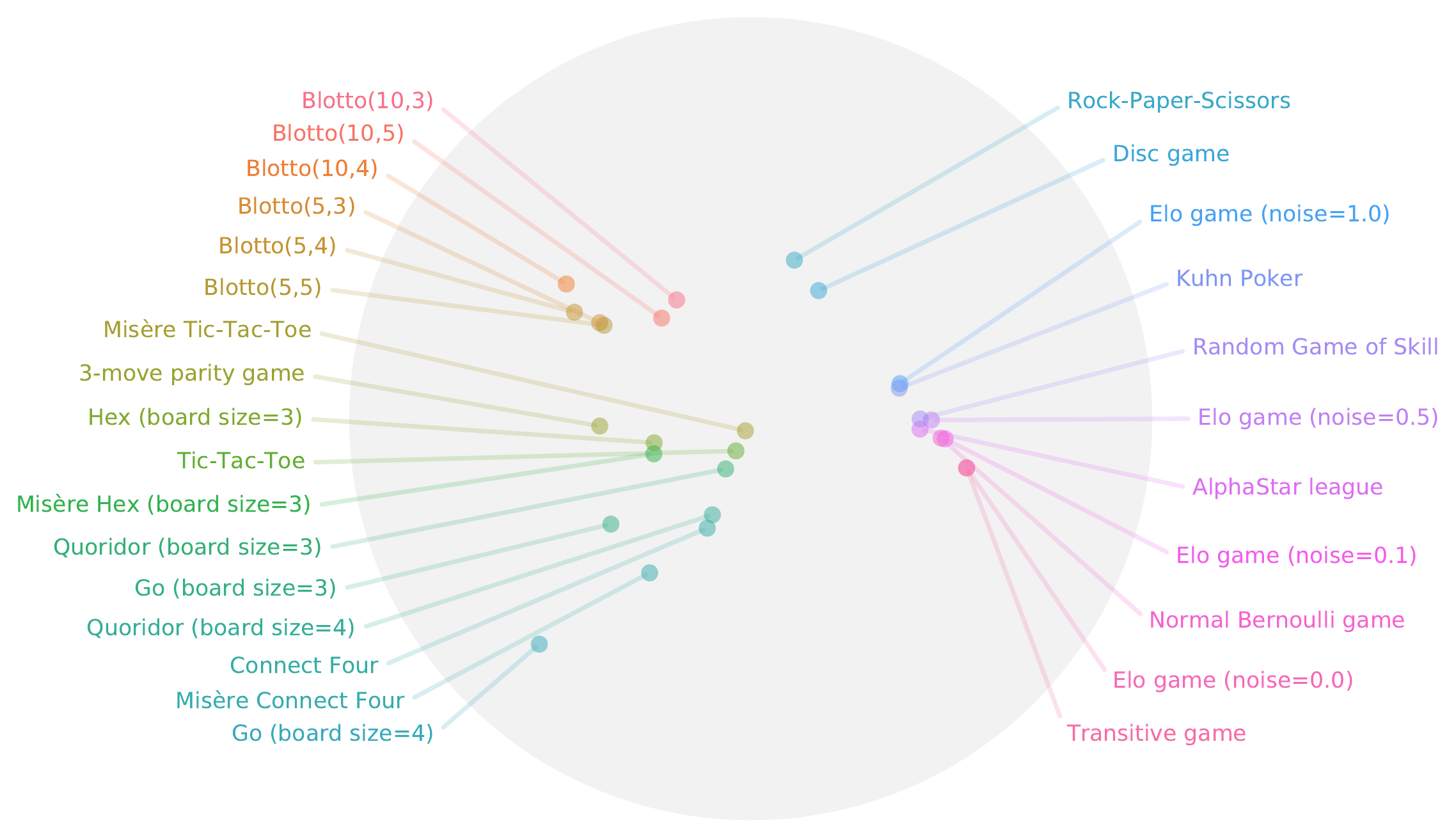}
    	\label{fig:rwg_embeddings_normCoG_True}
    \end{subfigure}
    \begin{subfigure}{\textwidth}
        \centering
    	\caption{}
    	\includegraphics[width=0.8\textwidth]{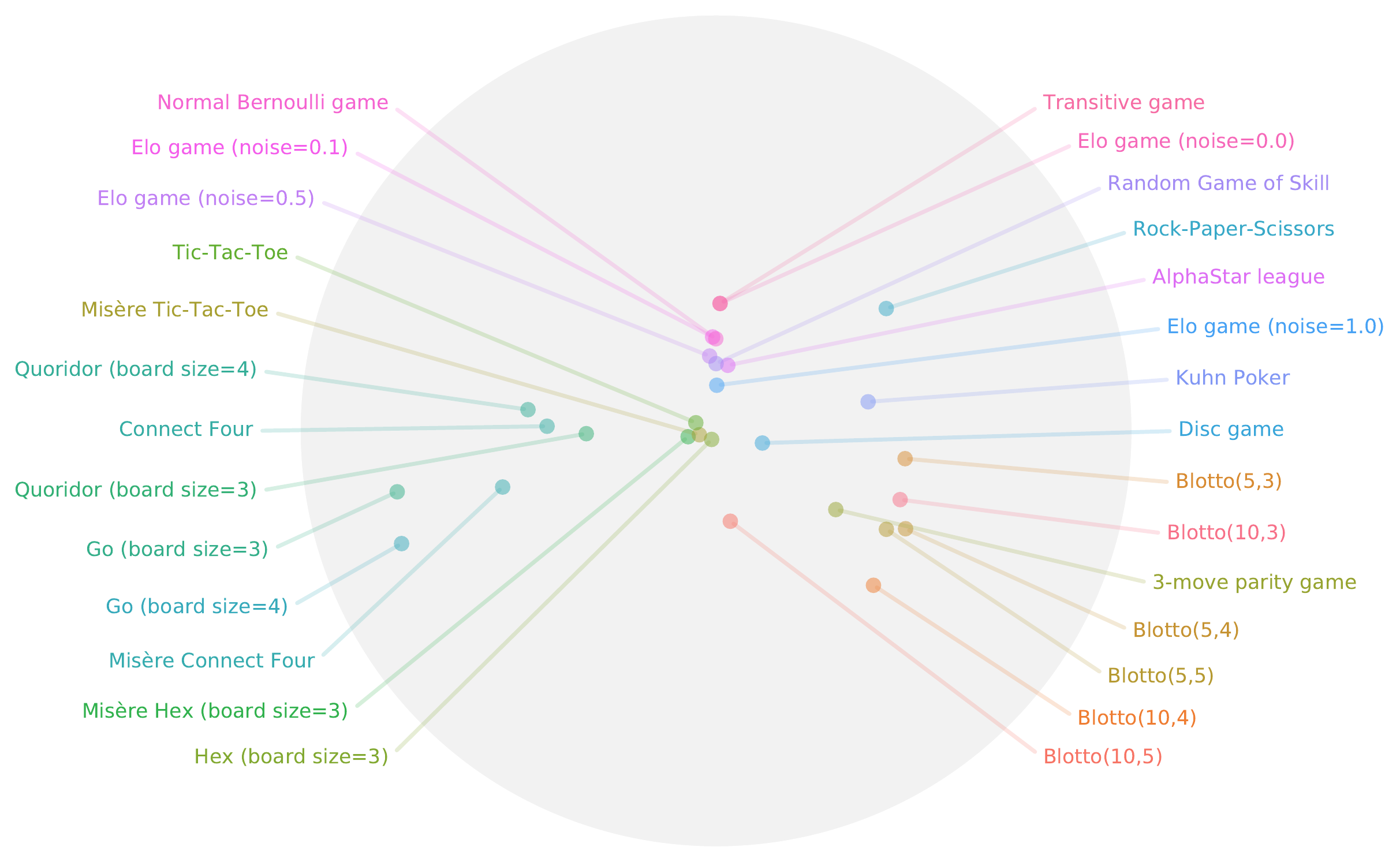}
    	\label{fig:rwg_embeddings_normCoG_False}
    \end{subfigure}
	\caption{Landscape of games generated via normalized and unnormalized response graph measures.
	\subref{fig:rwg_embeddings_normCoG_True} Normalized graph measures.
	\subref{fig:rwg_embeddings_normCoG_False} Unnormalized graph measures.
	}
	\label{fig:ablative_rwg_embeddings}
\end{figure}

We conduct additional ablative analysis of the transformations of graph complexity measures considered in the main text.
As noted in the main text, these results are not intended to to propose that a specific definition of complexity (e.g., with respect to Nash) is explicitly useful for defining a topology / classification over games, but merely investigate correlations between these measures and those related to the raw response graphs.
In \cref{fig:ablative_graph_vs_computational_complexity,fig:ablative_graph_vs_computational_complexity_cog_unnorm}, we plot all complexity results, over all combinations of normalizing/not normalizing the graph measures (y-axes) and/or the number of iterations to solve (x-axes).
Note that the case corresponding to the main paper is captured by \cref{fig:double_oracle_normDO_True_normCoG_True}, where both axes are normalized.
The primary motivation for conducting this normalization (across both axes) is that the measures considered here would otherwise vary with increasing game size, which would make it easy to artificially inflate a game's complexity via arbitrary increase of the strategy space size via filler strategies.

Under this view, even the standard (i.e., non-repeated) version of Rock--Paper--Scissors has high computational complexity, in the sense that the Nash equilibrium requires these learning agents to discover and play the three pure strategies with equal probability.
In other words, these results imply only that RPS is complex within the class of 3 $\times$ 3 games, as agents must {fully} explore the strategy space to equilibriate.
Moreover, all larger variants of RPS be equally as complex under this view, in the sense that their Nash equilibria all have full support;
thus, they would all deterministically require the same number of normalized iterations to solve (regardless of random initialization of the double oracle algorithm).
Without this normalization, this particular notion of the complexity of a game could be inflated by artificially increasing the strategy space size, without affecting the underlying topology or type of strategic interactions needed to solve it.

Note that different (and potentially less useful) conclusions may indeed be drawn by considering ablations over the type of normalization done.
For example, disabling this normalization yields \cref{fig:double_oracle_normDO_False_normCoG_True}, which implies that RPS has similar complexity to a fully Transitive Game (number 17), despite the latter's Nash equilibrium being deterministically found after a single double oracle iteration.
Moreover, in \cref{fig:ablative_graph_vs_computational_complexity,fig:ablative_graph_vs_computational_complexity_cog_unnorm}, we find that normalizing both types of measures consistently maximizes their Spearman correlation (with low p-value, see top-right of each subfigure).

We also consider now the impact that this normalization has on the landscape of games figure (see \cref{fig:ablative_rwg_embeddings}).
Here, note that there are only 2 ablations involving only normalization over the graph measures (as double oracle/iterations to solve are not used in the spectral analysis). 
The normalized landscape (\cref{fig:rwg_embeddings_normCoG_True}), used in the main paper, reveals coherent clusters of related games.
By contrast, the variety of underlying payoff table sizes and associated graph measures implies that the unnormalized landscape (\cref{fig:rwg_embeddings_normCoG_False}) is notably less structured.
For instance, while games such as Rock--Paper--Scissors and the Disc Game are conceptually very similar, their distance is relatively larger in the unnormalized landscape due to their widely different payoff sizes (3 $\times$ 3 vs. 1000 $\times$ 1000, respectively).

\subsection*{Effects of mixed policies}

\begin{figure}[t]
    \begin{subfigure}{0.5\textwidth}
        \centering
        \caption{}
        \includegraphics[width=\textwidth]{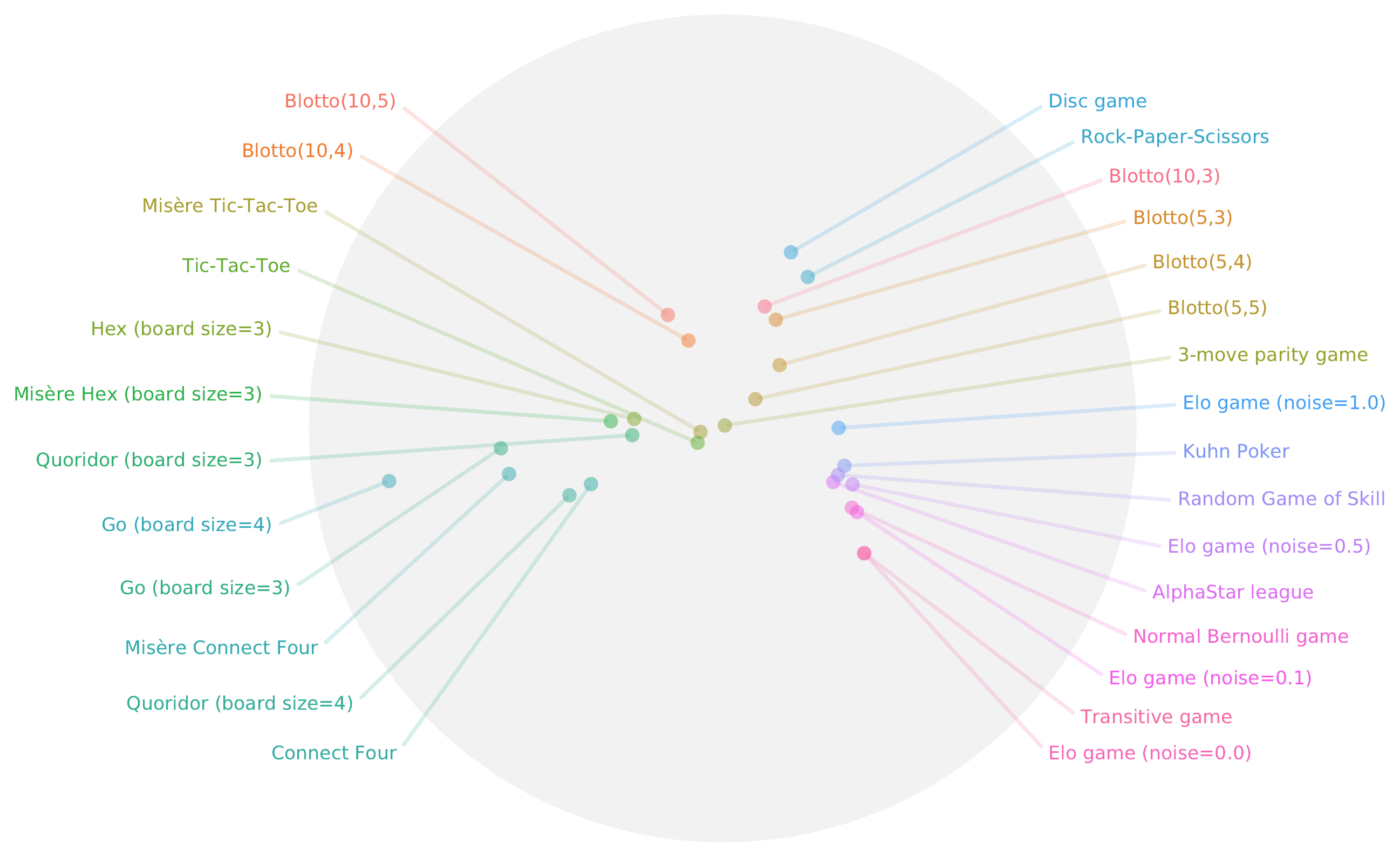}\\
        \label{fig:rwg_embeddings_bootstrap_trial_0_mixpolicies_True_nummix_100}
    \end{subfigure}%
    \begin{subfigure}{0.5\textwidth}
        \centering
        \caption{}
        \includegraphics[width=\textwidth]{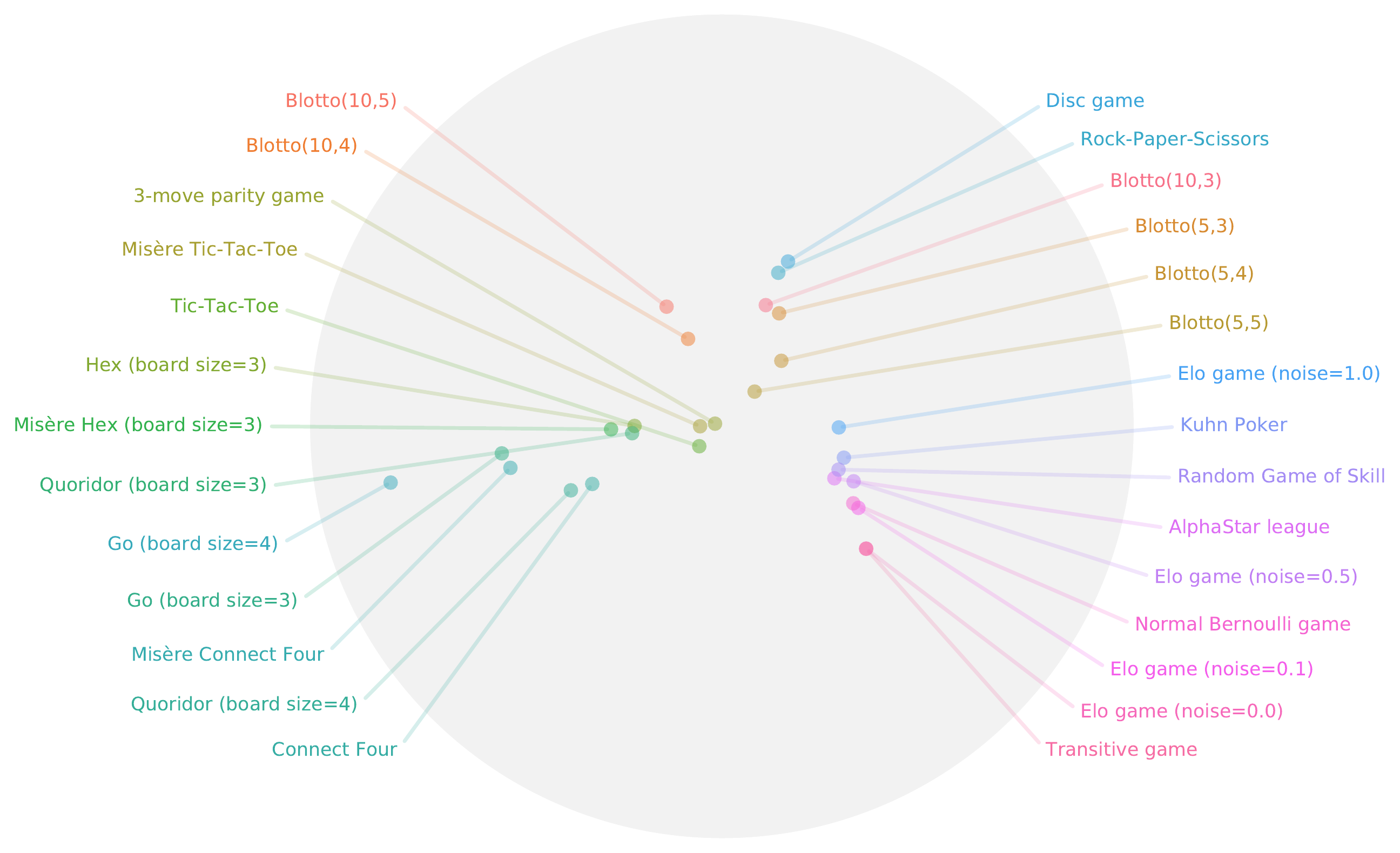}\\
        \label{fig:rwg_embeddings_bootstrap_trial_1_mixpolicies_True_nummix_100}
    \end{subfigure}\\
    \begin{subfigure}{0.5\textwidth}
        \centering
        \caption{}
        \includegraphics[width=\textwidth]{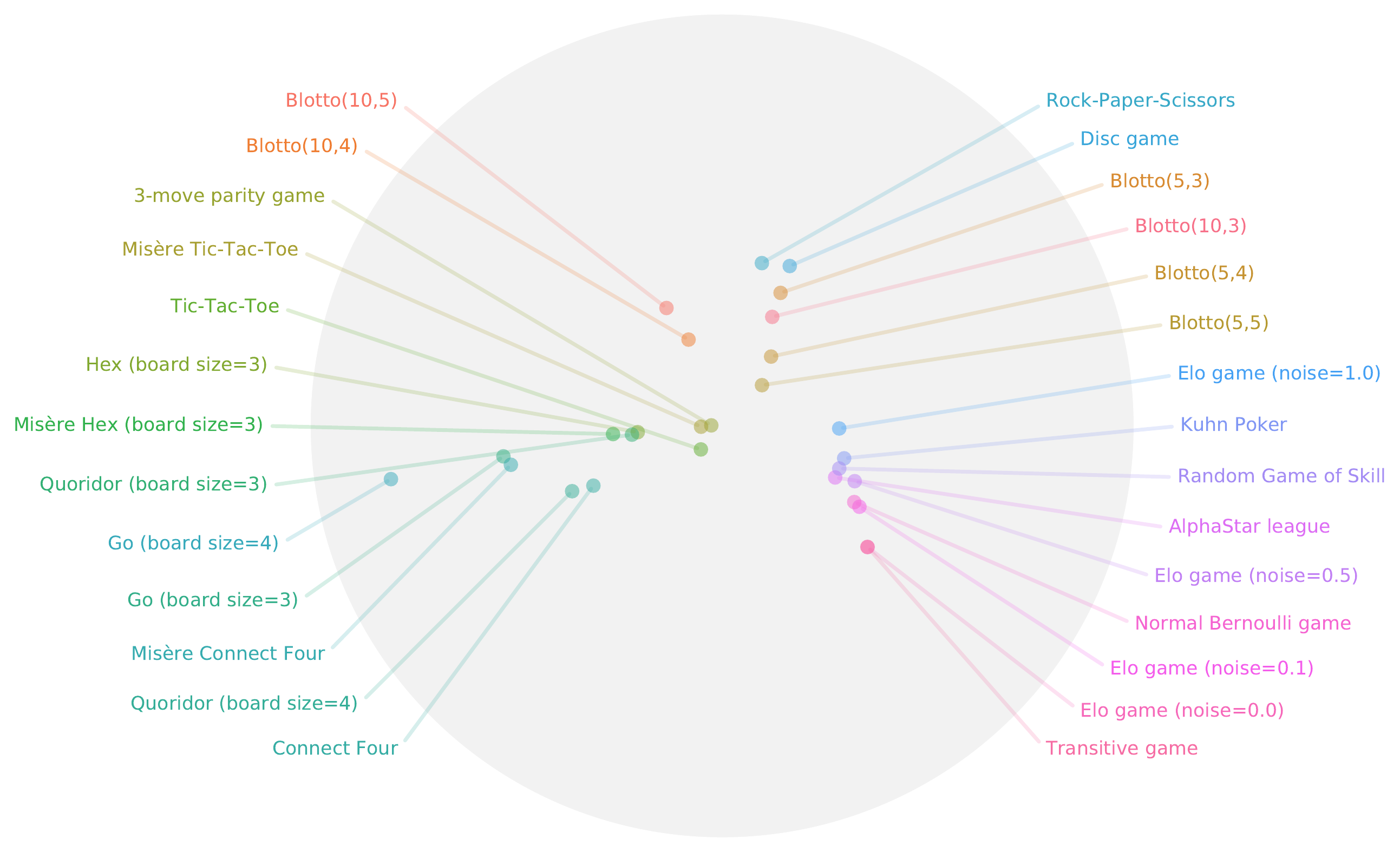}\\
        \label{fig:rwg_embeddings_bootstrap_trial_2_mixpolicies_True_nummix_100}
    \end{subfigure}%
    \begin{subfigure}{0.5\textwidth}
        \centering
        \caption{}
        \includegraphics[width=\textwidth]{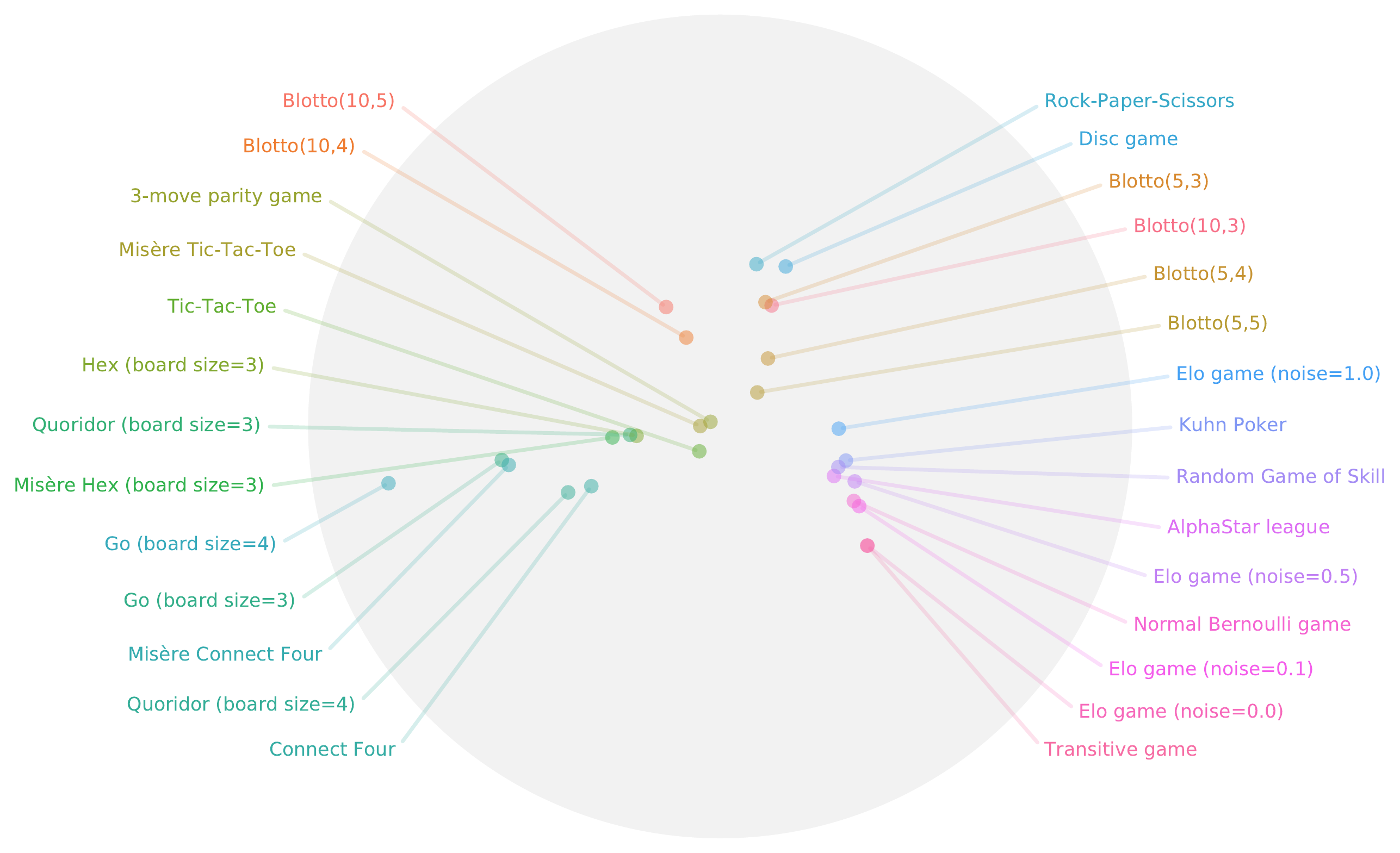}\\
        \label{fig:rwg_embeddings_bootstrap_trial_3_mixpolicies_True_nummix_100}
    \end{subfigure}\\
    \caption{Sensitivity to random mixtures of policies. In each figure, 100 additional policies are included per game, and the landscape subsequently regenerated. Note that game colors are kept the same as the original landscape of games visual (\cref{fig:rwg_embeddings_normCoG_True}) for easier comparison.
    \subref{fig:rwg_embeddings_bootstrap_trial_0_mixpolicies_True_nummix_100}--\subref{fig:rwg_embeddings_bootstrap_trial_3_mixpolicies_True_nummix_100} illustrate four such trials.
    }
    \label{fig:sensitivity_analysis_mixpolicies}
\end{figure}

It is interesting to consider the effect that additional mixing of policies would have on the results, and conduct a suite of experiments focused on this here.
Specifically, for all games in the landscape, we expand each payoff table by adding mixtures of policies.
For each game, we uniformly sample a mixture over a random selection of half of the base policies; 
we repeat this 100 times, expanding the payoff table accordingly.
For Rock--Paper--Scissors (due to the small size of the strategy space), we sample mixed policies uniformly over the full support of all 3 base strategies.
\cref{fig:sensitivity_analysis_mixpolicies} shows the results for 4 independent trials of policy mixing.

Some observations can be made here, in comparing the trials to one another (and to the original landscape, visualized in \cref{fig:rwg_embeddings_normCoG_True}).
At a high-level, most of the prominent clusters found in the landscape presented in the main paper are also present here, with some specific details as follows:
1) Rock--Paper--Scissors and the Disc game (closely clustered in both, despite the size of Rock--Paper--Scissors increasing from 3 $\times$ 3 in the original landscape to 103 $\times$ 103 in the mixed policy landscapes) 
2) Elo game(noise=0.0) and Transitive game;
3) Elo game(noise=0.1) and the Normal Bernoulli game;
4) Random Game of Skill, Elo game (noise=0.5), and the AlphaStar League;
5) Real-world games (e.g., Connect Four and Quoridor (board size=4));
6) The cluster of Blotto games have somewhat shifted towards that of Rock--Paper--Scissors and the Disc Game. 
Notably, all of these games are highly cyclical, and Blotto requires players to play uniformly across all permutations of the token-selection strategies (due to the game rules itself being permutation-invariant), a characteristic shared by Rock--Paper--Scissors.

\subsection*{A closer look at generated games}

\begin{figure}[t]
    \centering
    \includegraphics[width=.9\textwidth]{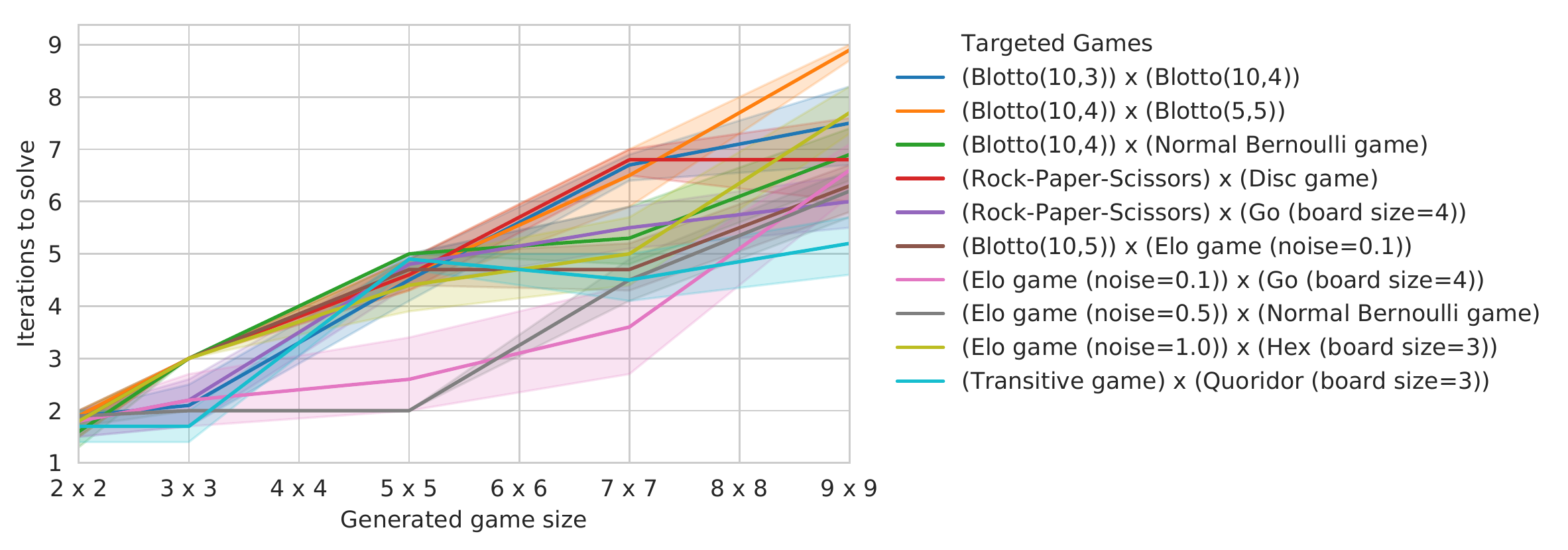}%
    \caption{Iterations to solve procedurally-generated game structures of varying sizes. Error ranges indicate 95\% confidence intervals over 10 trials.}
    \label{fig:gen_game_double_oracle}
\end{figure}

While raw hybrids of games alone are unlikely to play a key role in establishing useful curricula, we note that discovery of core features or mechanisms in games has, indeed, been a key driver of generation of new games (see, e.g., \citet{shaker2013automatic,charity2020mech,khalifa2019intentional}).
As such, we can observe interesting trends even in the specific class of normal-form games that we focus on here.

To more closely analyze trends in the games generated in the main paper, we conduct experiments that procedurally generate and subsequently train policies on these games via the double oracle algorithm. 
Specifically, we create a large class of generated games as follows. 
We first randomly sample 10 pairs of target games from the base games shown in the landscape of games (\cref{fig:rwg_embeddings_normCoG_True}).
We subsequently procedurally generate games of sizes 2 $\times$ 2, 3 $\times$ 3, 5 $\times$ 5, 7 $\times$ 7, and 9 $\times$ 9 targeting each sampled pair.
This yields a total of 50 generated games of varying characteristics.
We subsequently evaluate the complexity of solving these games via double oracle (of course, these can also be used for analysis of the dynamics of multiagent learning algorithms, as done in recent works such as \citet{BloembergenTHK15,hennes2020neural}.
\cref{fig:gen_game_double_oracle} summarizes the results of this experiment, with each generated instance evaluated using 10 randomly-seeded trials of the double oracle algorithm.

We notice several trends in these generated games.
First, the generated games targeting the three pairs (Blotto(10,3) $\times$ Blotto(10,4)), (Blotto(10,4) $\times$ Blotto(5,5)), and (Rock--Paper--Scissors $\times$ Disc game) consistently require the largest number of iterations to solve, across all game sizes.
This result is explained by the fact that each of these games targets a mixture over pairs of highly-cyclical underlying games; thus, each generated game likely has a large Nash support (despite the noise in the generation process), making them difficult to solve.
Next, we observe a lower bound roughly established by generated games involving the target pairs (Elo game (noise=0.1) $\times$ Go (board size=4)) and (Elo game (noise=0.5) $\times$ Normal Bernoulli Game).
Here, the noisy Elo game plays an important role, as its (roughly) transitive structure tends to reduce the number of strong strategies, thereby making it easier to solve.
Interestingly, for the largest generated game size (9 $\times$ 9), the pair (Transitive game $\times$ Quoridor (board size=3)) requires the fewest iterations to solve.
Effectively, the lack of noise in the Transitive game dominates the structure of these larger generated games, in contrast to the noisy Elo games.
Finally, the remaining generated games form a cluster with an intermediate number of iterations needed to solve them.
This latter class of games primarily targets a mixture of cyclical and transitive games (e.g., (Blotto(10,5) $\times$ Elo game (noise=0.1)), linking to the notion of interestingness from an AI perspective, as discussed in the main text.

Overall, generation of even the simple classes of games considered here exemplifies how one might use such an analysis to highlight relationships between generated games and existing, well-studied ones, to better understand and expand the space of multiplayer games. 

\subsection*{Motivating examples: results for randomly-generated games}
\begin{figure}[t]
    \def\mWidth{0.14\textwidth}
    \setcounter{subfigure}{0}%
    \centering
    \begin{tabular}{p{0.0in}M{\mWidth}M{\mWidth}M{\mWidth}M{\mWidth}M{\mWidth}M{\mWidth}}
        &Payoffs & Response graph & Cycle distribution & Spectral graph & Clustered graph & Contracted graph\\
        \gameRowMotivating{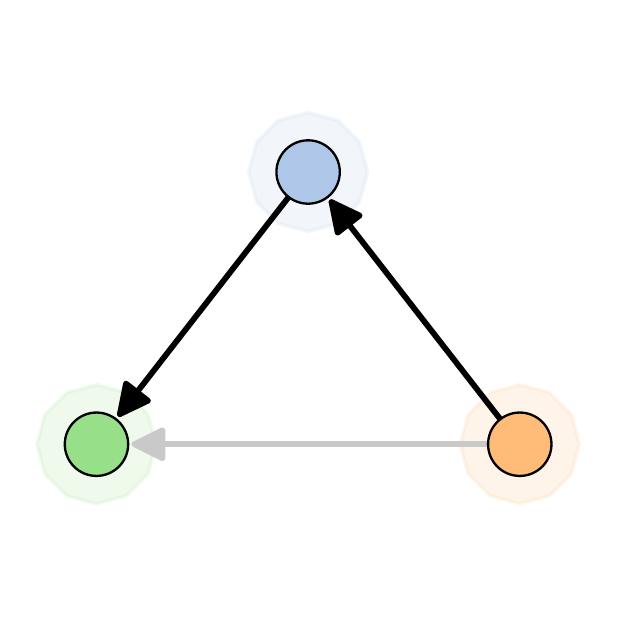}{Random game 1}
        \gameRowMotivating{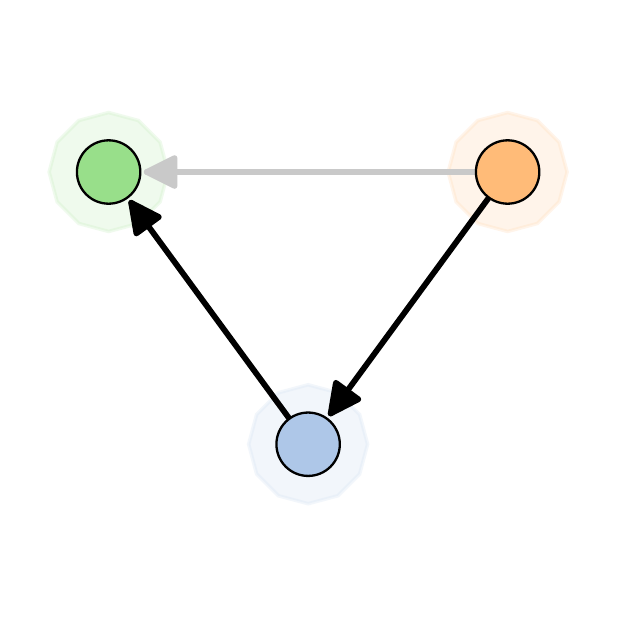}{Random game 2}
        \gameRowMotivating{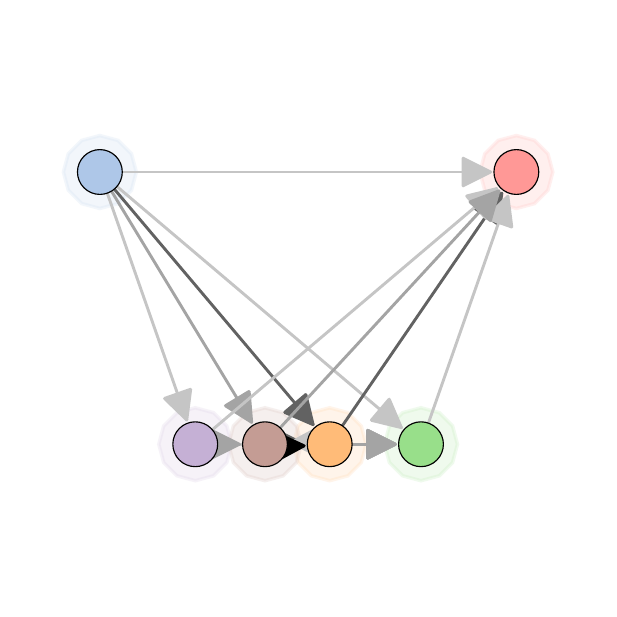}{Random game 3}
    \end{tabular}
    \caption{Results for randomly-generated games. Each column visualizes a different characteristic of the game or response graph, as discussed in the main text.}
    \label{fig:random_game_complexity_overview}
\end{figure}

\Cref{fig:random_game_complexity_overview} illustrates a sample of response graphs for generated games of random structure, as discussed in the Motivating Examples section of the main text.

\subsection*{Additional response graph analysis}
\begin{figure}[t]
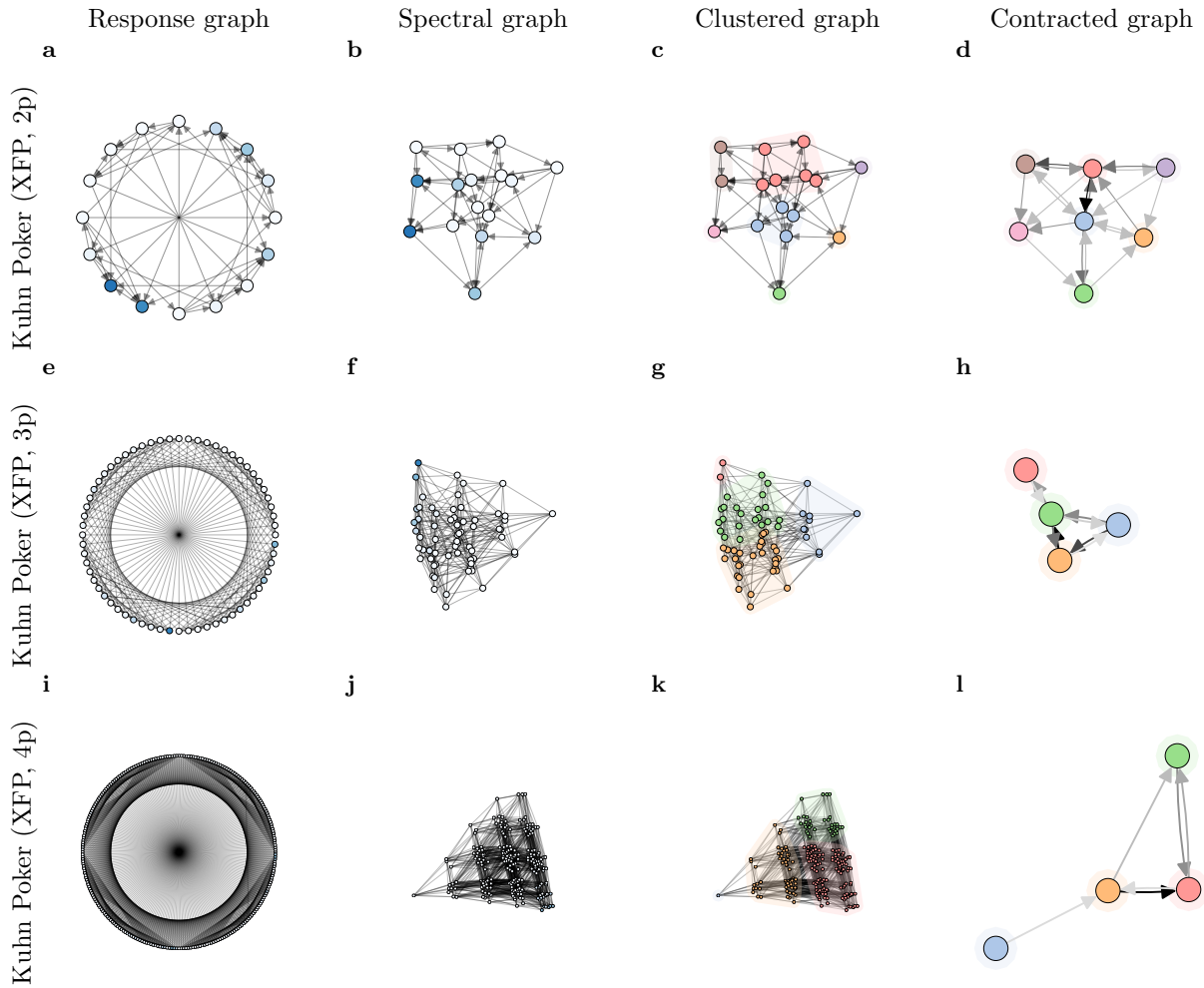

    \setcounter{subfigure}{0}%
    \centering
    \begin{tabular}{p{0.0in}cccc}
        &Response graph&Spectral graph & Clustered graph & Contracted graph\\
        \gameRowFiletypeNoPayoffs{kuhn_poker_2p}{Kuhn Poker (XFP, 2p)}{pdf}
        \gameRowFiletypeNoPayoffs{kuhn_poker_3p}{Kuhn Poker (XFP, 3p)}{pdf}
        \gameRowFiletypeNoPayoffs{kuhn_poker_4p}{Kuhn Poker (XFP, 4p)}{pdf}
    \end{tabular}
    \caption{Additional response graph analysis results I. Specifically, these results pertain to Kuhn Poker agents trained via extensive-form fictitious play (XFP)~\citep{heinrich2015fictitious}, with empirical games constructed as detailed in~\citet{omidshafiei2019alpha}.}
    \label{fig:additional_rg_results_1}
\end{figure}

For completeness, \cref{fig:additional_rg_results_1,fig:additional_rg_results_2,fig:additional_rg_results_3,fig:additional_rg_results_4,fig:additional_rg_results_5,fig:additional_rg_results_6} present the response graph-based analysis for the additional games considered in the main text.
\Cref{fig:additional_rg_results_1} is of particular note here, as it exemplifies the application of the methodology to asymmetric, many-player games.
Specifically, the empirical games constructed here correspond to agents trained via extensive-form fictitious play (XFP)~\citep{heinrich2015fictitious} in 2-, 3-, and 4-player variants of Kuhn Poker.  

\begin{figure}[t]
    \setcounter{subfigure}{0}%
    \centering
    \begin{tabular}{p{0.0in}cccc}
        &Payoffs & Spectral graph & Clustered graph & Contracted graph\\
        \gameRowFiletype{5_4_Blotto}{Blotto(5,4)}{pdf}
        \gameRowFiletype{5_5_Blotto}{Blotto(5,5)}{png}
        \gameRowFiletype{10_4_Blotto}{Blotto(10,4)}{png}
        \gameRowFiletype{10_5_Blotto}{Blotto(10,5)}{png}
    \end{tabular}
    \caption{Additional response graph analysis results II. Each column visualizes a different characteristic of the game or response graph, as discussed in the main text.}
    \label{fig:additional_rg_results_2}
\end{figure}

\begin{figure}[t]
    \setcounter{subfigure}{0}%
    \centering
    \begin{tabular}{p{0.0in}cccc}
        &Payoffs & Spectral graph & Clustered graph & Contracted graph\\
        \gameRowFiletype{3_move_parity_game_2}{3 move parity game}{png}
        \gameRowFiletype{Kuhn_poker}{Kuhn Poker}{pdf}
        \gameRowFiletype{Disc_game}{Disc game}{png}
        \gameRowFiletype{Normal_Bernoulli_game}{Normal Bernoulli Game}{png}
        \gameRowFiletype{Random_game_of_skill}{Random game of skill}{png}
    \end{tabular}
    \caption{Additional response graph analysis results III. Each column visualizes a different characteristic of the game or response graph, as discussed in the main text.}
    \label{fig:additional_rg_results_3}
\end{figure}

\begin{figure}[t]
    \setcounter{subfigure}{0}%
    \centering
    \begin{tabular}{p{0.0in}cccc}
        &Payoffs & Spectral graph & Clustered graph & Contracted graph\\
        \gameRowFiletype{Transitive_game}{Transitive game}{png}
        \gameRowFiletype{Elo_game}{Elo game (noise=0.0)}{png}
        \gameRowFiletype{Elo_game___noise_0_1}{Elo game (noise=0.1)}{png}
        \gameRowFiletype{Elo_game___noise_0_5}{Elo game (noise=0.5)}{png}
        \gameRowFiletype{Elo_game___noise_1_0}{Elo game (noise=1.0)}{png}
    \end{tabular}
    \caption{Additional response graph analysis results IV. Each column visualizes a different characteristic of the game or response graph, as discussed in the main text.}
    \label{fig:additional_rg_results_4}
\end{figure}

\begin{figure}[t]
    \setcounter{subfigure}{0}%
    \centering
    \begin{tabular}{p{0.0in}cccc}
        &Payoffs & Spectral graph & Clustered graph & Contracted graph\\
        \gameRowFiletype{connect_four}{Connect Four}{png}
        \gameRowFiletype{misere_game_connect_four___}{Mis\`ere Connect Four}{png}
        \gameRowFiletype{hex_board_size_3_}{Hex (board size=3)}{png}
        \gameRowFiletype{misere_game_hex_board_size_3__}{Mis\`ere Hex (board size=3)}{png}
        \gameRowFiletype{go_board_size_4_komi_6_5_}{Go (board size=4)}{png}
    \end{tabular}
    \caption{Additional response graph analysis results V. Each column visualizes a different characteristic of the game or response graph, as discussed in the main text.}
    \label{fig:additional_rg_results_5}
\end{figure}

\begin{figure}[t]
    \setcounter{subfigure}{0}%
    \centering
    \begin{tabular}{p{0.0in}cccc}
        &Payoffs & Spectral graph & Clustered graph & Contracted graph\\
        \gameRowFiletype{tic_tac_toe}{Tic--Tac--Toe}{png}
        \gameRowFiletype{misere_game_tic_tac_toe___}{Mis\`ere Tic--Tac--Toe}{png}
        \gameRowFiletype{quoridor_board_size_3_}{Quoridor (board size=3)}{png}
        \gameRowFiletype{quoridor_board_size_4_}{Quoridor (board size=4)}{png}
    \end{tabular}
    \caption{Additional response graph analysis results VI. Each column visualizes a different characteristic of the game or response graph, as discussed in the main text.}
    \label{fig:additional_rg_results_6}
\end{figure}

\subsection*{Additional complexity results}
\begin{figure}[t]
    \centering
    \includegraphics[width=\textwidth]{figs/double_oracle/double_oracle_game_results_legend.pdf}
    \begin{subfigure}{0.33\textwidth}
        \centering
        \caption{}
        \includegraphics[width=\textwidth]{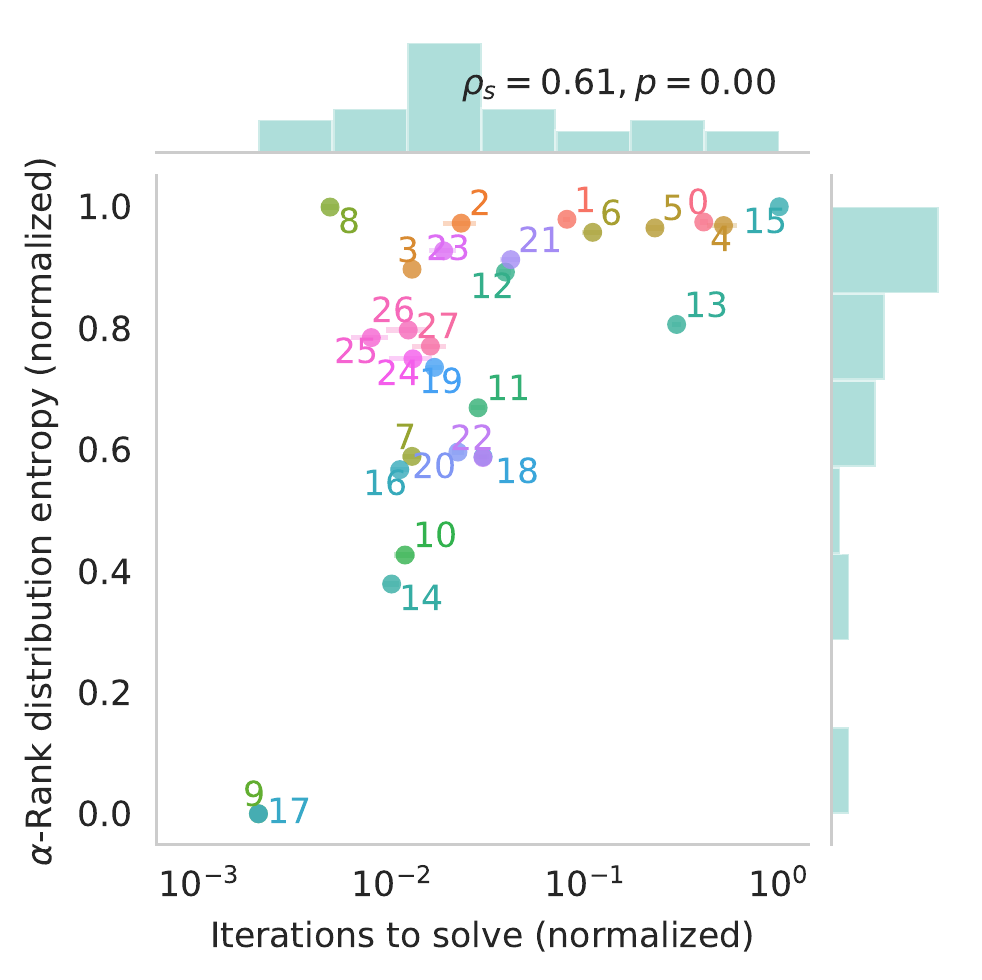}
        \label{fig:supp_info_double_oracle_game_results_alpharank_entropy}
    \end{subfigure}%
    \hspace{20pt}%
    \begin{subfigure}{0.33\textwidth}
        \centering
        \caption{}
        \includegraphics[width=\textwidth]{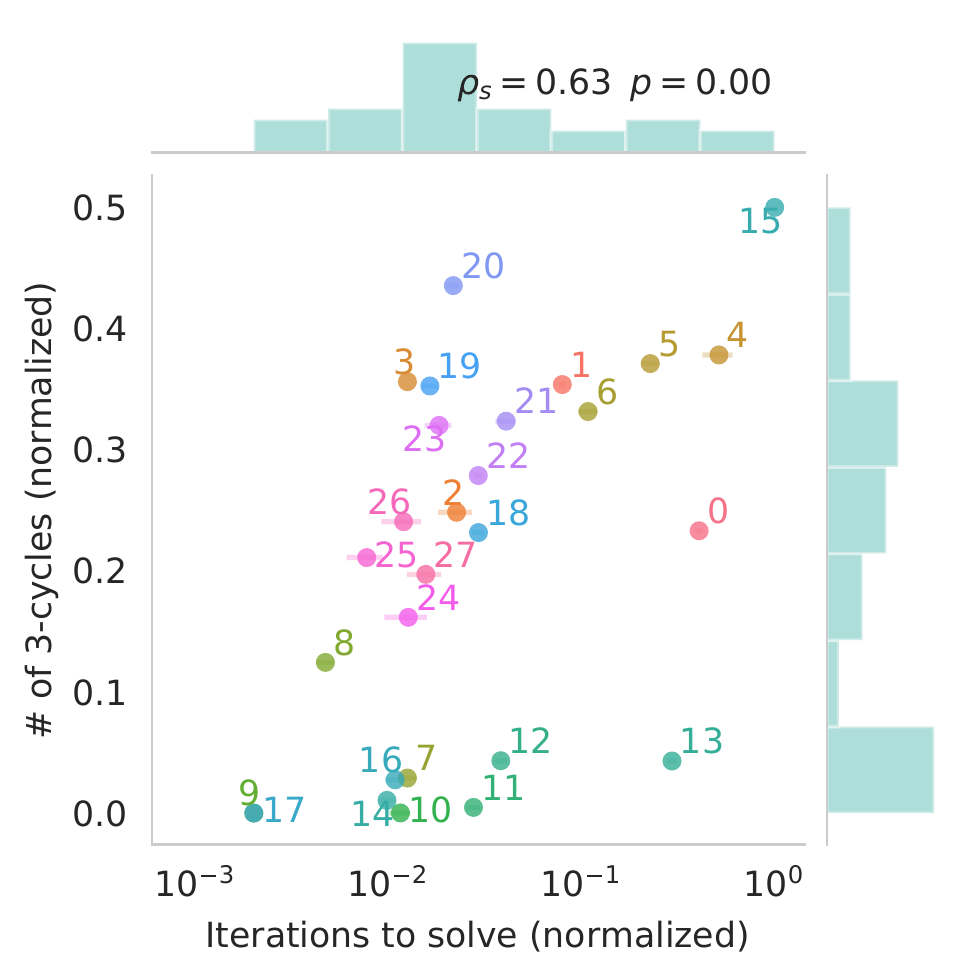}
        \label{fig:supp_info_double_oracle_game_results_num_3cycles}
    \end{subfigure}\\
    \begin{subfigure}{0.2\textwidth}
        \centering
        \caption{}
        \includegraphics[width=\textwidth]{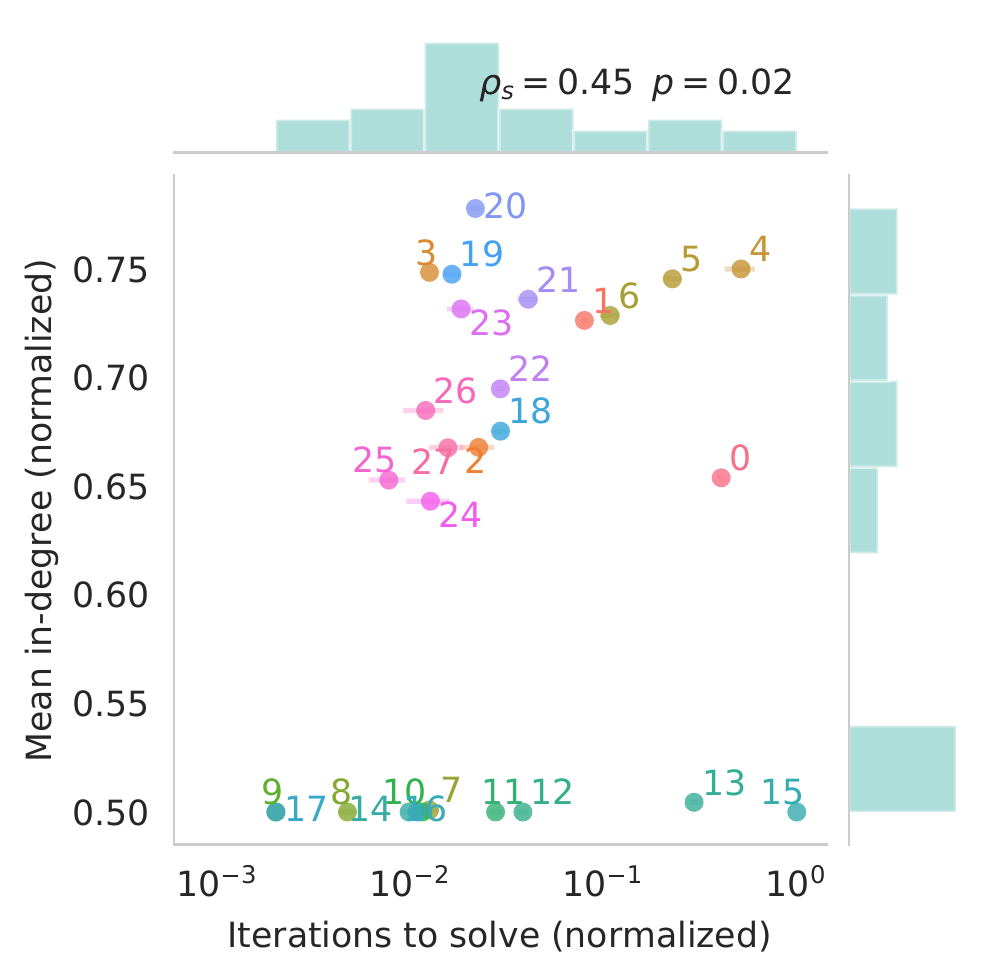}
        \label{fig:supp_info_double_oracle_game_results_indegree_stats_mean}
    \end{subfigure}%
    \hfill%
    \begin{subfigure}{0.2\textwidth}
        \centering
        \caption{}
        \includegraphics[width=\textwidth]{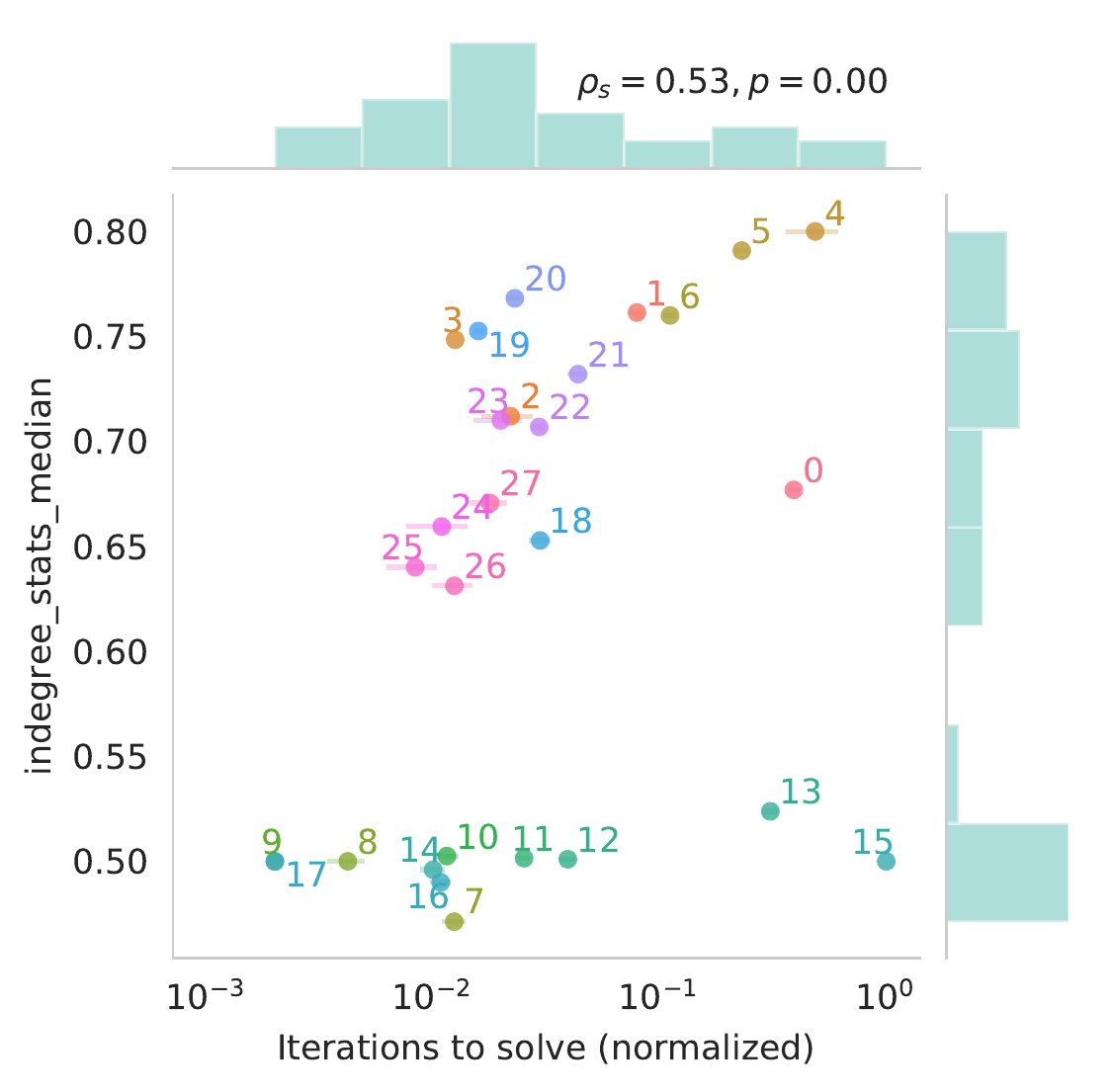}
    \end{subfigure}%
    \hfill%
    \begin{subfigure}{0.2\textwidth}
        \centering
        \caption{}
        \includegraphics[width=\textwidth]{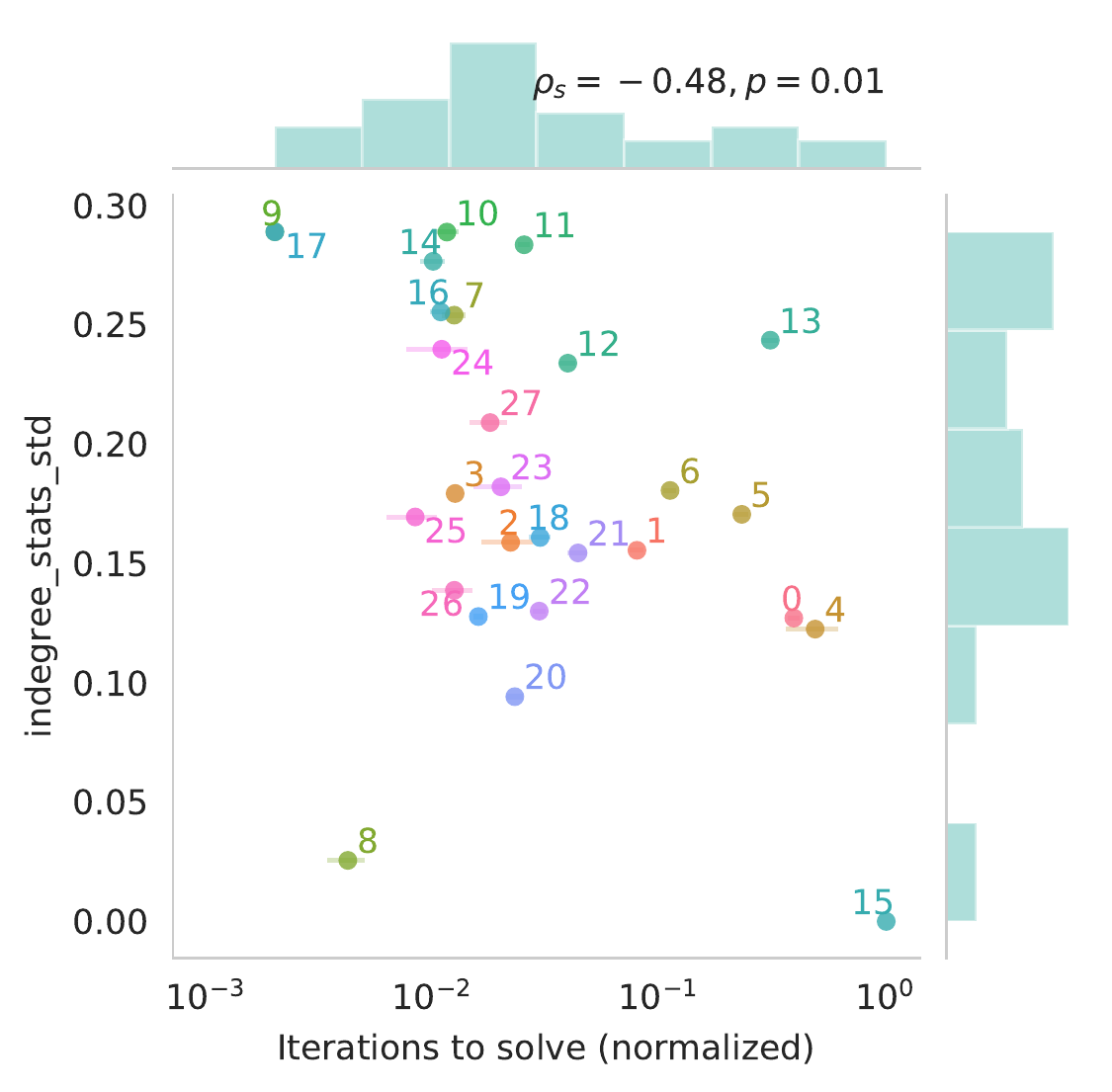}
    \end{subfigure}%
    \hfill%
    \begin{subfigure}{0.2\textwidth}
        \centering
        \caption{}
        \includegraphics[width=\textwidth]{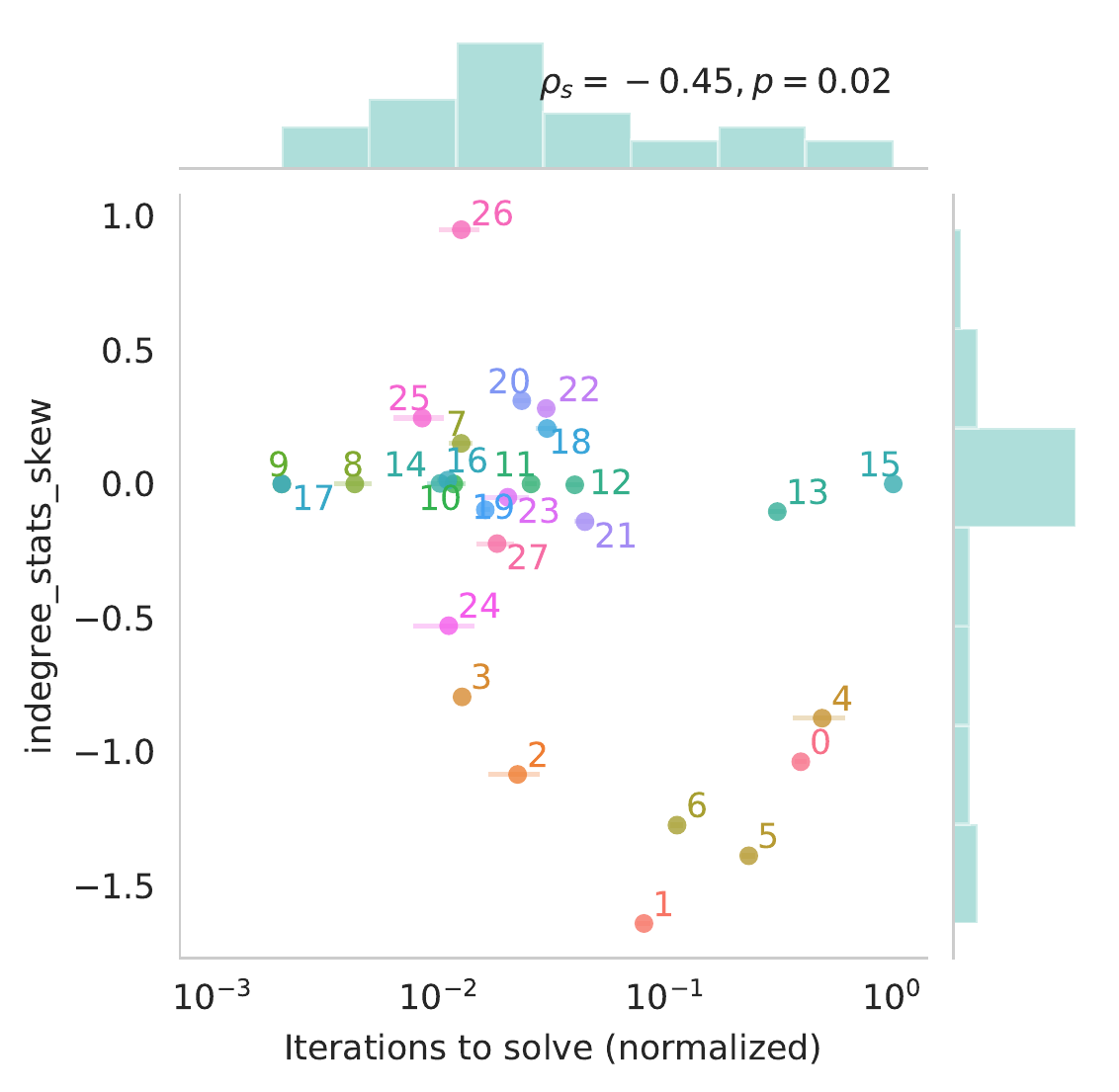}
    \end{subfigure}%
    \hfill%
    \begin{subfigure}{0.2\textwidth}
        \centering
        \caption{}
        \includegraphics[width=\textwidth]{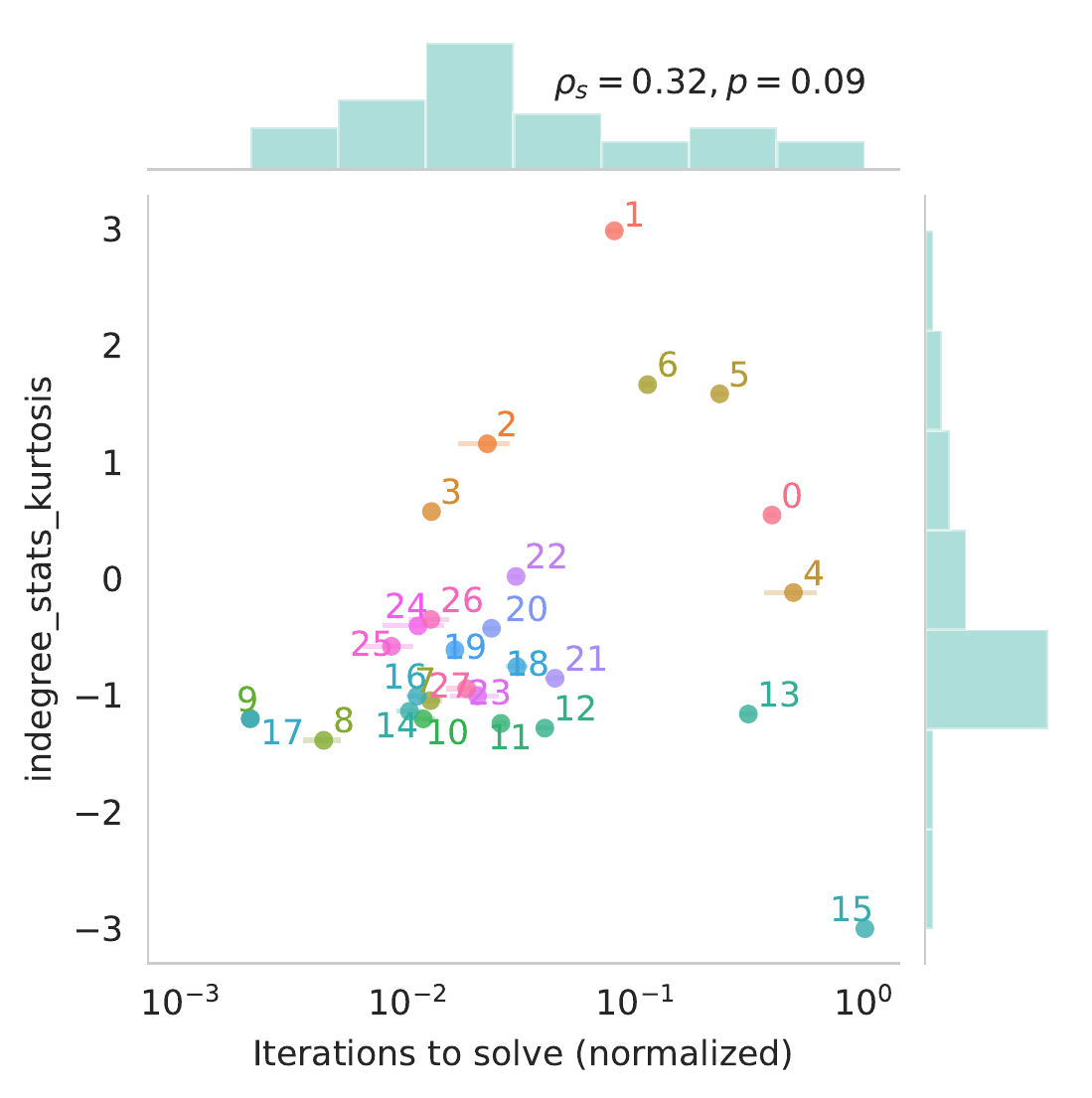}
    \end{subfigure}\\
    \begin{subfigure}{0.2\textwidth}
        \centering
        \caption{}
        \includegraphics[width=\textwidth]{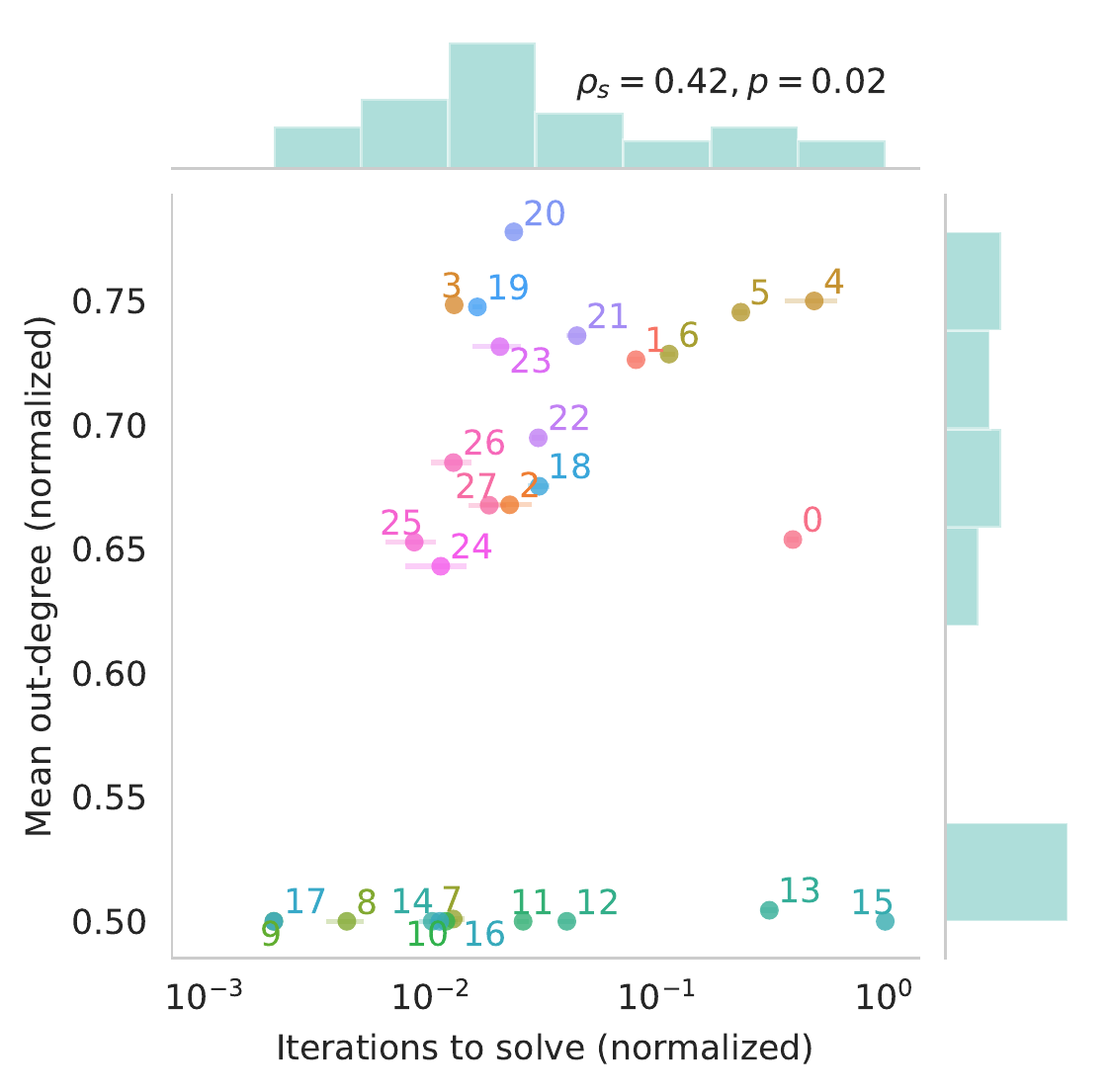}
        \label{fig:supp_info_double_oracle_game_results_outdegree_stats_mean}
    \end{subfigure}%
    \hfill%
    \begin{subfigure}{0.2\textwidth}
        \centering
        \caption{}
        \includegraphics[width=\textwidth]{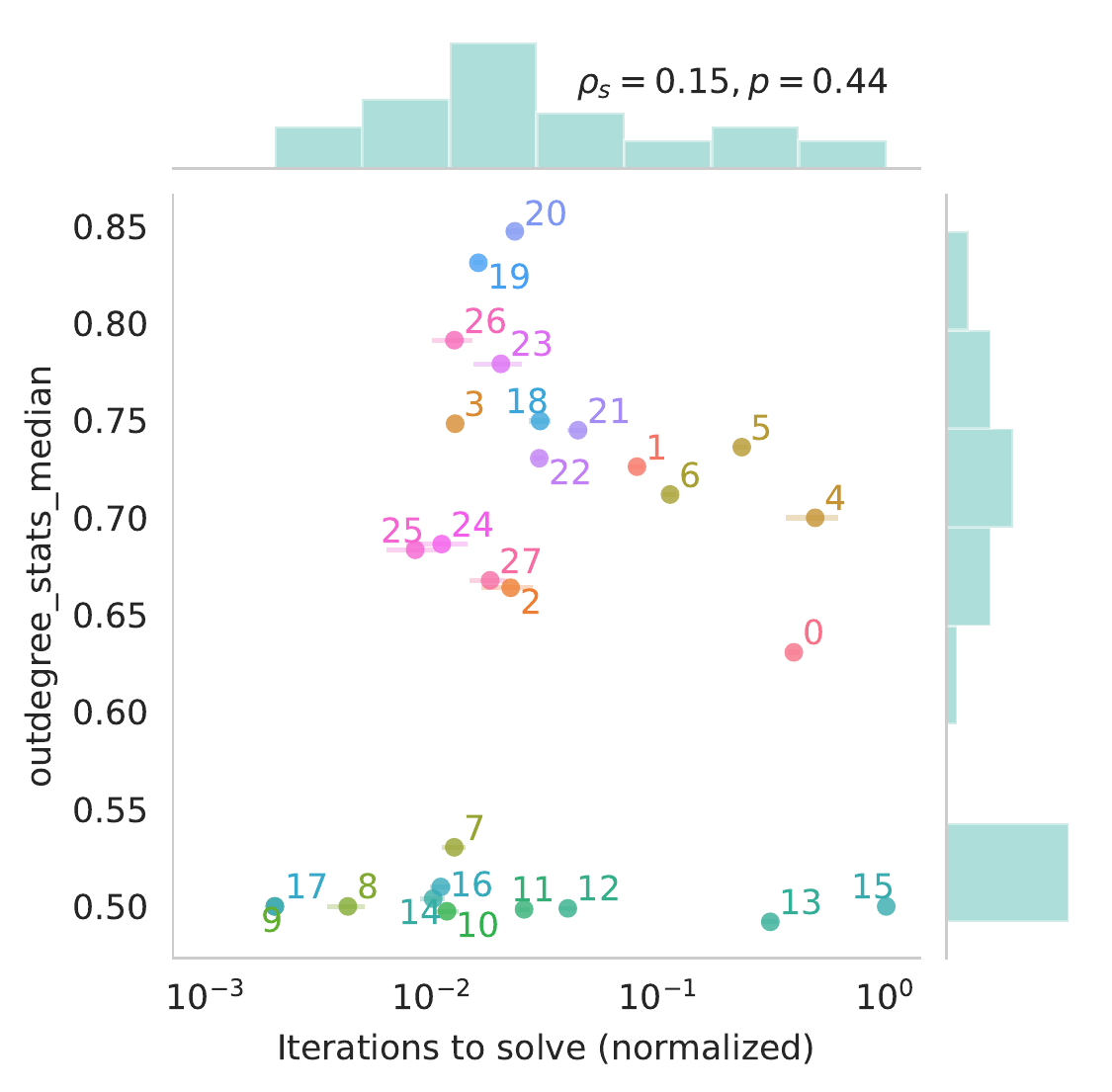}
    \end{subfigure}%
    \hfill%
    \begin{subfigure}{0.2\textwidth}
        \centering
        \caption{}
        \includegraphics[width=\textwidth]{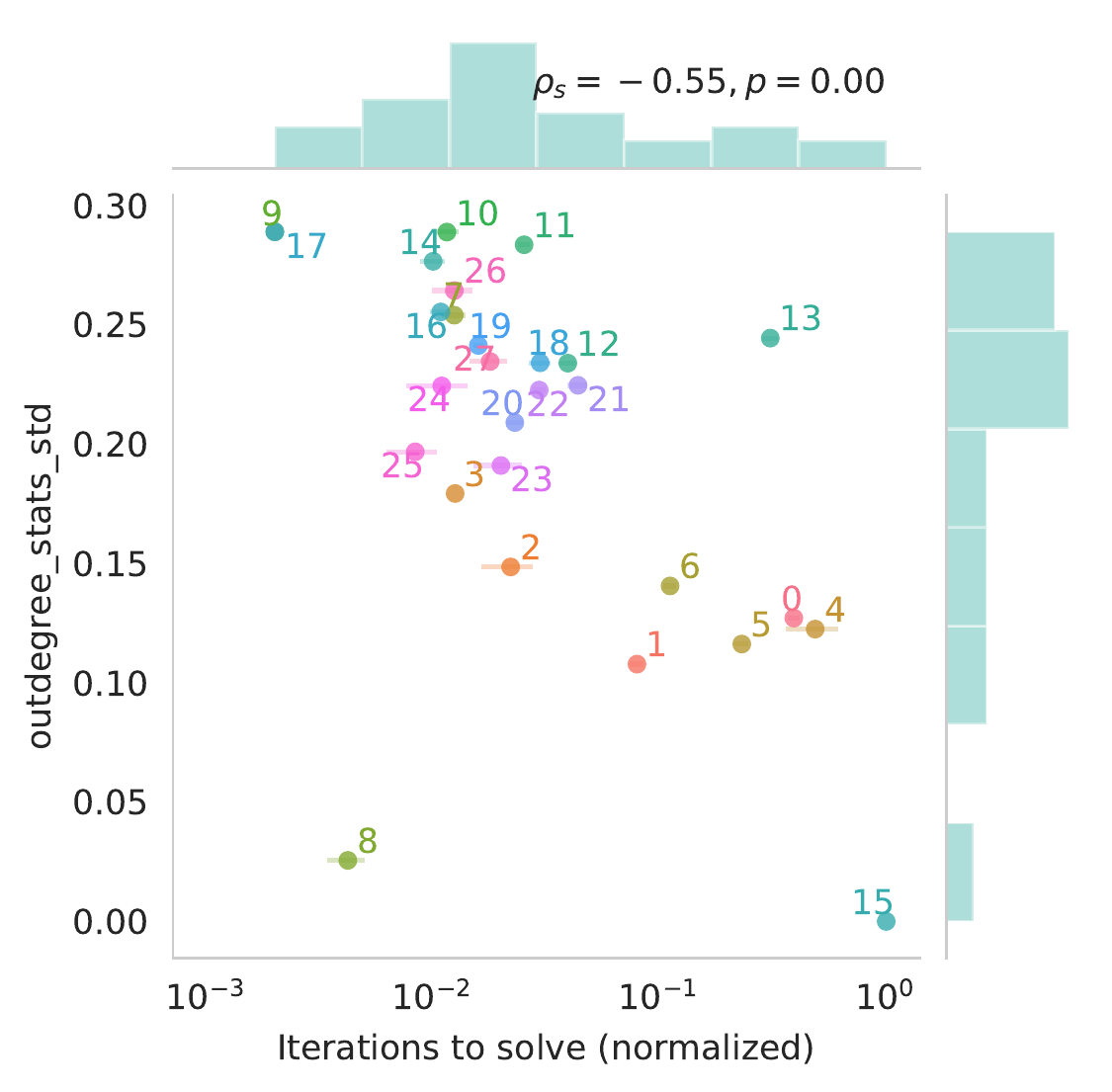}
    \end{subfigure}%
    \hfill%
    \begin{subfigure}{0.2\textwidth}
        \centering
        \caption{}
        \includegraphics[width=\textwidth]{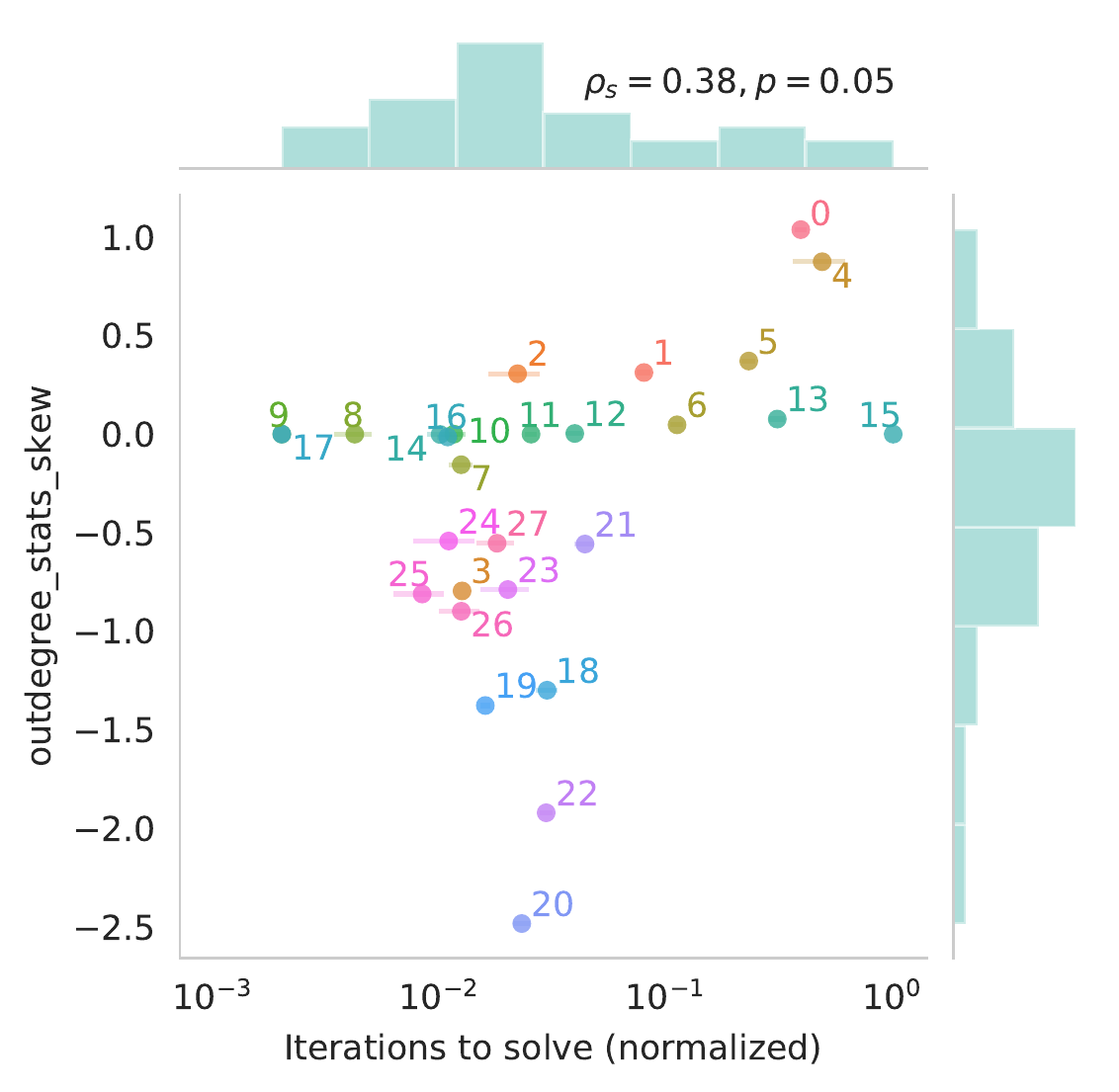}
    \end{subfigure}%
    \hfill%
    \begin{subfigure}{0.2\textwidth}
        \centering
        \caption{}
        \includegraphics[width=\textwidth]{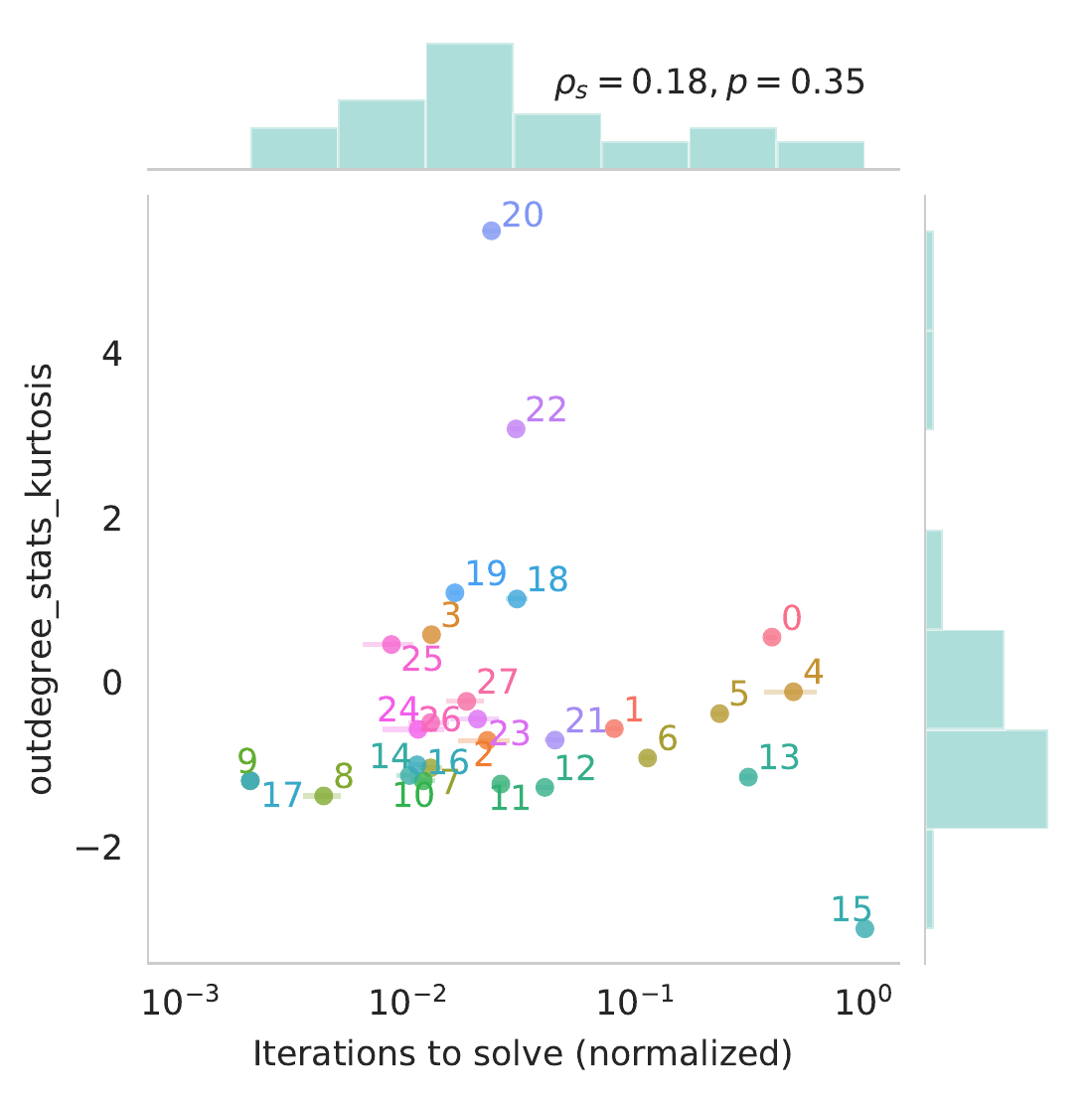}
    \end{subfigure}\\
    \caption{Response graph complexity vs. computational complexity of solving associated games. Each figure plots a respective measure of graph complexity against the normalized number of iterations needed to solve the associated game via the double oracle algorithm (with normalization done with respect to the total number of strategies in each underlying game).
    Note that mean node-wise in- and out-degrees (in \subref{fig:supp_info_double_oracle_game_results_indegree_stats_mean} and 
    \subref{fig:supp_info_double_oracle_game_results_outdegree_stats_mean}, respectively) match across the games here due to the degree sum formula;
    other distributional statistics, however, do not necessarily match across in- and out-degrees, as evident above.
    }
    \label{fig:supp_graph_vs_computational_complexity}
\end{figure}

\Cref{fig:supp_graph_vs_computational_complexity} provides an overview of additional response graph-based measures, in comparison to the normalized number of iterations required to solve each of the games considered in the main text.

